\PassOptionsToPackage{T2A}{fontenc}

\documentclass[onecolumn]{cohere}

\usepackage[utf8]{inputenc}

\usepackage{xcolor}
\usepackage{rotating}

\usepackage{lipsum}
\usepackage{pdfpages}
\usepackage{colortbl}
\usepackage{tabulary}
\usepackage{longtable}
\usepackage{array}
\usepackage{placeins}
\usepackage{tablefootnote}
\usepackage{tcolorbox}
\usepackage{wrapfig}
\usepackage{caption}
\usepackage{breqn} 
\usepackage{booktabs}

\usepackage{pifont}
\definecolor{deeppurple}{HTML}{9e02f7}
\definecolor{forestgreen}{HTML}{2e7d43}
\usepackage{multirow}
\usepackage{tabularx}
\usepackage{longtable}
\usepackage{tikz}
\usepackage{xcolor}
\usepackage{nicematrix}

\usepackage{arydshln}
\usepackage{xspace} 
\usepackage{makecell} 

\usepackage[framemethod=TikZ]{mdframed}
\usepackage{tcolorbox}
\usepackage{multicol}
\usepackage{multirow}
\usepackage{float}

\newtcolorbox{mybox}[2][]{
  colback=white, 
  colframe=lightblue,
  fonttitle=\bfseries,
  coltitle=black,  
  title=#2, 
  #1
}

\definecolor{ayad}{RGB}{148, 156, 229} 
\definecolor{ayadsymbol}{RGB}{76, 110, 230} 
\definecolor{lightblue}{RGB}{211, 227, 252} 

\definecolor{bgblue}{RGB}{247, 250, 255} 
\newcommand*\colourcheck[1]{%
  \expandafter\newcommand\csname #1check\endcsname{\textcolor{#1}{\ding{52}}}%
}
\newcommand*\colourcross[1]{%
  \expandafter\newcommand\csname #1cross\endcsname{\textcolor{#1}{\ding{55}}}%
}

\DeclareSymbolFont{extraup}{U}{zavm}{m}{n}
\DeclareMathSymbol{\vardiamond}{\mathalpha}{extraup}{87}
\definecolor{ayadsymbol}{RGB}{76, 110, 230} 

\colourcheck{blue}
\colourcheck{green}
\colourcheck{red}

\colourcross{blue}
\colourcross{green}
\colourcross{red}
\usepackage{nicematrix}
\usepackage{comment}
\usepackage{amssymb}
\usepackage{float}
\usepackage{hyperref}
\usepackage{adjustbox}
\usepackage{booktabs}
\usepackage{multirow}
\usepackage{tcolorbox}
\usepackage{graphicx}
\usepackage{wrapfig}

\setcitestyle{number,square}

\title{\includegraphics[scale=0.2]{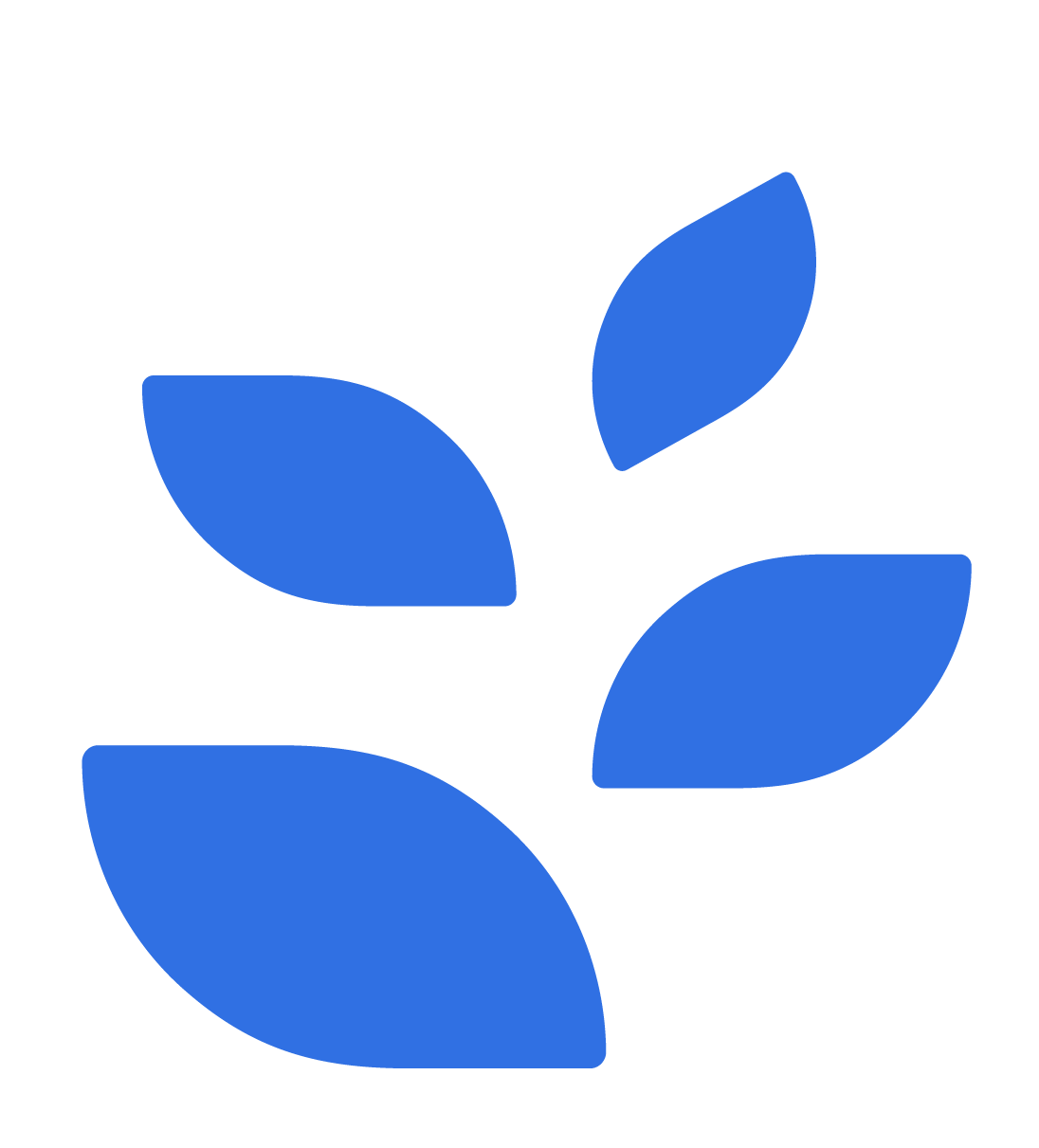}Aya Vision: Advancing the Frontier of Multilingual Multimodality}

\renewcommand{\fasymbol}{\textcolor{Cerulean}{$\vardiamond$}}
\renewcommand{\fasymbola}{\textsuperscript{\textcolor{NavyBlue}{$\bigstar$}}}

\multiauthors
\author{name={Saurabh Dash\faa},affiliation={1}}
\author{name={Yiyang Nan\faa},affiliation={1}}
\author{name={John Dang},affiliation={1}}
\author{name={Arash Ahmadian},affiliation={1,2}}
\author{name={Shivalika Singh},affiliation={1}}
\author{name={Madeline Smith},affiliation={1}}
\author{name={Bharat Venkitesh},affiliation={2}}
\author{name={Vlad Shmyhlo},affiliation={2}}
\author{name={Viraat Aryabumi},affiliation={2}}
\author{name={Walter Beller-Morales},affiliation={2}}
\author{name={Jeremy Pekmez},affiliation={2}}
\author{name={Jason Ozuzu},affiliation={2}}
\author{name={Pierre Richemond},affiliation={2}}
\author{name={Acyr Locatelli},affiliation={2}}
\author{name={Nick Frosst},affiliation={2}}
\author{name={Phil Blunsom},affiliation={2}}
\author{name={Aidan Gomez},affiliation={2}}
\author{name={Ivan Zhang},affiliation={2}}
\author{name={Marzieh Fadaee},affiliation={1}}
\author{name={Manoj Govindassamy},affiliation={2}}
\author{name={Sudip Roy},affiliation={2}}
\author{name={Matthias Gallé\fa},affiliation={1}}
\author{name={Beyza Ermis\fa},affiliation={1}}
\author{name={Ahmet Üstün\fa},affiliation={1}}
\author{name={Sara Hooker\fa},affiliation={1}}
\affiliations{
    \item[1] Cohere Labs
    \item[2] Cohere
}

\corresponding[*]{\{{saurabh}, {olivernan}, {matthias}, {beyza}, {ahmet}, {sarahooker}\}{@cohere.com}}

\date{\today}

\abstract{

Building multimodal language models is fundamentally challenging: it requires aligning vision and language modalities, curating high-quality instruction data, and avoiding the degradation of existing text-only capabilities once vision is introduced. These difficulties are further magnified in the multilingual setting, where the need for multimodal data in different languages exacerbates existing data scarcity, machine translation often distorts meaning, and catastrophic forgetting is more pronounced.
To address the aforementioned challenges, we introduce novel techniques spanning both data and modeling. First, we develop a synthetic annotation framework that curates high-quality, diverse multilingual multimodal instruction data, enabling Aya Vision models to produce natural, human-preferred responses to multimodal inputs across many languages. Complementing this, we propose a cross-modal model merging technique that mitigates catastrophic forgetting, effectively preserving text-only capabilities while simultaneously enhancing multimodal generative performance. Aya-Vision-8B achieves best-in-class performance compared to strong multimodal models such as Qwen-2.5-VL-7B, Pixtral-12B, and even much larger Llama-3.2-90B-Vision. We further scale this approach with Aya-Vision-32B, which outperforms models more than twice its size, such as Molmo-72B and LLaMA-3.2-90B-Vision. Our work advances multilingual progress on the multi-modal frontier, and provides insights into techniques that effectively bend the need for compute while delivering extremely high performance.

\textbf{Aya-Vision-8B}: \url{https://huggingface.co/CohereLabs/aya-vision-8B}

\textbf{Aya-Vision-32B}: \url{https://huggingface.co/CohereLabs/aya-vision-32B}

\textbf{AyaVisionBench}:
\url{https://huggingface.co/datasets/CohereLabs/AyaVisionBench}
}

\renewcommand{\FONTauthors}{\bfseries\Large}

\begin{document}

\begin{figure}[t]
    \centering
    \includegraphics[width=0.80\linewidth, trim={11cm 6cm 2cm 4cm},clip]{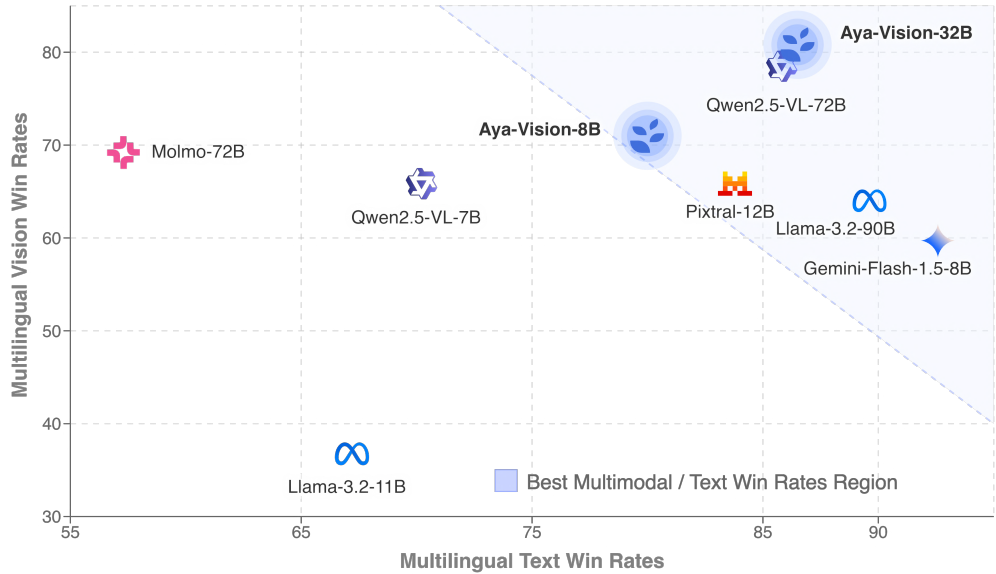}
    \caption{\textbf{Aya Vision models achieve state-of-the-art multilingual performance across both multimodal and text-only tasks.} We report multimodal and text-only win rates against Pangea-7B \citep{yue2024pangea}, averaged over 23 languages. Aya-Vision-8B achieves \textit{best-in-class} multimodal performance without compromising text capabilities, while Aya-Vision-32B outperforms all baselines, including much larger models such as Llama-3.2-90B-Vision \citep{grattafiori2024llama}, establishing an optimal balance between efficiency and cross-modal strength.} 
    \label{fig:enter-label}
\end{figure}
\section{Introduction}

\begin{quote}
    \textit{We do not describe the world we see, we see the world we can describe.} \textbf{--- René Descartes}
\end{quote}

Although multimodal large language models (MLLMs) \citep{liu2023visual, liu2024improved, deitke2024molmo, geminiteam2024gemini15unlockingmultimodal, laurenccon2024building, chen2024far, bai2025qwen25vl, kimiteam2025kimik15scalingreinforcement} have demonstrated remarkable success in jointly reasoning over various modalities, their performance remains predominantly confined to English. This linguistic limitation represents a substantial bottleneck in advancing multilingual AI, restricting global accessibility and impact.

Expanding multimodal models across languages exacerbates existing challenges at the frontier of AI. Foremost among these is the scarcity of high-quality multimodal data. While there has been expansion of languages served in language models \citep{ustun2024aya, aryabumi2024aya, dang2024aya, cohere2025commandaenterprisereadylarge}, the intersection of both images and languages remains severely underserved. 
High-quality multimodal instruction-tuning datasets are scarce and primarily composed of short, simplistic, task-oriented image-text pairs \citep{goyal2017making, wang2021screen2words, schwenk2022okvqa}. These datasets, while useful for benchmarking, inadequately prepare models for the rich, conversational scenarios encountered in real-world applications. 
Existing approaches primarily rely on machine translation to address this disparity \citep{li2023m, maaz2024palo, yue2024pangea}. However, translations often introduce linguistic artifacts (``translationese''), biases \citep{vanmassenhove2021machine, savoldi2021gender, hartung2023measuring, muennighoff2022crosslingual}, and fail to capture culture-specific nuances \citep{singh2024aya, salazar2025kaleidoscope}, contextual subtleties, and image-text alignments \citep{wang2022measuring, pudjiati2022post}. Creating high-quality, culturally and linguistically accurate multimodal instruction data across diverse languages thus remains an essential yet unsolved challenge.

The second significant challenge is the known tension between adding vision capabilities and maintaining robust text-only performance. Integrating vision modalities commonly results in catastrophic forgetting, where models lose previously acquired language skills \citep{qwenvl2023,deitke2024molmo,grattafiori2024llama,pozzobon2023goodtrieveradaptivetoxicitymitigation}. This decay is further amplified when expanding coverage across multiple languages. 

Equally pressing is the need for robust evaluations to measure progress. Any scientific pursuit requires a reliable metric of success. Existing multimodal benchmarks typically emphasize academic-style, multiple-choice tasks, evaluating models via rigid pattern-matching with predefined answer sets~\citep{changpinyo2022maxm, romero2024cvqa, yue2024pangea}. While useful for standardized comparisons, these fall short in capturing the nuanced, open-ended interactions that characterize real-world usage. Moreover, the few benchmarks that support more complex, open-ended interactions \citep{lu2024wildvision, agrawal2024pixtral} are currently only available in English-- leaving multilingual multimodal evaluation largely unexplored.

In this work, we tackle these challenges collectively. To address data scarcity, we replace naive translation pipelines with a hybrid method that pairs a specialized translation model with a larger LLM to correct and remove systematic translationese artifacts. We term this approach \textit{context-aware rephrasing}, which enables the creation of higher-quality, human-preferred multimodal instruction data. 
We also systematically explore the benefits of merging to mitigate catastrophic forgetting. We propose a \textbf{a novel cross-modal merging strategy} (\S~\ref{sec:balance}) that fuses capabilities across models, allowing for preservation and ``on-the-fly'' extension of capabilities across modalities. We see this as a powerful new paradigm to create adaptive models efficiently for new tasks. Our merging paradigm improves text-only tasks 50.2\% and multimodal tasks 20.5\% relative to the unmerged checkpoint, due to the inherent compositionality between the tasks and modalities. 

The result of our work is \textbf{Aya Vision}, a family of state-of-the-art multilingual multimodal models available in 8B and 32B sizes. In contrast to the many existing MLLMs, Aya Vision models are trained with a strong emphasis on multilingual and multimodal generation, yielding fluent chat performance. Aya-Vision-8B achieves \textit{best-in-class} performance, surpassing Qwen-2.5-VL-7B~\citep{bai2025qwen25vl}, Llama-3.2-11B-Vision~\citep{grattafiori2024llama}, Pixtral 12B~\citep{agrawal2024pixtral}, and Gemini-Flash-1.5-8B~\citep{geminiteam2024gemini15unlockingmultimodal}, with up to 79\% win rate across multimodal tasks in 23 languages. Aya-Vision-32B outcompetes models over twice its size, including Llama-3.2-90B-Vision~\citep{grattafiori2024llama}, Molmo-72B~\citep{deitke2024molmo}, and Qwen-2.5-VL-72B~\citep{bai2025qwen25vl}, with win rates up to 72.4\%.

\begin{figure}[t]
    \centering
    \includegraphics[width=0.80\linewidth, trim={11cm 6cm 2cm 4cm},clip]{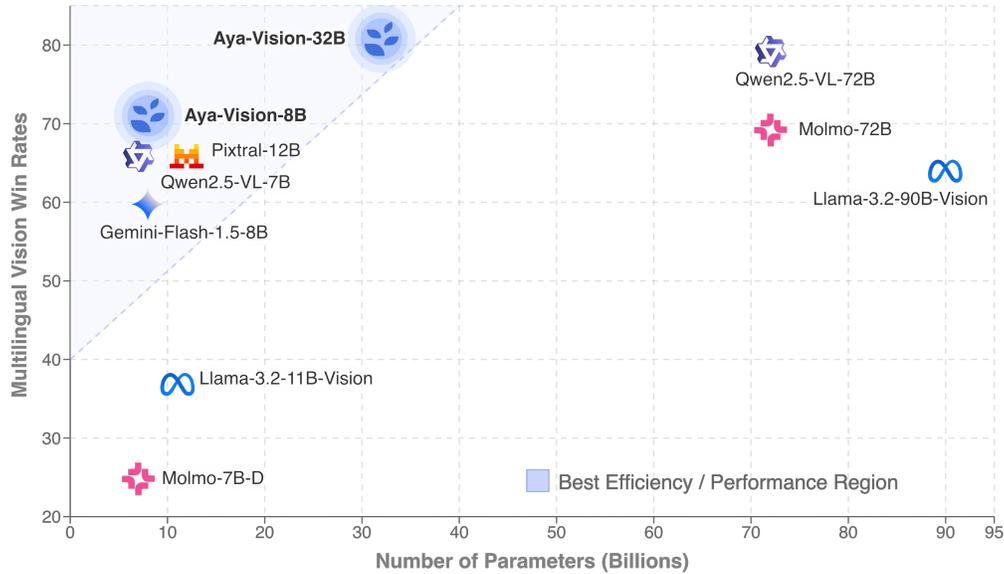}
    \caption{\textbf{Aya Vision establishes a new Pareto frontier in the performance-efficiency trade-off.} We show multimodal win rates against Pangea-7B, with respect to the number of parameters for each model.}
    \label{fig:aya-vision-best-region}
\end{figure}

Our primary contributions are as follows:
\begin{enumerate}
    \item \textbf{A family of state-of-the-art multilingual multimodal LLMs:} We introduce Aya-Vision-8B and 32B models, covering 23 languages spoken by half the worlds population. In contrast to the many existing multimodal LLMs, Aya Vision models are trained with a strong emphasis on multilingual, multimodal generation, yielding fluent chat performance preferred by humans. 
    
    \item \textbf{A novel multilingual multimodal synthetic annotation framework:} We develop a novel multilingual multimodal framework that combines synthetic data distillation, automated translation, and context-aware rephrasing to produce high-quality and diverse instruction data across languages, addressing data scarcity challenges.
    Recaptioning increases the average number of tokens from 27.2 to 140.8 and the measure of lexical diversity from 11.0 to 61.2. Our translation pipeline improves the translation quality by 11.24\% over the NLLB-3.3B \citep{costa2022no} translations. 
    
    \item \textbf{Optimizing performance across modalities with cross-modal model merging:} 
    We introduce a novel cross-modal model merging strategy that not only recovers text-only capabilities lost to catastrophic forgetting -- boosting text win-rates by up to 50.2\% but simultaneously enhances multilingual multimodal performance -- improving vision win-rates by up to 20.5\%, demonstrating an efficient, training-free path to stronger models across modalities.
        
    \item \textbf{A comprehensive benchmark suite for real-world multilingual multimodal evaluation:} We introduce \textit{AyaVisionBench}\footnote{\url{https://huggingface.co/datasets/CohereLabs/AyaVisionBench}}, a benchmark spanning 23 languages and 9 vision-language tasks, specifically designed to evaluate generative, open-ended instruction following. 
    To support multilingual evaluation further, we introduce \textit{m-WildVision}\footnote{ \url{https://huggingface.co/datasets/CohereLabs/m-WildVision}}, a high-quality translation of WildVision \citep{lu2024wildvision}. Together, they offer a meaningful and challenging testbed for multimodal models.
\end{enumerate}

\section{A Comprehensive Multilingual Multimodal Data Framework}
\label{sec:synth_data}

To solve for the scarcity of multilingual multimodal instruction data, prior efforts often depend on direct LLM-based translations of English-centric datasets. Approaches such as Pangea \citep{yue2024pangea} and Palo \citep{maaz2024palo} extend coverage across languages either through large-scale translation or multilingual caption alignment. However, these methods still struggle with limited linguistic diversity, the introduction of ``translationese'' from overreliance on translation, strict task formulations, and a lack of conversational naturalness.

\begin{figure}[t]
    \centering
    \makebox[\textwidth][c]{
        \includegraphics[width=\linewidth, trim=3cm 3cm 3cm 3cm, clip]{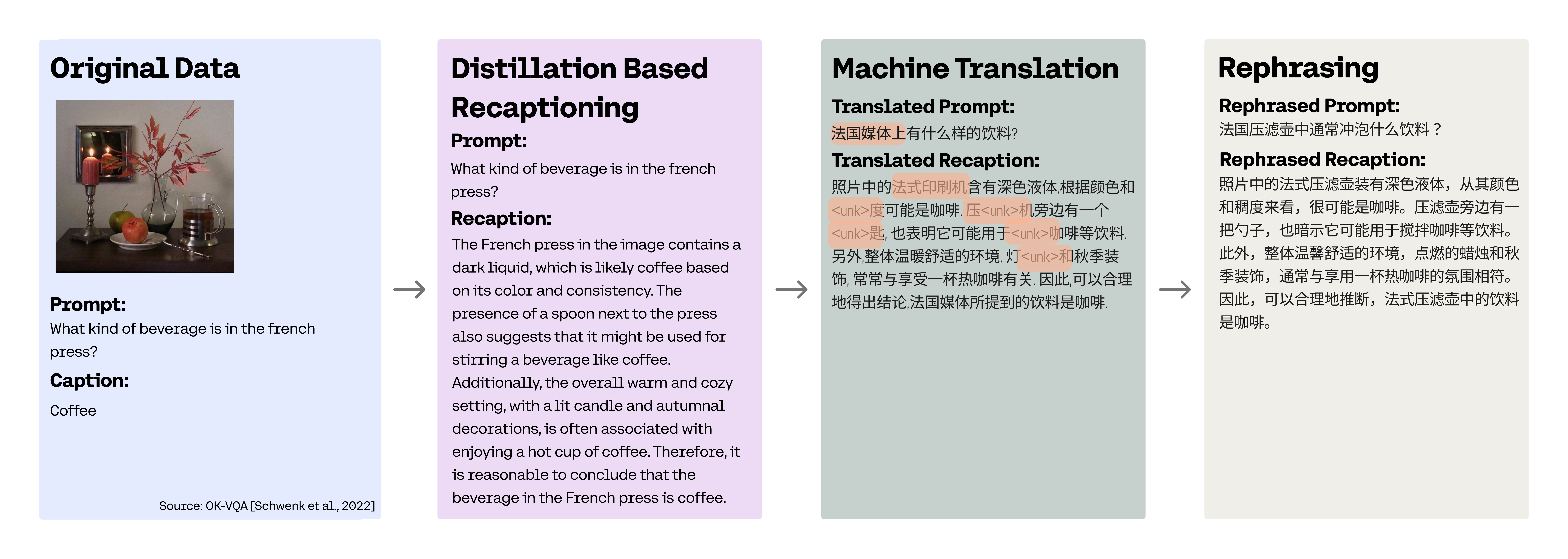}
    }
    \caption{\textbf{Our synthetic annotation pipeline enables diverse, high quality responses for multimodal instructions.} The pipeline consists of three core stages: (1) distillation-based recaptioning, (2) machine translation, and (3) rephrasing. We highlight common machine translation errors, such as unknown tokens (e.g.\ consistency, lit candle) or mistranslations, as in the case of `French press' rendered as `French media' due to lexical ambiguity in the word `press'. Rephrasing helps to resolve such issues, improving both the fluency and semantic accuracy of translations.
    } 
    \label{fig:data-pipeline}
\end{figure}

To address these gaps, we introduce a robust multimodal synthetic re-annotation pipeline for constructing high-quality multilingual multimodal datasets. As illustrated in \Cref{fig:data-pipeline}, our pipeline comprises three core stages: 1) \textit{distillation-based recaptioning} (\S~\ref{sec:synthetic-reannotation}), 2) \textit{dataset filtering} (\S~\ref{sec:filtering}), and 3) \textit{translation combined with multilingual rephrasing} (\S~\ref{sec:translation-and-rephrasing}).
This pipeline significantly enhances the dataset's quality, diversity, and linguistic coverage, resulting in a rich multilingual instruction dataset spanning 23 languages.

\subsection{Data Collection}
\label{sec:data_collection}

We began dataset construction by curating a diverse English-language multimodal instruction-tuning corpus. We constructed our dataset on well-established open-source resources, including \textit{Cauldron}\footnote{\url{https://huggingface.co/datasets/HuggingFaceM4/the_cauldron}}~\citep{laurenccon2024matters}, a large-scale collection of 50 vision-language datasets ($\sim$30M samples), and \textit{PixMo}\footnote{\url{https://huggingface.co/collections/allenai/pixmo-674746ea613028006285687b}}~\citep{deitke2024molmo}, a comprehensive dataset spanning seven multimodal tasks ($\sim$ 6M samples). We also drew from other sources such as \textit{SlideVQA} \citep{tanaka2023slidevqa}, \textit{PDFVQA} \citep{ding2023vqa}, and \textit{ScreenQA} \citep{hsiao2022screenqa}.
Our dataset follows Cauldron’s framework and covers a broad range of multimodal tasks: visual question answering (VQA), captioning, OCR and document understanding, chart and figure analysis, table comprehension, logical reasoning, academic or textbook questions, image comparison, and code generation from screenshots.
\begin{table}[h]
\centering
\caption{Task-wise distribution in our curated dataset, showing the proportion and the number of samples in the $\sim$2.29M collection.} 
\label{tab:task-distribution}
\scalebox{0.83}{
\begin{tabular}{l|ccccccccc}
\toprule
\textbf{Task} & \textbf{VQA} & \textbf{Capt.} & \makecell{\textbf{OCR/} \\ \textbf{Doc}} & \makecell{\textbf{Chart/} \\ \textbf{Fig}} & \makecell{\textbf{Table} \\ \textbf{Compr.}} & \makecell{\textbf{Logic.} \\ \textbf{Reasoning}} & \makecell{\textbf{2 Image} \\ \textbf{Diff.}} & \textbf{Textbook} & \makecell{\textbf{SS to} \\ \textbf{Code}} \\
\midrule
\textbf{Total Samples} & 560K & 220K & 490K & 289K & 222K & 252K & 239K & 20K & 9.5K \\
\textbf{Proportion}  & 24.5\% & 9.6\% & 21.4\% & 12.6\% & 9.2\% & 11.0\% & 10.4\% & 0.9\% & 0.4\% \\
\bottomrule
\end{tabular}
}
\end{table}
As Cauldron performs upstream filtering to remove duplicates across its aggregated sources, the subset we use does not contain repeated samples. Likewise, the PixMo data we incorporate -- primarily within the chart and figure category -- consists of synthetically generated content that is distinct from Cauldron and other sources, ensuring no overlap across datasets.

To ensure robust generalization across task types, we regulated the number of samples per category to construct a balanced and representative dataset. The final collection contains $\sim$2.29M samples, with the task-wise sample counts and distribution detailed in \Cref{tab:task-distribution}. This English data mixture serves as the basis for our further synthetic re-annotation and translation pipeline, forming multilingual instruction tuning set used to train Aya Vision.

\subsection{Distillation-based Recaptioning} 
\label{sec:synthetic-reannotation}

Our goal with recaptioning is to alter the data space such that it better reflects the data distribution we aim to represent in the real-world. To achieve this, we generate synthetic alternatives to the original completions across the $\sim$2.3M data points in our English dataset selection.

The original dataset is primarily composed of open-source, academic image captioning corpora, which exhibit limited linguistic diversity and constrained stylistic variation. Much of the data originates from a narrow set of sources such as MS-COCO \citep{lin2014microsoft}, Visual Genome \citep{krishna2017visual}, and Open Images \citep{kuznetsova2020open}, leading to repetitive content and reduced variation in captions for similar images. Furthermore, these English datasets are typically short and simplistic (average caption length across datasets is just 14.2), and often lack detailed descriptions or a conversational tone expected from state-of-the-art generative models.

Given these limitations, our goal with synthetic re-annotation is to generate recaptions that are more detailed, natural, and diverse in both tone and content. However, a key constraint in this process is that the recaptioned outputs also must remain anchored to the ground-truth answer.

\begin{table}[ht]
\centering
\caption{Examples of task-specific recaptioning outputs for different prompt strategies.}
\label{tab:recaption-examples}
\scalebox{0.87}{
\renewcommand{\arraystretch}{1.1}
\begin{tabular}{p{2.5cm} | p{7.0cm} | p{6.6cm}}
\toprule
\textbf{Task Type} & \textbf{Prompt Instruction (Simplified)} & \textbf{Sample Recaptioned Output} \\
\midrule
\textbf{Captioning} & Rewrite the original caption to be more detailed, descriptive, and human-like. Avoid introducing unseen elements. & A man wearing a red helmet rides a mountain bike along a forest trail, surrounded by tall green trees. \\
\midrule
\multirow{2}{=}{\textbf{Reasoning / Math}} 
& Solve the visual/mathematical problem with a clear, step-by-step explanation. Ensure logical correctness and clarity. & 
\multirow{2}{=}{To find the total, we multiply 4 by 3 because there are 4 rows with 3 items each. 4 $\times$ 3 = 12. So, the final answer is 12.} \\
& The response should be logical, clear, and easy to follow. Include intermediate reasoning steps. & \\
\bottomrule
\end{tabular}
}
\end{table}

The effectiveness of the recaptioning depends heavily on the quality of the prompt templates, which play a critical role in shaping the richness and relevance of the generated annotations \citep{guo2024mammoth, fang2024vila}.
To enhance the quality of our synthetic data, we design task-specific prompt templates for the teacher model, which guide the recaptioning process. These prompt strategies are adapted to rewrite captions based on the ground-truth and to meet the requirements of different vision-language tasks. For example, templates for reasoning tasks are more structured to elicit step-by-step explanations; captioning prompts emphasize more detailed and informative descriptions; and VQA prompts are designed to have accurate and image-grounded answers. 
Table~\ref{tab:recaption-examples} presents examples for two different tasks, illustrating how recaptioning instructions vary by context. For additional task types and full prompt formats, see \Cref{app:recaption-templates}.

Taken together, recaptioning serves to bridge the gap between limited, narrowly scoped training data and  the rich, diverse language found in real-world contexts. To quantify its linguistic impact, we analyze several textual properties --average word count, number of tokens, and lexical diversity -- using the Measure of Textual Lexical Diversity (MTLD) \citep{shen2022lexicalrichness}. Following recaptioning, the average word count increases \textbf{from 14.2 to 100.1}, token count rises \textbf{from 27.2 to 140.8}, and MTLD improves \textbf{from 11.0 to 61.2}. Higher MTLD scores indicate greater vocabulary variation; a score of 61.2 suggests strong lexical richness comparable to fluent language use~\citep{mccarthy2010mtld,ploeger2024tailored}.
These more expressive and natural annotations support better generalization and improved robustness in downstream multimodal tasks. Examples of recaptioned outputs are provided in \Cref{app:recaption_examples}.

\subsection{Verifying and Filtering Recaptioned Instruction Data} 
\label{sec:filtering}

Recaptioning offers a scalable approach to improving the quality of model responses. However, synthetic generations can still introduce errors or hallucinated content that is not grounded in the image \citep{rohrbach2018object,liu2023mitigating,li2023evaluating,gunjal2024detecting}. 
Training on such data may amplify a model's tendency to hallucinate or generate inaccuracies that compromise overall quality. To mitigate these risks and ensure both fluency and correctness, we implement a two-stage filtering pipeline that enhances the overall reliability of the recaptioned dataset. While some methods apply single-pass alignment filtering, e.g CLIP score \citep{Nguyen2023}, or train models to avoid hallucinations using reward learning \citep{ben2023mocha, wang2024mdpo}, our two-stage pipeline adds an extra safeguard against the inclusion of fluent yet hallucinated outputs.

\textbf{Stage 1: Keyword-based filtering.}
We begin with simple keyword detection to identify recaptioned samples that exhibit common failure modes, such as refusals to respond or repeated phrases from the input prompt. To catch these issues, we compile a list of keywords and phrases that automatically flag such responses. Flagged samples are either sent back to the model for regeneration or discarded if the issue persists.

While keyword-matching can detect basic errors, it still struggles to identify more subtle inaccuracies. This limitation is particularly critical for tasks that require deterministic or subjective answer, such as question answering or mathematical reasoning. In these cases, the teacher model may ignore the provided ground truth or hallucinate details, resulting in flawed or incorrect answers.

\textbf{Stage 2: LLM-based semantic filtering.}
To address more nuanced errors, we apply a second-stage filtering using \texttt{command-r-plus-08-2024}\footnote{\url{https://huggingface.co/CohereLabs/c4ai-command-r-plus-08-2024}\label{command-r-plus}} for semantic verification (see \Cref{app:filitering_template_and_samples} for the prompt). In this stage, the original and rephrased captions are presented to the model, which acts as a semantic judge to assess whether the answer to the original caption remains valid given the rephrased version. This ensures that recaption do not alter the underlying meaning or contradict with the ground truth answer.
All corrupted samples identified at this stage are discarded. This step reveals an overall error rate of 3.2\% (62,370 samples) in the recaptioned data. Task complexity significantly influences error frequency -- for example, reasoning tasks exhibit a higher error rate (4.6\%) than simpler VQA tasks (2.5\%). This trend aligns with findings from prior work \citep{yue2024mmmu, wang2024charxiv, song2025visualpuzzles}. By integrating keyword-based filtering with nuanced semantic evaluation capabilities of an LLM, our pipeline generates a recaptioned dataset that is cleaner, more reliable, and better optimized for visual instruction tuning. Examples of filtered samples are provided in \Cref{app:filitering_template_and_samples}.

\subsection{Hybrid Translation Pipeline for Multilingual Instruction Data} \label{sec:translation-and-rephrasing}

Our approach diverges from prior efforts that either rely exclusively on proprietary LLMs for translation \citep{yue2024pangea, maaz2024palo} or highlight disparities in translation quality between high- and low-resource languages without directly addressing how to mitigate them \citep{hendy2023good}. For example, \citet{hendy2023good} find that GPT models perform competitively on high-resource languages but struggle significantly with low-resource ones.
Although machine translation has inherit limitations, it remains essential for broad language coverage, especially in-language human-curated datasets in many languages are scarce and typically reserved for evaluation \citep{singh2024aya,romanou2024include, aakanksha2024multilingual, singh2024global, salazar2025kaleidoscope}. 
Prior work has also shown that translating instruction data can significantly improve cross-lingual generalization in language models \citep{ranaldi2023does,dang2024aya,ermis-etal-2024-one,ustun2024aya}.

However, while machine translation models offer broad coverage, they often introduce artifacts that compromise fluency and fidelity. These include unnatural phrasing, incorrect lexical choices, or incomplete renderings as documented in prior studies \citep{bizzoni2020human, vanmassenhove2021machine, ustun2024aya, singh2024aya}. 
Large language models may struggle with translation, especially in low-resource language contexts \citep{zhu2023multilingual}.
To balance language coverage with translation quality, we adopt a \textbf{hybrid approach}:
\vspace{-2mm}
\begin{itemize}
    \item We begin with machine translation, using the NLLB-3.3B model\footnote{\url{https://huggingface.co/facebook/nllb-200-3.3B}} \citep{costa2022no}. Specifically, we translate our re-annotated English dataset (see §\ref{sec:synthetic-reannotation}) into the following 22 languages: \textit{Arabic, Chinese, Czech, Dutch, French, German, Greek, Hebrew, Hindi, Indonesian, Italian, Japanese, Korean, Persian, Polish, Portuguese, Romanian, Russian, Spanish, Turkish, Ukrainian,} and \textit{Vietnamese}.
    
    \item We then apply a post-editing step using a capable multilingual language model, \texttt{command-r-pl-\\us-08-2024}\footref{command-r-plus}, to refine the translations. This step uses the initial machine-translated output as an in-context example to guide the model toward generating more fluent and accurate outputs \citep{zhu2023multilingual, raunak2023leveraging}. In doing so, we correct common machine translation artifacts while preserving the original semantic content.
\end{itemize}
The prompt used for this rephrasing step and some examples illustrating improvements from rephrasing are in \Cref{app:rephrasing_template_and_enhacements}.

This two-stage process ensures higher translation quality across languages by combining broad coverage from machine translation with fluency improvements from LLM-based post-editing.
To further improve training efficiency and generalization, we do not translate the full English dataset into all 22 languages; instead, we randomly sample subsets of the English pool of examples for each languages. This approach improves efficiency and helps avoid overfitting by reducing repeated exposure to identical content across languages. Prior work has shown that partial translation can achieve strong multilingual generalization while significantly reducing data size \citep{geigle2023mblip,shaham2024pinch}, and is commonly used in large-scale multilingual datasets to enhance linguistic diversity without unnecessary duplication \citep{muennighoff2022crosslingual, nguyen2024diverse,ustun2024aya,dang2024aya,aryabumi2024aya23openweight}.

To evaluate translation quality, we report \textbf{COMET}\footnote{\url{https://github.com/Unbabel/COMET}} \footnote{\url{https://huggingface.co/Unbabel/wmt23-cometkiwi-da-xxl}} \citep{rei2020comet, rei2023scaling}, a reference-free machine translation evaluation metric. The translations from NLLB-3.3B achieve an \textbf{average score of 0.7455} across the 22 languages. After post-editing, the \textbf{average score increases to 0.8293}, indicating the effectiveness of our hybrid strategy. \textbf{COMET scores typically range from 0 to 1}, with higher values indicating better adequacy and fluency. Thus, a gain of over 0.08 reflects a substantial quality improvement. Detailed per-language COMET improvements are reported in \Cref{tab:translated_recaption_comet} in \Cref{app:breakdown_by_language}.

\section{Balancing Performance across Languages, Modalities and Tasks} 
\label{sec:balance}

For multimodal LLMs, carefully sampling the fine-tuning mixture with high-quality and task-oriented visual instructions is crucial for optimal performance \citep{liu2023visual,laurenccon2024matters,tong2024cambrian,dai2024nvlm}. In multilingual multimodal LLMs, this challenge intensifies as the balancing should be optimized for both multilingual and multimodal dimensions. Previous works \citep{ustun2024aya, aryabumi2024aya, dang2024aya} have shown that a skewed distribution of languages in the training mixture hampers the model’s ability to learn reliably, leading to measurable drops in accuracy on a subset of languages. Furthermore, a state-of-the-art multimodal LLM should also retain its text-only capabilities, as these models are often deployed in real-world scenarios that encompass both multimodal and text-only use cases. 

Retaining the text-only performance of the backbone LLM, while acquiring strong multimodal capabilities through multimodal training is challenging for several reasons.
Firstly, choosing the data mixture to strike a balance between multimodal and text datasets is a challenging problem, as finding the right balance is non-trivial and requires a multitude of ablations. 
For instance, Molmo \citep{deitke2024molmo} and Pangea \citep{yue2024pangea} include approximately 10\% text-only data in their multimodal SFT mixture to retain text performance. While this might enable minimal degradation on text-only academic benchmarks, we observe in practice that both models suffer a significant drop in open-ended generation performance measured by the preference evaluation as shown in \Cref{fig:text-degradation}.

Secondly, reintroducing previously seen text-only data can potentially lead to overfitting with minimal improvement in text performance and a higher degradation in multimodal performance~\citep{marafioti2025smolvlm}. We further investigate this pattern through ablations, presented in detail in \Cref{sec:model-merging-more-effective}. Moreover, post-training of state-of-the-art LLMs typically involves several steps of SFT and preference optimization \citep{dang2024aya, lambert2024t, cohere2025commandaenterprisereadylarge}, which could cause instability due to the shift in the data distribution in the multimodal fine-tuning step. This highlights the importance of striking a balance during multimodal fine-tuning to maintain model robustness and generalization. 

In this work, we explore a variety of mitigation to this solution including (1) systematic weighting of different sources of data to preserve both language balancing and diversity, (2) Cross-modal model merging to seamlessly integrate multimodal and text-only capabilities.

\begin{figure}[t]
\centering
\definecolor{Original}{HTML}{66c2a5}
\definecolor{Multilingual}{HTML}{fc8d62}
\definecolor{Synthetic}{HTML}{8da0cb}
\begin{minipage}[t]{0.51\textwidth}
\vspace*{0pt}  
\centering
\footnotesize
\renewcommand{\arraystretch}{1.1}
\scalebox{0.87}{
\begin{NiceTabular}{l>{\columncolor{Original!80}}l>{\columncolor{Multilingual!80}}l>{\columncolor{Synthetic!80}}ll|l}
\toprule
\textbf{Task} & 
\rotatebox{90}{\textbf{Orig.}} & 
\rotatebox{90}{\textbf{Multi.}} & 
\rotatebox{90}{\textbf{Synth.}} & 
\rotatebox{90}{\textbf{Total}} & 
\rotatebox{90}{\textbf{Per(\%)}} \\
\midrule
General VQA              &  269.0k &  311.2k &  168.2k &   748.4k  &  27.2 \\
Captioning               &   ~~~-  &   74.6k &  109.0k &   183.6k  &  6.7 \\
OCR                      &  231.8k &   60.7k &  188.8k &   481.3k  &  17.5 \\
Figures/Charts           &  290.0k &   31.3k &  159.6k &   480.9k  &  17.5 \\
Table Compr.             &   77.5k &  260.7k &   56.5k &   394.7k  &  14.4 \\
Reason./Logic/Math       &   ~~~- &  136.4k  &   60.9k &   197.2k  &  7.2 \\
Multi Image              &   39.6k &   78.0k &   97.3k &   214.8k  &  7.8 \\
Textbook/Academic        &   19.1k &   ~~~-  &   12.8k &   31.9k   &  1.2 \\
Screenshot → Code        &    9.5k &    5.2k &   ~~~-  &   14.7k   &  0.5 \\
\hdashline
\noalign{\smallskip}
\textbf{Total}           &  936.3k &  958.1k &  853.0k & \textbf{2.75M} & 100\% \\
\bottomrule
\end{NiceTabular}}
\end{minipage}
\hfill
\begin{minipage}[t]{0.47\textwidth}
\vspace*{8pt}  
\centering
\includegraphics[width=0.92\linewidth, trim=189 200 120 250, clip]{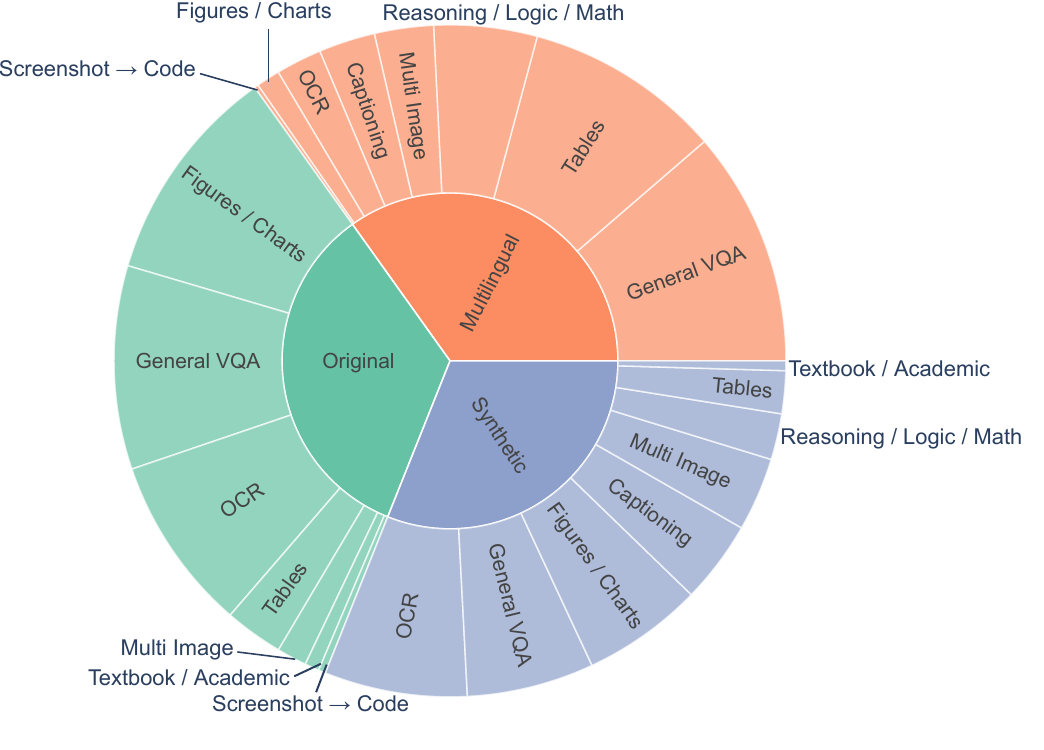}
\end{minipage}

\vspace{0.5em}
\caption{\textbf{Overview of our multilingual multimodal SFT mixture from various task categories.} Left: Number of samples across data sources and tasks categories used in training. Right: Visual breakdown of dataset source distributions.}
\label{fig:dataset-summary-sidebyside}
\end{figure}

\subsection{Sampling Visual Instructions from Multiple Sources and Languages}

To balance coverage and preserve diversity, we mix and weight three buckets of data: 
\begin{enumerate}
    \item \textbf{Synthetically Re-annotated data in English:} This data was generated after the first phase of our data framework (\S~\ref{sec:synthetic-reannotation}), $2.29\text{M}$ samples in total. Inside this bucket, we upsample datasets with a small number of samples, such as science or textbook questions, to avoid underrepresenting any task categories. Additionally, we also upsample datasets deemed to be of higher quality upon manual inspection leading to a total of $3.5\text{M}$ samples from this bucket being seen by the model.
    \item \textbf{Multilingual datasets:} This data was generated by using a subset of re-annotated English dataset through our data framework (\S~\ref{sec:translation-and-rephrasing}). We uniformly sample data across 22 languages (except English) and maintain a similar task distribution to the first bucket. While the total data volume in this bucket amounts to a total of $5\text{M}$ samples, we sample $3.4\text{M}$ uniformly distributed across 22 languages (except English) to preserve the balance between tasks.
    \item \textbf{High-quality original datasets:} In addition to the fully synthetic data, we also use a selection of original datasets, based on their quality. This bucket is required since some downstream VQA evaluations expect syntactically accurate answers that match their training distribution and penalize semantically correct generations (for example $0.5$ instead of $1/2$). However, we downsample the original corpus to avoid a drop in overall quality, as this data penalizes natural generations and completion length -- thereby degrading the model's free-form conversational abilities. While the total number of samples in this bucket is $6\text{M}$, we sample $3.7\text{M}$ for training.
\end{enumerate}

In each data bucket, we ensure a diverse set of tasks is represented. To enhance multilingual performance, we experiment with varying proportions of multilingual data -- these results are presented in \S~\ref{sec:multilingual-percentage}. Based on these findings, we use approximately 66\% of synthetically re-annotated datasets out of which 35\% corresponds to the multilingual datasets; while the remaining 34\% are the high-quality original datasets. \Cref{fig:dataset-summary-sidebyside} illustrates the composition of the training data across buckets and tasks, totaling $2.75\text{M}$ sequence-packed final training samples.

\subsection{Unifying Multimodal Performance with State-of-the-Art Text Capabilities}

\begin{wrapfigure}{r}{0.5\textwidth}
    \vspace{-0.5cm}
  \centering
  \includegraphics[width=0.5\textwidth]{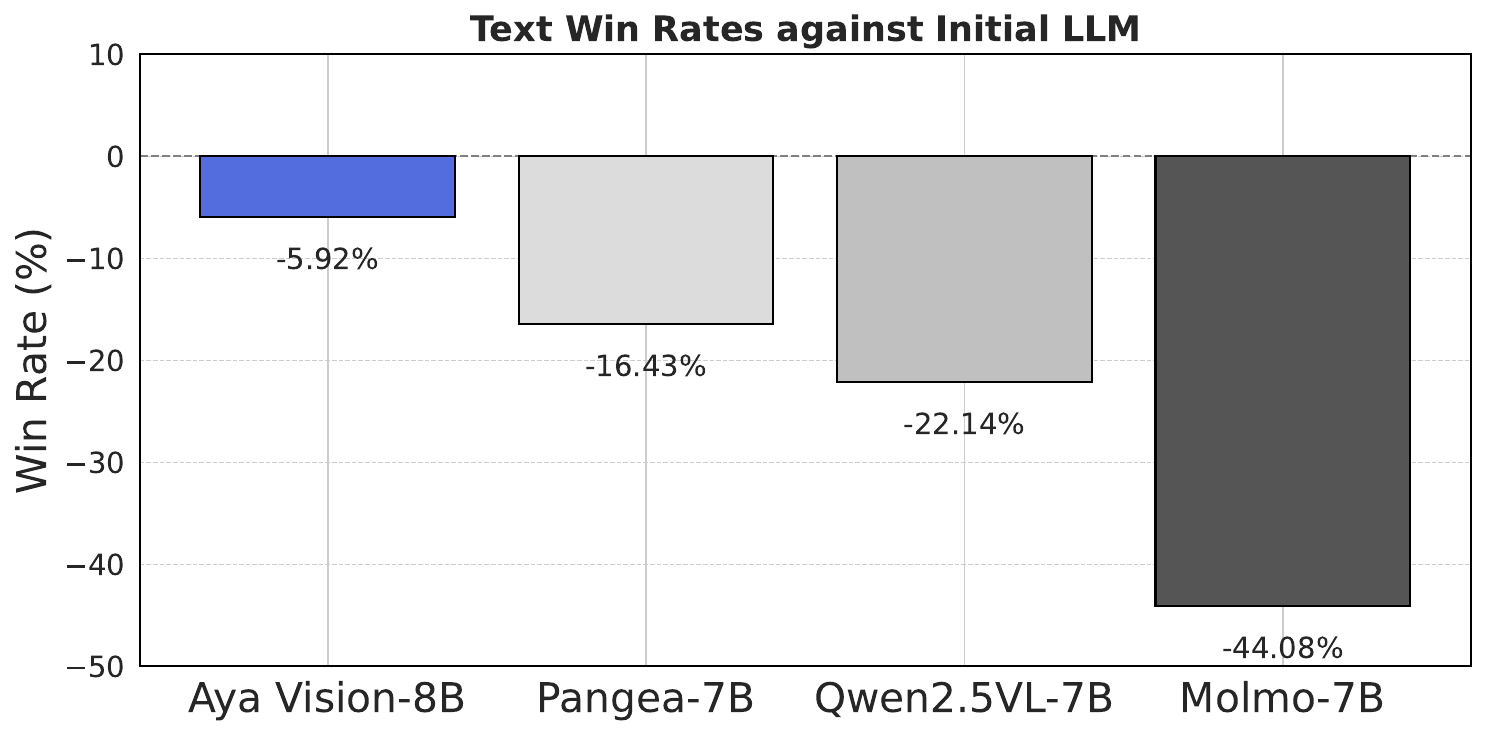}
  \caption{\textbf{Degradation in text-only win-rates after multimodal training.} Each model is compared to their initial LLM on mArenaHard \citep{dang2024aya}. We see that only including a percentage of text-only data in the final multimodal training mix is insufficient to retain open-ended generative performance.}
  \label{fig:text-degradation}
  \vspace{-0.2cm}
\end{wrapfigure}

In Aya Vision, instead of balancing multimodal and text-only abilities in the data space via a sweep over data mixtures, we introduce a novel cross-modal model merging inspired by the recent body of work in model merging \citep{ wortsman2022model, matena2022merging, yadav2023ties,aakanksha2024mixdatamergemodels, goddard2024arcee}. Concretely,  we posit that since the multimodal model is initialized from the final preference-tuned LLM checkpoint, sharing a part of the optimization trajectory \citep{izmailov2018averaging, frankle2020linear, ilharco2022editing} makes the multimodal LLM and the backbone LLM amenable to merging. Cross-modal model merging introduces an efficient, training-free recovery solution for retaining text-only performance by balancing multimodal and text-only capabilities in the weight space \textit{aposteriori}. We conduct systematic study of merging techniques applied to the weights of the original text-only LLM and the LLM backbone of the multimodal model (see \S~\ref{sec:merging-ablations}).

We perform a linear interpolation between the text-only LLM and the backbone LLM of the multimodal model as the merging method, as shown in Equation \ref{eq:merging}. Since the text-only language model lacks the vision encoder and alignment layer, we simply inherit them from the vision-language model.
 \begin{equation}
    W_{\text{merged}} =  \alpha.W_{\text{mm-LLM}} + (1-\alpha).W_{\text{text-LLM}} \;\;\;\;\;\;\;\;\;  
    \label{eq:merging}
\end{equation}

\section{Aya Vision's Architecture and Training Details}

\subsection{Architecture}
Aya Vision models follow the common architecture design for vision-language models \citep{liu2023visual, liu2024improved, laurenccon2024matters, mckinzie2024mm1, chen2024far, deitke2024molmo} that is based on late-fusion \citep{team2024chameleon} of (1) a vision encoder to compute image patch embeddings which is pre-trained on billions of image-text pairs \citep{radford2021learning, zhai2023sigmoid, chen2024far, tschannen2025siglip}, (2) a connector that maps the embeddings from the output space of the vision encoder to the input embedding space of the language model, (3) a large language model.

\label{sec:architecture}
\textbf{Vision Encoder:}
We use \texttt{siglip2-so400m} \citep{tschannen2025siglip} as the initialization for the vision encoder, which has been pretrained with an auto-regressive decoder-based loss in addition to the original sigmoidal loss \citep{zhai2023sigmoid}. This primes the vision encoder to generate high-quality dense feature representations for generative tasks, making it the perfect candidate for a multilingual vision language model. Specifically, we use \texttt{siglip2-so400m-patch14-384}\footnote{\url{https://huggingface.co/google/siglip2-so400m-patch14-384}} in Aya-Vision-8B for a reduced activation footprint, making it widely accessible on cheaper hardware. For Aya-Vision-32B, we opt for the higher resolution \texttt{siglip2-so400m-patch16-512}\footnote{\url{https://huggingface.co/google/siglip2-so400m-patch16-512}} to achieve better performance\citep{laurenccon2024matters}.

\textbf{Image Processing:}
The performance of multimodal LLMs improves with higher input resolution \citep{mckinzie2024mm1, laurenccon2024matters}, however, most vision encoders are pretrained on a fixed resolution. To enable Aya Vision models to process images with arbitrary resolutions, similar to \cite{chen2024far}, we map the input images to the nearest supported resolution that minimizes distortion in the aspect ratio.
After resizing, we split the image into up to 12 non-overlapping tiles based on the image encoder's resolution to be processed independently by the vision encoder. In addition to tiles, we include a thumbnail (resized) for a low-resolution overview of the image.

\textbf{Vision-Language Connector:}
Following the image encoder, the vision-language connector maps features from the vision encoder to the language model's input embedding space. We use a 2-layer MLP with SwiGLU activation function \citep{shazeer2020glu}. To reduce the number of image tokens passed to the language model, we perform Pixel Shuffle \citep{chen2024far}, which downsamples the image tokens in the spatial dimensions by stacking $2\times2$ patch embeddings along the embedding dimension before passing through the connector layer. This decreases the number of image tokens by $4\times$, resulting in a maximum of 2,197 and 3,328 image tokens for our 8B and 32B models respectively.
When passing image tokens to LLM, we use special delimitation tokens to denote the start and the end of image token sequences. Additionally, we inject 1D-tile tags \citep{dai2024nvlm} to denote image tiles as a form of explicit positional encoding for the tiles. We use regular text tokens (\texttt{TILE_1,...,TILE_N} and \texttt{TILE_GLOBAL} for thumbnail) for potential inference-time scaling.

\textbf{Language Model:}  
Although some previous works initialize the language model from a pre-trained base checkpoint \citep{beyer2024paligemma}, we initialize the language model from a multilingually post-trained LLM to inherit strong capabilities in various tasks including chat, instruction-following, and multilingual. For Aya-Vision-8B, we use an LLM based on Command-R7B\footnote{\url{https://huggingface.co/CohereLabs/c4ai-command-r7b-12-2024}} which is further post-trained with the Aya Expanse recipe \citep{dang2024aya}, and for Aya-Vision-32B, we use the Aya-Expanse-32B \citep{dang2024aya}.

\subsection{Multimodal Training}
Following previous work that use late-fusion as in our models \citep{liu2023visual, liu2024improved, laurenccon2024matters, mckinzie2024mm1, chen2024far, deitke2024molmo}, we train Aya Vision models in two steps: (1) Vision-Language Alignment and (2) Supervised Fine-tuning.

\textbf{Vision-Language Alignment:}
In this step, we only train the vision-language connector by keeping both the vision encoder and the language model frozen. Freezing the language model and vision encoder allows for using a high learning rate to quickly map the image features to the input embedding space. We use a peak learning rate of $10^{-4}$  and $10^{-3}$ for Aya-Vision-8B and 32B models respectively. Additionally, we find that the 32B model requires longer training in this step due to the much larger connector size. While Aya-Vision-8B includes a $190\text{M}$ vision-language connector, the parameter size of the connector in 32B model is $428\text{M}$. Therefore, we train the 8B model for $9.7\text{k}$ steps (1 epoch) and the 32B model for $19\text{k}$ steps (2 epochs). Similar to previous works \citep{liu2023visual, yue2024pangea} we use LLaVa-Pretrain\footnote{\url{https://huggingface.co/datasets/liuhaotian/LLaVA-CC3M-Pretrain-595K}} as the primary source of data in this step. However, since this data is English-only, we add a small fraction of the multilingual data generated by our data framework amounting to $14\%$ of the total data seen during this step. All training details can be found in \Cref{tab:train-hyperparams} in the appendix.

\textbf{Visual Instruction Fine-tuning:}
In the instruction fine-tuning step (i.e., supervised fine-tuning with visual instructions), we train both the vision-language connector and the language model but keep the vision encoder frozen. We experiment with both full model fine-tuning and LoRA \citep{hu2022lora}. For both Aya-Vision-8B and Aya-Vision-32B, we use a batch size of 128 and train for $31\text{k}$ iterations with $\mu\text{P}$ enabled on about $10\text{M}$ samples. The peak learning rates are set to $10^{-4}$  and $5\times10^{-4}$ respectively established via hyperparameter tuning. We utilize sequence packing to pack multiple samples into a single sequence of length 8192 for improved training efficiency. A breakdown of the SFT training data can be found in \Cref{fig:dataset-summary-sidebyside} with detailed discussion presented in \S~\ref{sec:balance}.

\definecolor{highlightcolor}{RGB}{211, 227, 252}

\begin{table}[t]
    \small
    \centering
    \renewcommand{\arraystretch}{1.3}
    \setlength{\tabcolsep}{8pt} 
    \begin{NiceTabular}{lllc}[code-before=\rowcolors{3}{gray!15!white}{white}]
        \toprule
        \textbf{Dataset} & \textbf{Task} & \textbf{Metric} & \textbf{\# Languages} \\
        \midrule
        \RowStyle[rowcolor=highlightcolor]{\bfseries}Multimodal Academic Bench.\\
        xMMMU~\citep{yue2024pangea} & Multimodal Understanding & Accuracy & 7 \\
        MaXM~\citep{changpinyo2022maxm} & VQA & Accuracy & 7 \\
        CVQA~\citep{romero2024cvqa} & VQA & Accuracy & 31 \\
        MTVQA \citep{singh2019towards} & VQA & VQA Score & 9 \\
        Kaleidoscope \citep{salazar2025kaleidoscope} & VQA & Accuracy & 18 \\
        \midrule
        \RowStyle[rowcolor=highlightcolor]{\bfseries}Multimodal Open-Ended Bench.\\
        AyaVisionBench & Multimodal Chat & Win-Rates & 23 \\
        m-WildVision \citep{lu2024wildvision} & Multimodal Chat & Win-Rates & 23 \\
        xChat \citep{yue2024pangea} & Multimodal Chat & LLM-Score & 7 \\
        \midrule
        \RowStyle[rowcolor=highlightcolor]{\bfseries} Text-only Bench.\\
        m-ArenaHard \citep{dang2024aya} & Open-Ended Generations & Win-Rates & 23 \\
        MGSM \citep{shi2022language} & Math. Reasoning & Accuracy & 6 \\
        Global MMLU-Lite \citep{singh2024global} & Language Understanding & Accuracy &  15 \\
        FLORES \citep{guzmán2019flores} & Language Understanding & SpBLEU &  23 \\
        IFEval \citep{zhou2023instruction} & Instruction Following & Accuracy & 1 \\
        \bottomrule
    \end{NiceTabular}
    \caption{\textbf{Multilingual multimodal evaluation suite used in Aya Vision.} Our evaluation suite consists of multilingual multimodal benchmarks, multimodal open-ended benchmarks for preference evaluation, and finally, text-only benchmarks include open-ended, generative, and discriminative evaluation sets.}
    \label{tab:eval_suite}
\end{table}

\section{Evaluation}

\subsection{Multilingual Multimodal Preference Evaluation}
\label{sec:aya_vision_bench}

\subsubsection{Open-ended Multimodal Evaluation}
\label{sec:open-ended}

While recent efforts have explored multilingual evaluation for multimodal LLMs \citep{changpinyo2022maxm, romero2024cvqa, tang2024mtvqa, yue2024pangea}, existing benchmarks still fall short of enabling robust, real-world evaluation. Most current suites focus on static, single-turn tasks with predefined answers, failing to capture the nuanced, open-ended, and dynamic nature of real-world user interactions.
To address this, we introduce:

\noindent\textbf{AyaVisionBench} \footnote{\url{https://huggingface.co/datasets/CohereLabs/AyaVisionBench}}, a benchmark explicitly designed to evaluate not only multimodal understanding and reasoning but also generation quality along human-centric dimensions, such as relevance, fluency, and engagement. AyaVisionBench targets the question: \textit{How well can a multimodal model respond to complex, open-ended instructions across languages and modalities?}

AyaVisionBench spans 23 languages and comprises 135 image-question pairs per language, covering 9 diverse task categories: captioning, chart and figure understanding, identifying differences between two images, general visual question answering, OCR, document understanding, text transcription, mathematical or logical reasoning, textbook questions and converting screenshots to code. This multilingual, multi-task design supports comprehensive evaluation of cross-lingual multimodal understanding. Most samples include ground-truth responses for reference. Further construction details are available in \Cref{app:ayavisionbench}. The benchmark is publicly released for community use and broader evaluation.

\textbf{Multilingual WildVision (m-WildVision) and xChatBench} To complement AyaVisionBench, we release \textbf{m-WildVision}\footnote{\url{https://huggingface.co/datasets/CohereLabs/m-WildVision}}, a multilingual extension of WildVision-Bench \citep{lu2024wildvision}, featuring translated prompts in 22 languages. WildVision is curated from real-world user interactions and provides practical, context-rich evaluation scenarios. We also incorporate \textbf{xChatBench} \citep{yue2024pangea}, which supports fine-grained, score-based assessments across 7 languages and various interaction types.

To evaluate model performance across all three benchmarks, we follow the VLM-as-a-judge protocol used in prior multilingual studies \citep{ustun2024aya, dang2024aya}, conducting pairwise comparisons between Aya Vision and baseline models. For scoring and preference ranking, we use \textbf{claude-3-7-sonnet-20250219} \citep{claude3.7systemcard} as the multimodal judge. 
This choice is based on a comparative study using the translated Multimodal RewardBench \citep{yasunaga2025multimodal} across 8 languages,\footnote{English (original), Arabic, Farsi, French, Hindi, Portuguese, Turkish, Vietnamese, Simplified Chinese.} where Claude-3-7-Sonnet outperformed GPT-4o \citep{openai2024gpt4o} and Gemini-2.0-Flash \citep{team2024gemini} by 6.4\% and 25.8\% respectively in preference ranking accuracy. Full details on the evaluation prompt are provided in \Cref{app:judge_prompt}.

\subsubsection{Academic Multilingual Multimodal Benchmarks} 

In addition to the preference-based open-ended multimodal evaluation, we evaluate Aya Vision on visual question answering and reasoning style benchmarks that require the generations to adhere to a prescribed format, such as multiple-choice style or short-form answers, for easy automated evaluation. Specifically, we use \textbf{xMMMU} \citep{yue2024pangea}, \textbf{MaXM} \citep{changpinyo2022maxm}, \textbf{CVQA} \citep{romero2024cvqa}, \textbf{MTVQA} \citep{tang2024mtvqa} and \textbf{Kaleidoscope} \citep{salazar2025kaleidoscope}. These benchmarks, covering a range of languages, measure multimodal understanding, knowledge, and reasoning capabilities of multimodal LLMs. The number of languages in each dataset is shown in \Cref{tab:eval_suite}, and details of these benchmarks are given in \Cref{app:eval-details}.
       
\subsection{Multilingual Text-Only Evaluations}

As a final component of our multilingual evaluation suite, we evaluate Aya Vision models and baselines on various text-only benchmarks. This is important to reflect real-world deployment scenarios where models are used with both multimodal and text-only inputs. However, as shown in \S~\ref{sec:balance}, many vision-language models experience some degree of degradation in their text-only performance. Therefore, to evaluate models' performance in various tasks, we include a set of representative text-only evaluations. 

\textbf{Open-ended evaluation} Similar to AyaVisionBench, we use \textbf{m-ArenaHard} \citep{li2024crowdsourced,dang2024aya} to evaluate and compare models' performance in open-ended text generations in 23 languages.\footnote{We use \texttt{gpt-4o-2024-11-20} \citep{openai2024gpt4o} as the LLM-judge following \citet{dang2024aya}.}

\textbf{Task-specific benchmarks} Additionally, we included \textbf{MGSM} \citep{shi2022language}, \textbf{Global MMLU-Lite} \citep{singh2024global}, and \textbf{FLORES} \citep{guzmán2019flores} covering mathematical reasoning, multilingual language understanding, and machine translation, respectively. Each of these benchmarks includes a different set of languages, as listed in \Cref{tab:eval_suite}. For FLORES, we evaluate models' translation performance from English to the target language (\texttt{En$\rightarrow$X}) as translating from English is a harder task and a good indication for multilingual performance. Finally, we also include \textbf{IFEval} \citep{zhou2023instruction}, although it is English-only, as it measures instruction-following capabilities of models, which potentially impacts the performance in other multimodal and text-only benchmarks. Metrics for these benchmarks are given in \Cref{tab:eval_suite}, and additional details can be found in \Cref{app:eval-details}. 

\begin{figure}[t]
    \centering
    \includegraphics[width=1\linewidth]{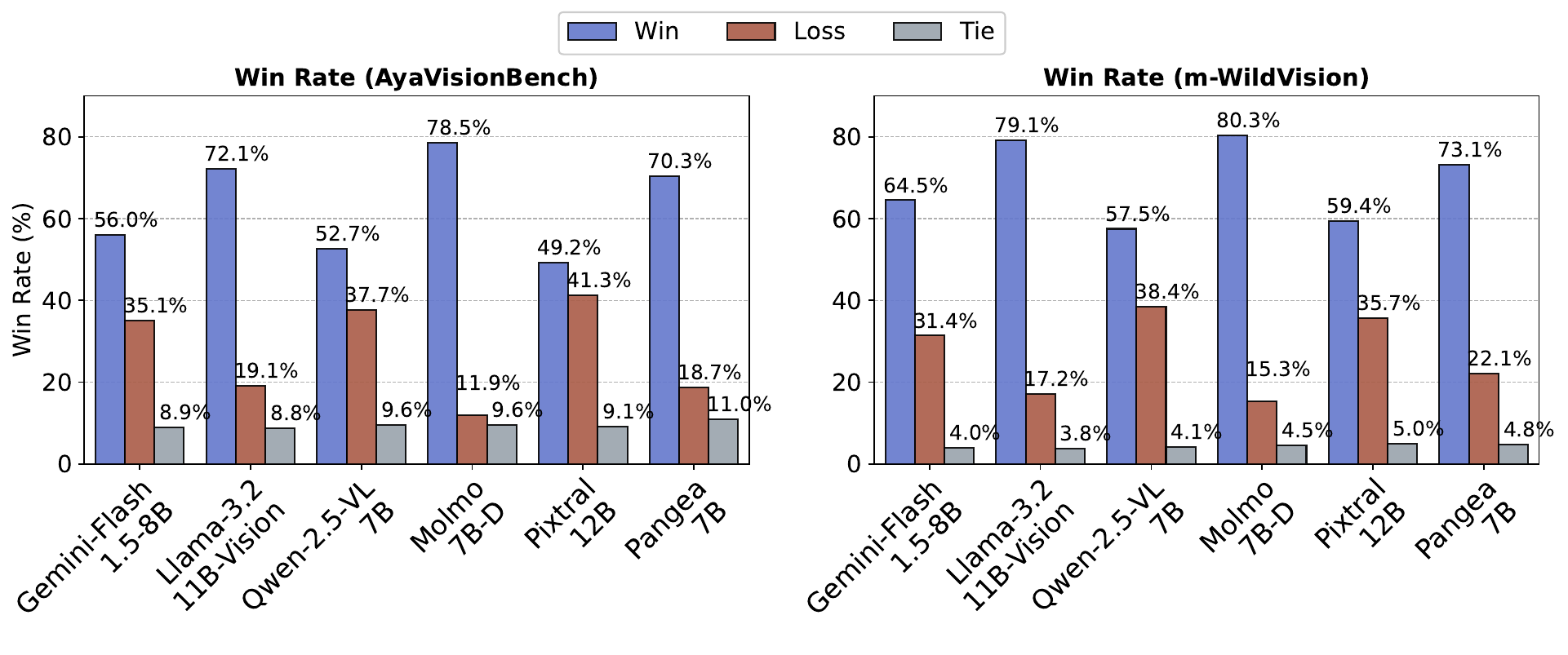}
    \caption{\textbf{Aya-Vision-8B achieves best-in-class performance on preference evaluation.} Pair-wise win-rates on AyaVisionBench and m-WildVision \citep{lu2024wildvision} averaged across 23 languages. We compare Aya-Vision-8B with Gemini-Flash-8B, Llama-3.2-11B-Vision, Qwen-2.5--VL-7B, Pixtral-12B and Pangea-7B on AyaVisionBench (left) and m-WildVision (right). Language-specific breakdown for the results can be found in \Cref{tab:aya_vision_bench-aya_vision_8b} \& \Cref{tab:wildvision-aya_vision_8b} in the Appendix.}
    \label{fig:mm-win-rates-8b}
\end{figure}

\begin{figure}[t]
    \centering
    \includegraphics[width=0.8\linewidth]{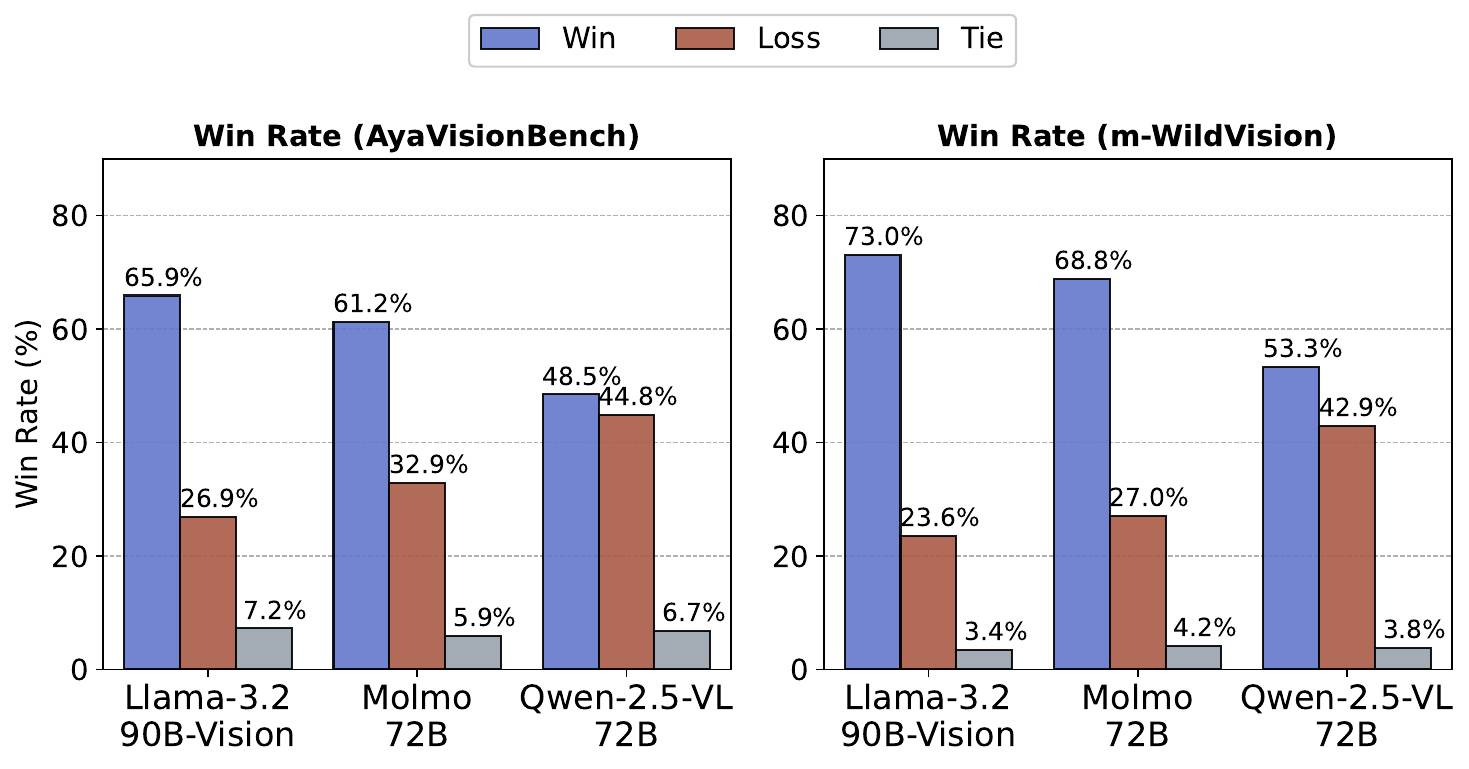}
    \caption{\textbf{Aya-Vision-32B outperforms models more than double its size.} Pairwise win-rates on AyaVisionBench and m-WildVision \citep{lu2024wildvision} averaged across 23 languages. We compare Aya-Vision-32B with Llama-3.2-90B-Vision, Molmo-72B and Qwen-2.5-VL-72B on AyaVisionBench (left) and m-WildVision (right). Language-specific breakdown for the results can be found in \Cref{tab:aya-vision-bench-aya_vision_32b} \& \Cref{tab:mwildvision-aya_vision_32b} 
 in the Appendix.}
    \label{fig:mm-win-rates-32b}
\end{figure}

\definecolor{highlightcolor}{RGB}{211, 227, 252}
\definecolor{textgray}{gray}{0.5}

\begin{table}[H]
    \small
    \centering
    \renewcommand{\arraystretch}{1.3}
    \begin{tabular}{lccccccc}
        \toprule
        Models / Evaluations & \textbf{MaxM} & \textbf{xMMMU} & \textbf{CVQA} & \textbf{MTVQA} & \textbf{Kaleidoscope} & \textbf{xChat} & \textbf{avg} \\
        \midrule
        Pangea-7B & 51.27 & \underline{44.00} &  60.53 & 18.32 & 29.46 & 32.21 & 39.30 \\
        Molmo-7B-D & 44.16 & 37.87 & 58.53 &  16.89 & 36.42 & 23.36 & 36.21 \\
        Llama-3.2-11B-Vision & 39.30 & 42.73 & 58.92 & 16.40 & 36.50 & 28.59 & 37.07 \\
        Pixtral-12B  & 44.43 & 42.27 &  \underline{63.54} & \underline{19.81} & 36.08  & \textbf{64.50} & 45.11 \\ 
        Qwen-2.5-VL-7B & \underline{52.65} & \textbf{46.77} & \textbf{73.22} & \textbf{29.57} &  \textbf{39.64} & 58.14 & \textbf{50.00}  \\
        \rowcolor{highlightcolor} Aya-Vision-8B & \textbf{58.21} & 39.94 & 61.86 &  19.33 &   \underline{38.62} & \underline{58.64} & \underline{46.16} \\
        \midrule
        Molmo-72B & 55.62 & 51.53 & 72.77 & 18.66 & 50.34 & 45.43 & 49.06 \\ 
        Llama-3.2-90B-Vision & \textbf{64.17} & \underline{52.40}& \underline{81.88} & \underline{27.44} & \underline{48.41} & 51.12 & \underline{54.24}\\
        Qwen-2.5-VL-72B & 56.42 & \textbf{61.74} & \textbf{82.10} & \textbf{31.92}  & \textbf{55.02} & \textbf{71.13}  & \textbf{59.72} \\
        \rowcolor{highlightcolor} Aya-Vision-32B & \underline{62.28} & 45.11 &  74.06 & 23.46 & 41.73 & \underline{70.07} & 52.81 \\
        \bottomrule
    \end{tabular}
    \caption{\textbf{Evaluation on multilingual multimodal benchmarks for Aya-Vision-8B and Aya-Vision-32B together with the baselines}. For each benchmark, we include languages that are in the list of Aya Vision's 23 languages.
    The full results on all available languages are given in \Cref{app:breakdown_by_language}.}
    \label{tab:academic-evals}
\end{table}

\subsection{Baselines}
We compare Aya Vision models against a range of state-of-the-art multimodal LLMs, both open- and closed-weight, to evaluate multilingual, multimodal, and text-only capabilities. We select models based on architecture, model size, base model family, and language coverage. The selected models cover a range of sizes (7B to 90B), base models (Llama-3.2, Qwen-2.5, Molmo), and language coverage (including both English and multilingual models).
Our evaluation includes open-weight models (Pixtral \citep{agrawal2024pixtral}, Molmo \citep{deitke2024molmo}, Qwen-2.5-VL \citep{bai2025qwen25vl} and Pangea \citep{yue2024pangea}) as well as the closed-weight (Gemini-Flash-1.5 \citep{geminiteam2024gemini15unlockingmultimodal}). For model families, Qwen, Molmo, and Llama, we  report results across multiple sizes ranging from 7B to 90B parameters.

Among the baseline models, Pangea, Qwen, Pixtral, Llama, and Gemini explicitly report multilingual support. We also include Molmo, which does not explicitly claim to support multiple languages, however in practice, they are heavily used by multilingual users relative to some multilingual models like Pangea-7B \citep{yue2024pangea}. Hence, we think it is important to include. Furthermore, we also find that these models achieve considerable performance in many multilingual tasks, as shown in our evaluation.

\begin{figure}[h!]
    \centering
    \includegraphics[width=\linewidth]{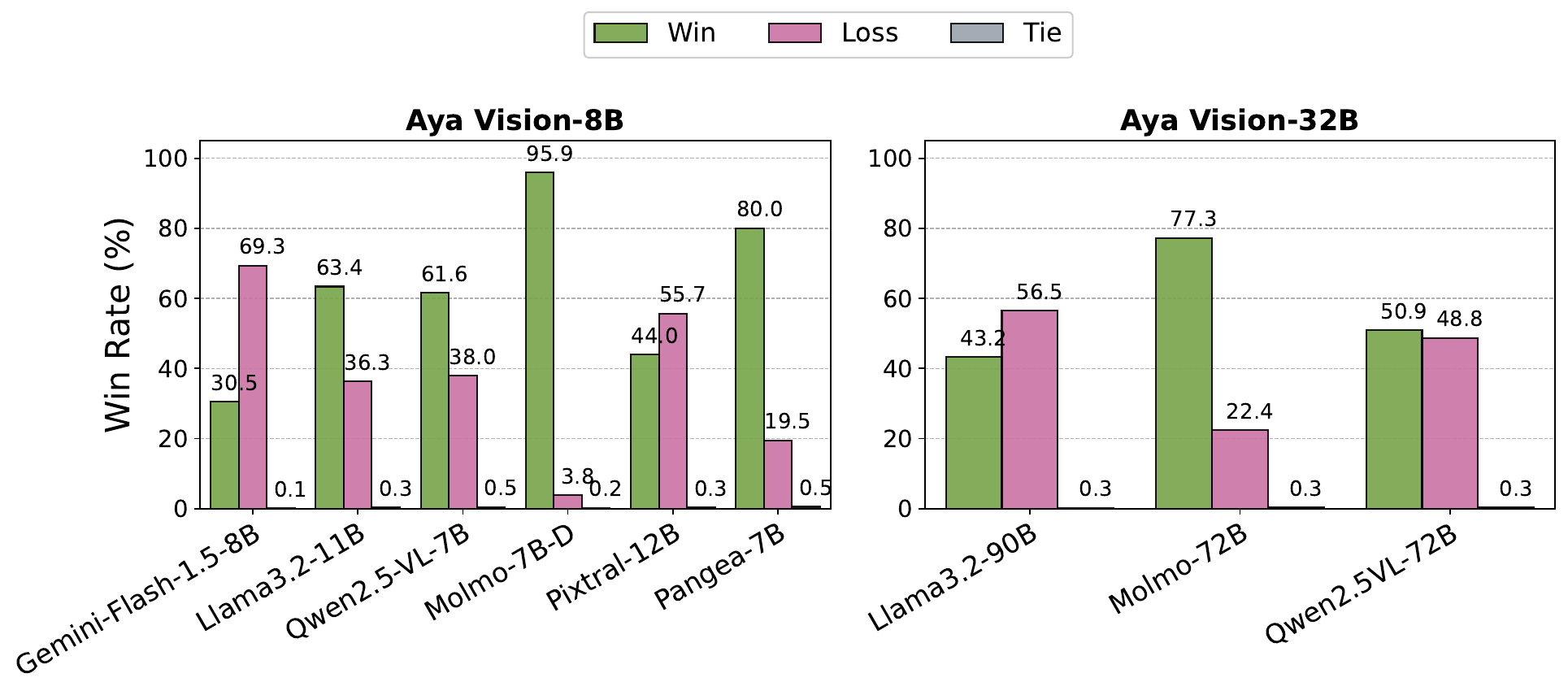}
  \caption{\textbf{Aya Vision models are amongst the best models in text-only preference evaluation compared to models with much larger size.} Pairwise win-rates for Aya-Vision-8B (left) and 32B (right) on m-ArenaHard \citep{li2024crowdsourced,dang2024aya} averaged across 23 languages. Language-specific breakdown for the results can be found in \Cref{tab:marena-aya_vision_8b} \& \Cref{tab:marena-Aya_Vision-32B} in the Appendix.}
  \label{fig:m-arena-combined}
\end{figure}
\definecolor{highlightcolor}{RGB}{211, 227, 252}
\definecolor{textgray}{gray}{0.5}

\begin{table}[H]
    \small
    \centering
    \renewcommand{\arraystretch}{1.3}
    \begin{tabular}{lccccc}
        \toprule
        Models / Evaluations & \textbf{G‑MMLU (Lite)} & \textbf{MGSM} & \textbf{FLORES} & \textbf{IFEval} & \textbf{avg} \\
        \midrule
        Pangea‑7B             & 49.35 & 50.51      &  28.04      & 23.99 & 37.97    \\
        Molmo‑7B‑D            & 39.63 & 49.94  & 15.74  & 56.10  & 40.35 \\
        Llama‑3.2‑11B-Vision         & 60.75  & 72.84  & \underline{31.84} & \textbf{83.43} & \underline{62.22}  \\
        Pixtral‑12B           & \textbf{66.09}  & \textbf{77.62} & 29.29  & 65.59  & 59.65 \\
        Qwen‑2.5‑VL‑7B        & \underline{64.82}  & 60.90  & 27.98  & 72.46 & 56.54 \\
        \rowcolor{highlightcolor} Aya-Vision‑8B   & 62.52  & \underline{76.42}  & \textbf{35.90}  & \underline{82.78} & \textbf{64.41}  \\
        \midrule
        Molmo‑72B             & 71.02 & \underline{86.00} & 32.52  & 78.10 & 66.91 \\
        Llama‑3.2‑90B-Vision  & \underline{77.46}  & 66.67  & \textbf{38.25}  & \underline{88.14} & \underline{67.63} \\
        Qwen‑2.5‑VL‑72B       & \textbf{81.49}  & \textbf{89.61}  & 35.71  & \textbf{89.74} & \textbf{74.14}  \\
        \rowcolor{highlightcolor} Aya-Vision‑32B  & 63.58  & 79.46  & \underline{37.79}  & 78.50 & 64.83 \\
        \bottomrule
    \end{tabular}
    \label{tab:text-only-academic-evals}
    \caption{\textbf{Evaluation on multilingual text-only academic benchmarks for Aya-Vision-8B and Aya-Vision-32B together with the baselines}. For each benchmark, we include languages that are in the list of Aya Vision's 23 languages.
    The full results on all languages are available in \Cref{app:breakdown_by_language}. 
    }
\end{table}

\section{Results and Discussion}

\subsection{Multilingual Multimodal Open-Ended Performance}
\textbf{Aya-Vision-8B achieves best-in-class performance in preference evaluation.}
Figure \ref{fig:mm-win-rates-8b} shows pairwise win-rates on AyaVisionBench and m-WildVision, averaged over 23 languages for Aya-Vision-8B against the other state-of-the-art multimodal LLMs. Overall, Aya-Vision-8B achieves the best-in-class performance, outperforming all the models by win-rate, ranging from 49.6\% to 80.3\%. We find that Aya-Vision-8B achieves slightly higher win-rates on m-WildVision compared to AyaVisionBench -- 6\% on average, potentially due to the challenging characteristic of AyaVisionBench -- higher tie rates also indicate failure cases for both models in the comparison. Aya-Vision-8B outperforms both Qwen-2.5-VL-7B and Pixtral-12B by 54.8\% win-rate averaged across the two datasets, even though Pixtral-12B is a larger model. Additionally, Aya-Vision-8B also outperforms strong proprietary models like Gemini-Flash1.5-8B with a win-rate of 60.3\% on average. Notably, Aya-Vision-8B outperforms Pangea-7B by a significant margin (71.7\% win-rate) even though Pangea includes a large proportion of multilingual data in its training. 

Aya-Vision-8B also outperforms Pangea-7B across all 23 languages -- ranging from $56\%$ in English to $83.6\%$ in Greek. Given that the ``curse of multilinguality" leads to drop in per-language performance as the number of languages covered increases, Aya-Vision-8B is still extremely competitive with Molmo-7B (specifically optimized for English) with a win-rate of $48.3\%$ in English while outperforming it over the other 22 languages with an average win-rate of $80\%$.

Finally, in addition to AyaVisionBench and m-WildVision, Aya-Vision-8B outperforms all models in the same parameter class on xChatBench as shown in Table \ref{tab:academic-evals}. Notably, Aya-Vision-8B not only achieves a significant margin against models like Pangea-7B, Molmo-7B-D, and Llama-3.2-11B, but also outperforms much larger models such as Molmo-72B and Llama-3.2-90B by 28.5\% and 14.7\% relative increase, validating its strong conversational ability.

\textbf{Aya Vision outperforms far larger models.} While scaling model size has demonstrated tangible gains in model performance \citep{kaplan2020scaling}; complementing this with careful data and model optimization techniques yields significant efficiency gains. Such optimizations improve the underlying scaling dynamics, reducing the parameter count needed for equivalent performance \citep{hooker2024limitationscomputethresholdsgovernance}.
Figure \ref{fig:mm-win-rates-32b} shows pairwise win-rates averaged over 23 languages for Aya-Vision-32B on AyaVisionBench and m-WildVision. Across both AyaVisionBench and m-WildVision, Aya-Vision-32B consistently outperforms models over $2\times$ larger, such as Molmo-72B, Qwen-2.5-VL-72B, and Llama-3.2-90B-Vision by win-rates ranging from 48.5\% to 73\%. Notably, Aya-Vision-32B outperforms Llama-3.2-90B-Vision on AyaVisionBench and m-WildVision by 65.9\% and 73\% win-rates, respectively. The closest competitor to Aya-Vision-32B is Qwen-2.5-VL-72B, where Aya-Vision-32B outperforms Qwen-2.5-VL-72B by 50.8\% win-rate on average across both datasets. This showcases our critical focus on efficiency by achieving more using less compute. This also enables greater support for the research community, who often have more limited access to compute resources.

\subsection{Multilingual Multimodal Academic Benchmarks}

\textbf{Aya Vision models achieve competitive performance in multiple-choice or short-form academic benchmarks.} 
Aya Vision models are optimized for open-ended real-world usage rather than academic benchmarks featuring multiple-choice or short-form answers. These benchmarks, typically designed as visual question answering tasks, tend to prioritize constrained, static evaluation formats and often fail to capture the full generative capabilities of modern MLLMs. As noted in prior work \citep{muennighoff2022crosslingual, agrawal2024pixtral, deitke2024molmo, ustun2024aya}, performance on such benchmarks correlates weakly with real-world open-ended tasks. Nonetheless, Aya Vision models demonstrate strong performance across these evaluations. Results are reported in Table~\ref{tab:academic-evals}.

Notably, on MaxM, a short-form VQA benchmark, Aya-Vision-8B outperforms all models in its parameter class, including larger ones like Pixtral-12B and LLaMA-3.2-11B-Vision. Similarly, on Kaleidoscope, it performs competitively with Qwen-2.5-VL-7B and surpasses all other baselines.

Finally, our 32B model Aya Vision model exhibits competitive performance on academic benchmarks against models more than $2\times$ its size. Aya-Vision-32B outperforms Molmo-72B on all benchmarks except xMMMU, and closely matches Llama-3.2-90B-Vision, despite being nearly 3$\times$ smaller.

\subsection{Text-Only Performance} 

\textbf{Aya Vision models punch above their size in text-only preference evaluation.} A key concern with multimodal models is that introducing vision can degrade existing text performance. Hence, we evaluate the final overall performance in text performance. \Cref{fig:m-arena-combined} shows win-rates for Aya Vision models against the baselines on m-ArenaHard dataset, averaged over 23 languages. At 8B parameter scale, Aya-Vision-8B outperforms all the models except Gemini-Flash1.5-8B, which is a proprietary model. Compared to models that are larger than ours, while Aya-Vision-8B beats Llama-3.2-11B-Vision with a 63.4\% win-rate, it is outperformed by Pixtral-12B with 44.0\% win-rate. For the larger model comparison, Aya-Vision-32B outperforms Molmo-72B and Qwen-2.5-VL-72B by win-rates of 77.3\% and 50.9\% respectively. Our 32B model is competitive with Llama-3.2-90B-Vision with a 43.2\% win-rate. Considered together with superior multimodal win-rates (\Cref{fig:mm-win-rates-8b} \& \Cref{fig:mm-win-rates-32b}), these results show the relative preservation in text performance while adding best-in-class multimodal abilities.

\textbf{Aya Vision recovers open-ended text-only performance in a significantly higher degree than the baselines.} As an additional perspective on text-only performance, \Cref{fig:text-degradation} compares the text-only win-rates on mArenaHard for Aya-Vision-8B, Pangea-7B, Qwen-2.5-7B, and Molmo-7B compared to the LLMs they were initialized from. Here, Aya-Vision-8B with cross-modal merging makes significant strides towards much closer performance to the initial LLM -- limiting the degradation to within $5.9\%$. This degradation, however, is significantly higher in the other models evaluated, 16.4\% for Pangea, 22.1\% for Qwen-2.5, and 44.1\% for Molmo compared to their initial LLMs. These results highlight the benefits of our cross-modal merging framework.

\textbf{It is easier to recover text-only performance in academic benchmarks compared to open-ended evaluation.} As we show in \S~\ref{sec:balance}, maintaining the base LLM's text-only performance in academic benchmarks is much easier than preference evaluation due to the nature of these benchmarks.
Hence, the performance of similar-sized models is closer in these benchmarks. At 8B parameter scale, Aya-Vision-8B achieves the best average performance across text-only benchmarks of $64.41\%$, where it outperforms all models in FLORES (\texttt{En$\rightarrow$X}, 23 languages) and reaches the second-best performance in both MGSM and IFEval, after Pixtral-12B and Llama-3.2-11B-Vision respectively. Notably, both models are much larger than our 8B model. For Aya-Vision-32B, our model achieves second-best performance for FLORES, but falls behind other models on other tasks. We relate this to the original performance of base LLMs in these benchmarks, where recovery is relatively straightforward. It is important to note that the models compared to Aya-Vision-32B are over $2\times$ its size (72B and 90B models). 
Overall, we observe that both multimodal and text-only academic benchmark results have poor alignment with their open-ended generation counterparts; as demonstrated in prior works \citep{muennighoff2022crosslingual,ustun2024aya} due to their rigid metrics emphasizing precise format compliance at the expense of semantic correctness and quality of generations.

\begin{figure}[t]
    \centering
    \includegraphics[width=0.9\linewidth]{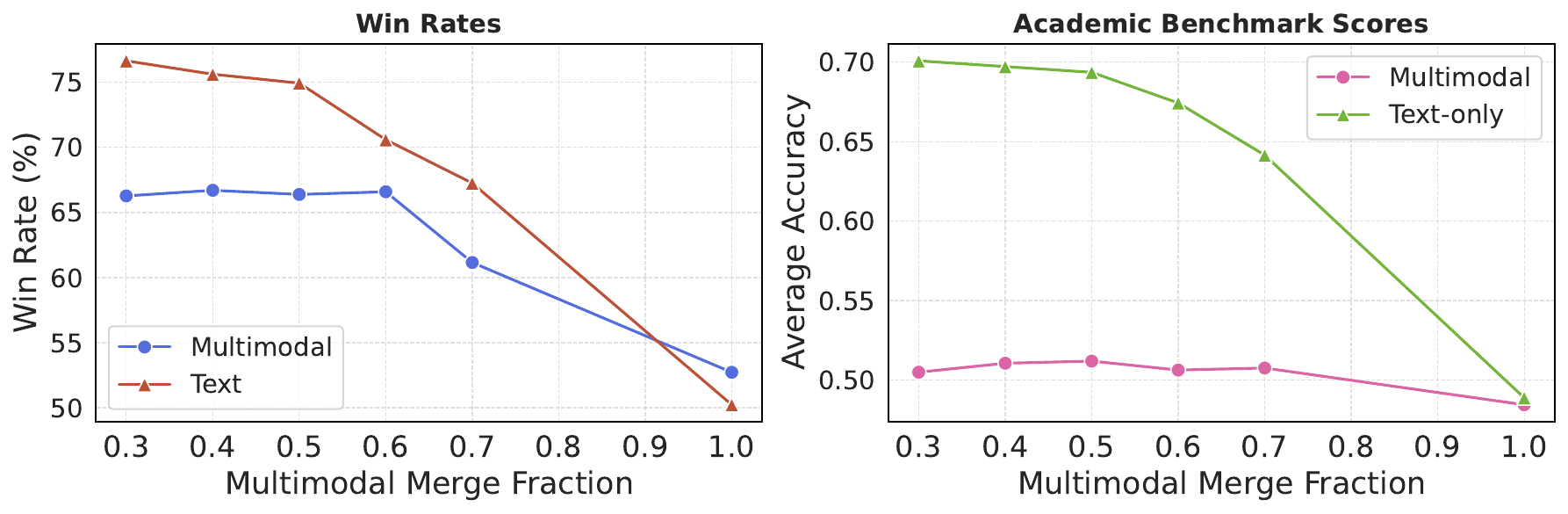}
    \caption{\textbf{Impact of cross-modal merging across various merge ratios.} Multimodal and text win-rates are calculated against Pangea-7B on AyaVisionBench and m-ArenaHard respectively over 7 languages. Multimodal academic benchmark is an average of CVQA and xMMMU; Text-Only academic benchmarks are averaged over IFEval, MGSM and MMMLU (subset).}
    \label{fig:merging-sweep}
\end{figure}

\section{Key Ablations and Discussion}
To isolate the impact of our design choices, we perform a set of controlled ablations focusing on -- (1) cross-modal model merging, (2) comparison with the addition of text-only data, (3) multilingual data percentage during SFT, (4) the vision encoder, and (5) comparison of full model fine-tuning with low-rank adaptation, all at the 8B parameter scale. In each of these ablations, we only vary a single variable of interest, while keeping the rest of the experimental setup fixed. To evaluate each ablation, we use multimodal win-rates on AyaVisionBench and text win-rates on mArena-Hard using a subset of languages\footnote{English, French, Hindi, Arabic, Turkish, Japanese, Chinese} against Pangea-7B. In addition, we also report scores on various academic benchmarks based on the ablation. 

\subsection{Model Merging Improves Multilingual Performance Across Tasks and Modalities}
\label{sec:merging-ablations}

To understand the impact of our cross-modal model merging as the merging ratio changes, we ablate the interpolation weight $\alpha$ in \Cref{eq:merging} for the multimodal LLM, and evaluate the resulting merged multimodal LLMs. An $\alpha$ of $0$ corresponds to purely the text-only model whereas an $\alpha$ of $1$ corresponds to just the post-multimodal training model. In addition to the win-rates for both multimodal and text-only, we report the average of CVQA and xMMMU for academic vision benchmarks and IFEval, MMMLU (subset), and MGSM for text-only academic benchmarks.

While our original motivation for model merging was retention of performance on text-only multilingual benchmarks, \Cref{fig:merging-sweep} (left) shows that our novel cross-modal merging recipe additionally boosts multilingual vision win-rates as the interpolation weight for text-only model increases. Below 0.6 multimodal interpolation weight, the text-only win-rates keep climbing; however, the vision win-rates saturate. For academic benchmarks, we again observe a similar trend -- as the ratio of the text-only model increases, text-only benchmarks rapidly increase until 0.5, following which the gains are minimal. Interestingly, even academic multimodal benchmarks see a minor gain due to model merging. Based on these results, we chose $0.4$ as the merging ratio for both our 8B and 32B models.

\subsection{Model merging is more effective than adding \textit{seen} text data for cross-modal transfer}
\begin{wrapfigure}{r}{0.6\textwidth}
  \centering
  \includegraphics[width=0.6\textwidth, keepaspectratio=true]{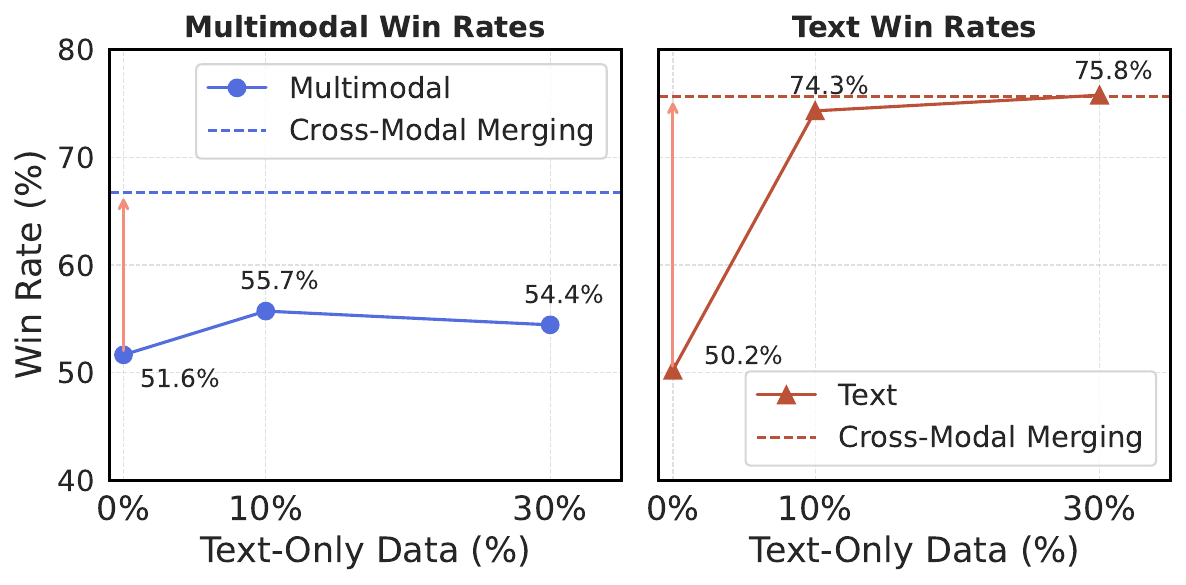}
  \caption{\textbf{Modal merging is an efficient way to enable cross-modal transfer.} Multimodal and text-only win-rates on AyaVisionBench and m-ArenaHard against Pangea-7B. We increase the amount of text-only mixture in SFT and compare to cross-modal merging (dashed line).}
  \label{fig:text-data-ablations}
  \vspace{0.4cm}
\end{wrapfigure}

An alternate approach to recover performance on text-only tasks is to include a certain percentage of text-only data in the training mixture. To understand the role of text-only data on multimodal and text-only win-rates and specifically compare it with our cross-modal merging approach, we train 3 variants with varying proportions of text data -- $0\%$, $10\%$, and $30\%$. For the variants with text-data added, we evaluate the final checkpoints without merging, and compare with the model where our merging recipe is applied on the variant with $0\%$ text-data. \Cref{fig:text-data-ablations} shows the results of these experiments. 

\label{sec:model-merging-more-effective}
While increasing the amount of text-only data improves the quality of generations for textual prompts as indicated by win-rates going from $50.2\%$ to $74.8\%$; these gains do not translate to multimodal prompts. In fact, as seen in \Cref{fig:text-data-ablations} these win-rates are substantially lower than those obtained by training on purely multimodal data followed by merging with a weight of 0.4. Additionally, increasing the amount of text data added from $10\%$ to $30\%$ leads to a slight decrease in the multimodal win-rates due to increasing share of model capacity being used for text-only modeling. This highlights the simplicity and efficacy of our model merging framework at cross-modal transfer of capabilities.

\begin{figure}[t]
    \centering
    \includegraphics[width=0.9\linewidth]{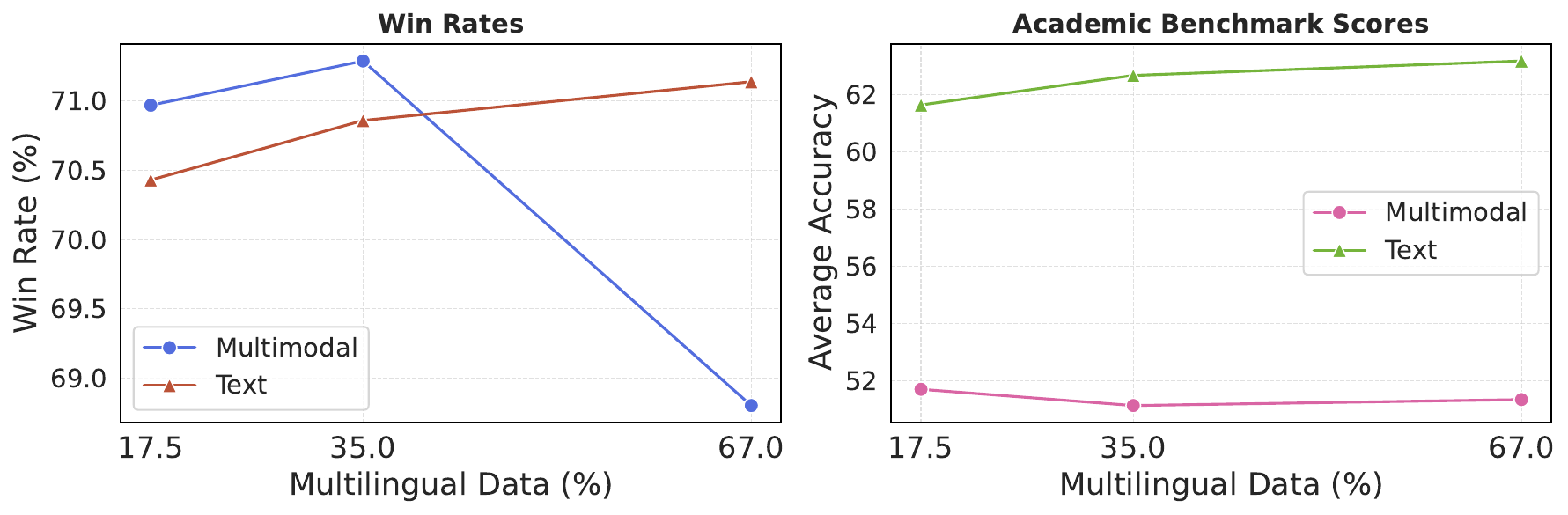}
    \caption{\textbf{A balanced data mixture is essential for multilingual multimodal performance.} Multimodal and text win-rates are calculated against Pangea-7B on AyaVisionBench and m-ArenaHard respectively over 7 languages. Multimodal academic benchmark is an average of CVQA and xMMMU; Text-Only academic benchmarks are averaged over IFEval, MGSM and MMMLU (subset).}
    \label{fig:multilingual-data}
\end{figure}

\newpage
\subsection{Data Improvements has the Highest Impact on the Quality of Generations}

\begin{wrapfigure}{r}{0.5\textwidth}
    \vspace{-0.4cm}
  \centering
  \includegraphics[width=0.5\textwidth]{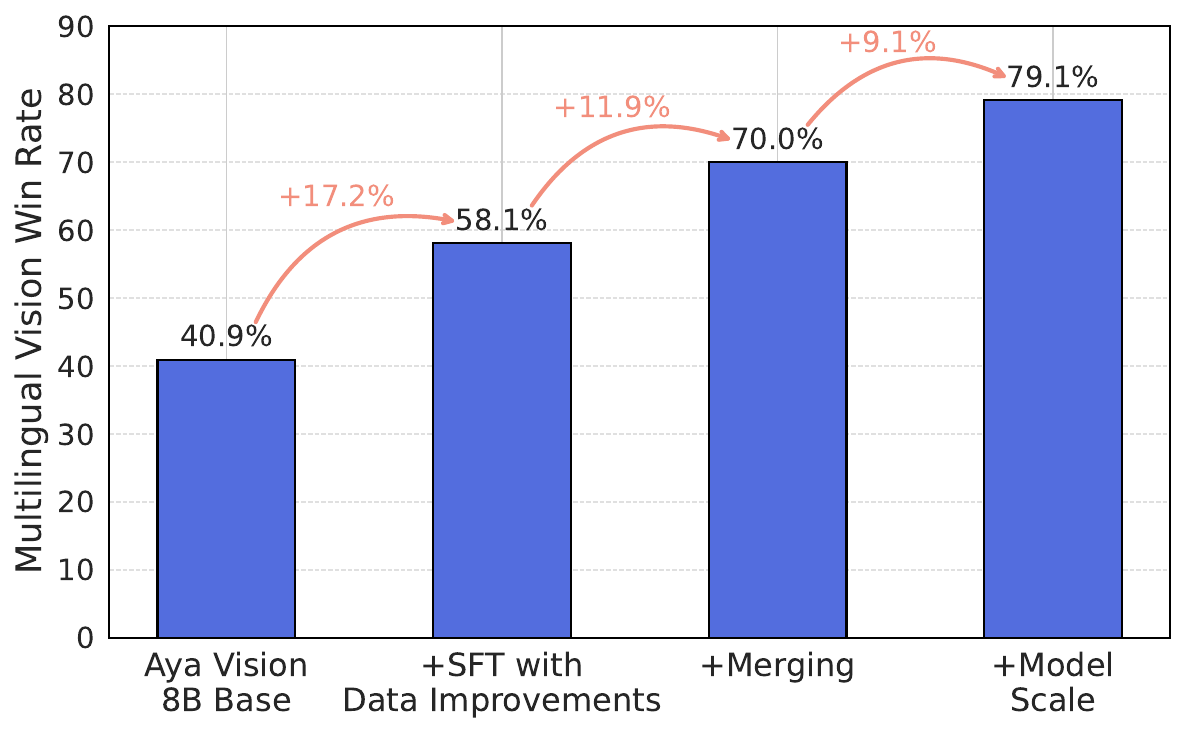}
  \caption{\textbf{Impact of various interventions.} Step-by-step improvements in Aya Vision 8B's pairwise win-rates against Pangea-7B.}
  \label{fig:stepwise}
  \vspace{-0.3cm}
\end{wrapfigure}
Our data generation framework has a strong emphasis on the quality but can we quantify the importance of the data improvement process? To answer this question, we train 2 variants -- (1) with \underline{only} existing open-source data, (2) with the data mixture proposed in \S~\ref{sec:balance} -- holding the amount of data and iterations during training fixed; and measure the multimodal win-rates. Please note that no merging is performed here to allow for a cleaner comparison. \Cref{fig:stepwise} shows the impact of synthetic annotations on the win-rates. Compared to variant (1) trained purely on original task-specific data, our data improvements lead to the largest jump in win-rates -- $17\%$ amongst our various interventions. This underscores the importance of fluent, detailed and diverse completions in the training data mixture towards building a strong conversational multimodal model. Additionally, when paired with cross-modal merging the total improvement increases to nearly $30\%$.

\subsection{A Balanced Data Mixture is Essential for Multilingual Multimodal Performance}
\label{sec:multilingual-percentage}

An important question in building a multilingual multimodal model is -- \textit{What is the right ratio of multilingual data in the training mixture?} 

To answer this question, we train 3 variants with varying proportions of multilingual multimodal data -- $17.5\%$, $35\%$, and $67\%$, which is uniformly distributed across 22 languages (except English). We compare these variants using preference evaluation (win-rates), and a subset of multimodal and text-only academic benchmarks. Note that we merge each trained checkpoint with the text-only model with the same interpolation factor ($\alpha$) to make it consistent with our final recipe. \Cref{fig:multilingual-data} shows the results. 

\textbf{Balanced multilingual data leverages cross-lingual transfer from English for best performance across modalities and languages.} 
We observe that increasing the ratio of multilingual multimodal data from 35\% to 67\% leads to degradation in the quality of generations -- reducing the win-rates from $71.4\%$ to $68.7\%$, and also hurts multimodal academic benchmarks, emphasizing the importance of the balance between English and multilingual data. Given the scarcity of high-quality multilingual multimodal data, upsampling this bucket requires repeating the data multiple times, limiting its benefit in multilingual multimodal performance. Additionally, a sufficient percentage of the more diverse English data
is crucial for cross-lingual transfer.
Therefore, we use 35\% of multilingual data in our final recipe, leaving 65\% for a diverse set of English datasets, which includes selected original datasets (34\%), and a high-quality synthetically re-annotated dataset (31\%) as presented in Section \ref{sec:balance}.

\subsection{Low Rank Finetuning is Comparable to Full Finetuning}
\begin{wrapfigure}{r}{0.5\textwidth}
  \centering
    \vspace{-0.5cm}
    \includegraphics[width=\linewidth]{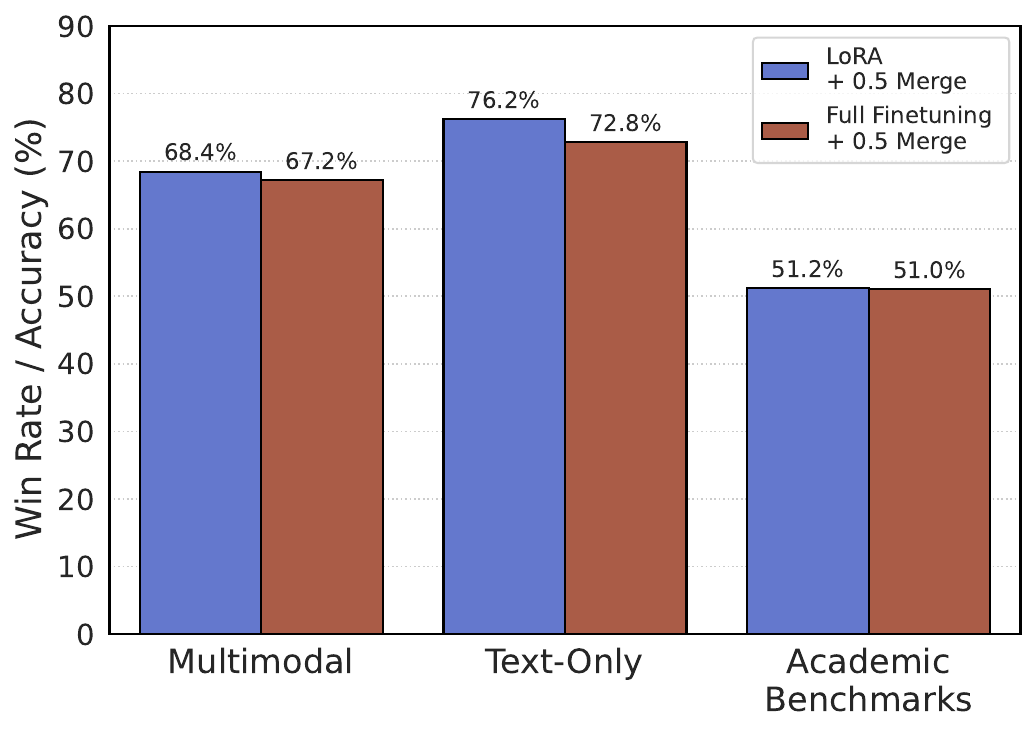}
    \caption{\textbf{Impact of training with LoRA vs. Full-Finetuning.} We compare vision win-rates (left) and text-only win-rates (center) against Pangea-7B averaged across 7 languages. We also report the average of CVQA and xMMMU (right).}
    \label{fig:lora-vs-fullft}
\end{wrapfigure}

Low-rank training (LoRA) is an extremely performant method to reduce the hardware footprint during training for improved efficiency. LoRA drastically reduces the number of trainable parameters and optimizer states to be stored in the accelerator memory \citep{zadouri2023pushing}. Furthermore, freezing the LLM and constraining the rank of updates has the potential to prevent catastrophic forgetting on text-only prompts. To understand the impact of the rank of training updates during the SFT stage, we train 2 variants on the same data -- (1) trained with LoRA ($\text{rank} = 256$, $\alpha=512$) \citep{hu2022lora} while (2) is trained with full finetuning (all network weights are updated). Once both the models are trained, we merge the multimodal updates to the text-only language model with a weight ($\alpha$) of 0.5. Finally, we evaluate both variants on multimodal and text win-rates; and academic benchmarks like CVQA and xMMMU. \Cref{fig:lora-vs-fullft} shows the results on all the above tasks.

On academic tasks like CVQA and xMMMU, we observe that both variants perform equally well, 51.2 vs 51.0 average accuracy for LoRA and full model fine-tuning, respectively. On multimodal win-rate evaluations, both models are extremely close -- with $68.4\%$ and $67.2\%$ win-rates for the LoRA and fully-finetuned variants respectively. Any improvement exhibited by the LoRA variant on win-rates is well within the noise-margin. On text-only win-rates, the LoRA variant is $3.4\%$ better than full-finetuning which can be attributed to the frozen LLM backbone during training and the amenability of LoRA model to merging due to the shared optimization trajectory.

\section{Related Work}
\textbf{Visual Instruction Tuning} Visual instruction tuning \citep{liu2023visual, chen2023pali, liu2024improved, chen2024far, agrawal2024pixtral, wang2024qwen2, deitke2024molmo, bai2025qwen25vl} combines a pre-trained vision encoder \citep{radford2021learning, zhai2023sigmoid,  chen2024far, tschannen2025siglip} with an off‐the‐shelf large language model via a dedicated vision–language connector. This process extends the LLM’s text capabilities into the visual domain while retaining its desirable attributes-- such as in-context learning, reasoning, and instruction following. As a result, visual instruction tuning has emerged as a highly effective method to achieve state-of-the-art performance on a wide range of tasks -- even outperforming certain proprietary models.

\textbf{Multilingual Multimodal Models} 
Initial works on multilingual multimodal models \citep{Ni_2021_CVPR, jain2021mural, zeng2023crossviewlanguagemodelingunified} focused on learning robust, universal representations for retrieval tasks across modalities. However, these models require further downstream training to be used as generative models. On the other hand, \citep{geigle2023mblip, chen2023pali, yue2024pangea} perform large-scale multilingual multi-task fine-tuning to enable multilingual understanding and generation. However, they focus only on vision-language academic benchmarks which are reference based -- focusing on exact matches rather than free-form holistic evaluations of the generations.

\textbf{Multilingual Multimodal Evaluations}
Multilingual multimodal evaluation benchmarks have traditionally focused on visual question answering (VQA) tasks, where the model-generated response must exactly match a human-provided reference answer \citep{changpinyo2022maxm,romero2024cvqa, tang2024mtvqa}. This approach often penalizes responses that are semantically correct but differ syntactically from the reference \citep{agrawal2024pixtral}. To address these limitations, recent work \citep{yue2024pangea, maaz2024palo} has proposed multilingual multimodal chat benchmarks. Instead of relying solely on exact matches, these benchmarks evaluate free-form responses by employing a Vision-Language model as an adjudicator--either by scoring responses against a detailed rubric or by selecting the superior generation from a pair of outputs.

\textbf{Multimodal Merging}
Recent work by \citet{zhu2025remedy} introduces REMEDY, a method for merging VLM weights -- including the connector layer -- after low-rank fine-tuning on various VLM tasks. However, REMEDY does not address the merging of weights that have been trained for different modalities. In a closely related concurrent work, \citet{li2025transferring} merge a text-only reward model with a vision-language model with the goal to specifically transfer the reward modeling capabilities from the text-based reward model to build a multimodal reward model.

\section{Conclusion}

In this work, we introduced Aya Vision, a family of multilingual vision-language models (8B and 32B) designed to improve multimodal understanding across 23 languages. Addressing key challenges in this space, we propose a scalable synthetic annotation framework to overcome multilingual data scarcity, and a training-free model merging approach to preserve text-only performance during multimodal training. Our models outperform existing open-weight baselines and are supported by AyaVisionBench, a benchmark tailored for evaluating generative multilingual multimodal systems. By releasing our models and evaluation suite,  we aim to lower barriers for research in this area and support continued progress toward more inclusive and linguistically diverse multimodal AI.

\section{Acknowledgements}

Thank you to all our colleagues across Cohere who jumped in to help test Aya Vision in their language: Ivan Zhang, Irem Ergün, Eddie Kim, Hemant Jain, Wei-Yin Ko, Adrian Morisot, Rod Hajjar, Gokce Keskin, Trushant Kalyanpur, Julia Kreutzer, Olivia Lasche, Dennis Aumiller, Felipe Cruz Salinas, Alice Schoenauer Sebag, Dwarak Talupuru, Diana Abagyan, Ammar Khairi, Huey Sun, Varun Kumethi, Viraat Aryabumi, Sungjin Hong, Trent Fowlers, Lidiya Murakhovska, Aidan  Peppin, Jay Alammar, Samuel Cahyawijaya, Brittawnya Prince, Daniel D’souza and Vivek Muppalla.

And to the members of our Open Science Community who shared their expertise and insights: Ahmad Anis, Amir Nuriyev, Bronson Bakunga,, Daniel Laurin, Danylo Boiko, David Cairuz, Dina Kliuchareva, Dominik Krzemiński, Erika Watanabe, Fernanda Guerriero Antunes, Gimei Alex, Jie Gao, Joana da Matta, Joseph Pollack, Karthik Kanjula, Kenny Rebelo, Kentaro Kojima, Kian Kyers, Lana Ludmila, Leticia Mie Otani, Louisa Chang, Marek Suppa, Mayuko Koizumi, Mei E., Mei Hirata, Micol Altomare, Nicole Mak, Ning Sun, Rami Rao, Reuben Fernandes, Reza Rob, Selina Tong, Shayekh Bin Islam, Shirley Au, Silvia Fernandez, Sree Harsha Nelaturu, Tai Guratti, Teresa Shiho Waddell, Thiago Correia, Xuelong An Wang, and Yanny Li.

The authors would also like to thank Fraser Greenlee for his contributions during the early stages of this project.

\bibliography{addon}

\begin{thebibliography}{130}
\providecommand{\natexlab}[1]{#1}
\providecommand{\url}[1]{\texttt{#1}}
\expandafter\ifx\csname urlstyle\endcsname\relax
  \providecommand{\doi}[1]{doi: #1}\else
  \providecommand{\doi}{doi: \begingroup \urlstyle{rm}\Url}\fi

\bibitem[Aakanksha et~al.(2024{\natexlab{a}})Aakanksha, Ahmadian, Goldfarb-Tarrant, Ermis, Fadaee, and Hooker]{aakanksha2024mixdatamergemodels}
Aakanksha, Arash Ahmadian, Seraphina Goldfarb-Tarrant, Beyza Ermis, Marzieh Fadaee, and Sara Hooker.
\newblock Mix data or merge models? optimizing for diverse multi-task learning, 2024{\natexlab{a}}.
\newblock URL \url{https://arxiv.org/abs/2410.10801}.

\bibitem[Aakanksha et~al.(2024{\natexlab{b}})Aakanksha, Ermis, Goldfarb-Tarrant, Kreutzer, Fadaee, Hooker, et~al.]{aakanksha2024multilingual}
Arash Aakanksha, Ahmadian, Beyza Ermis, Seraphina Goldfarb-Tarrant, Julia Kreutzer, Marzieh Fadaee, Sara Hooker, et~al.
\newblock The multilingual alignment prism: Aligning global and local preferences to reduce harm.
\newblock \emph{arXiv preprint arXiv:2406.18682}, 2024{\natexlab{b}}.

\bibitem[Acharya et~al.(2019)Acharya, Kafle, and Kanan]{TallyQA}
Manoj Acharya, Kushal Kafle, and Christopher Kanan.
\newblock Tallyqa: Answering complex counting questions.
\newblock In \emph{AAAI}, 2019.

\bibitem[Agrawal et~al.(2024)Agrawal, Antoniak, Hanna, Bout, Chaplot, Chudnovsky, Costa, De~Monicault, Garg, Gervet, et~al.]{agrawal2024pixtral}
Pravesh Agrawal, Szymon Antoniak, Emma~Bou Hanna, Baptiste Bout, Devendra Chaplot, Jessica Chudnovsky, Diogo Costa, Baudouin De~Monicault, Saurabh Garg, Theophile Gervet, et~al.
\newblock Pixtral 12b.
\newblock \emph{arXiv preprint arXiv:2410.07073}, 2024.

\bibitem[{Anthropic}(2025)]{claude3.7systemcard}
{Anthropic}.
\newblock Claude 3.7 sonnet system card.
\newblock \url{https://assets.anthropic.com/m/785e231869ea8b3b/original/claude-3-7-sonnet-system-card.pdf}, February 2025.
\newblock Accessed: 2025-04-17.

\bibitem[Aryabumi et~al.(2024{\natexlab{a}})Aryabumi, Dang, Talupuru, Dash, Cairuz, Lin, Venkitesh, Smith, Campos, Tan, Marchisio, Bartolo, Ruder, Locatelli, Kreutzer, Frosst, Gomez, Blunsom, Fadaee, Üstün, and Hooker]{aryabumi2024aya23openweight}
Viraat Aryabumi, John Dang, Dwarak Talupuru, Saurabh Dash, David Cairuz, Hangyu Lin, Bharat Venkitesh, Madeline Smith, Jon~Ander Campos, Yi~Chern Tan, Kelly Marchisio, Max Bartolo, Sebastian Ruder, Acyr Locatelli, Julia Kreutzer, Nick Frosst, Aidan Gomez, Phil Blunsom, Marzieh Fadaee, Ahmet Üstün, and Sara Hooker.
\newblock Aya 23: Open weight releases to further multilingual progress, 2024{\natexlab{a}}.
\newblock URL \url{https://arxiv.org/abs/2405.15032}.

\bibitem[Aryabumi et~al.(2024{\natexlab{b}})Aryabumi, Dang, Talupuru, Dash, Cairuz, Lin, Venkitesh, Smith, Campos, Tan, et~al.]{aryabumi2024aya}
Viraat Aryabumi, John Dang, Dwarak Talupuru, Saurabh Dash, David Cairuz, Hangyu Lin, Bharat Venkitesh, Madeline Smith, Jon~Ander Campos, Yi~Chern Tan, et~al.
\newblock Aya 23: Open weight releases to further multilingual progress.
\newblock \emph{arXiv preprint arXiv:2405.15032}, 2024{\natexlab{b}}.

\bibitem[Bai et~al.(2023)Bai, Bai, Yang, Wang, Tan, Wang, Lin, Zhou, and Zhou]{qwenvl2023}
Jinze Bai, Shuai Bai, Shusheng Yang, Shijie Wang, Sinan Tan, Peng Wang, Junyang Lin, Chang Zhou, and Jingren Zhou.
\newblock Qwen-vl: A versatile vision-language model for understanding, localization, text reading, and beyond.
\newblock \emph{arXiv preprint arXiv:2308.12966}, 2023.

\bibitem[Bai et~al.(2025)Bai, Chen, Liu, Wang, Ge, Song, Dang, Wang, Wang, Tang, Zhong, Zhu, Yang, Li, Wan, Wang, Ding, Fu, Xu, Ye, Zhang, Xie, Cheng, Zhang, Yang, Xu, and Lin]{bai2025qwen25vl}
Shuai Bai, Keqin Chen, Xuejing Liu, Jialin Wang, Wenbin Ge, Sibo Song, Kai Dang, Peng Wang, Shijie Wang, Jun Tang, Humen Zhong, Yuanzhi Zhu, Mingkun Yang, Zhaohai Li, Jianqiang Wan, Pengfei Wang, Wei Ding, Zheren Fu, Yiheng Xu, Jiabo Ye, Xi~Zhang, Tianbao Xie, Zesen Cheng, Hang Zhang, Zhibo Yang, Haiyang Xu, and Junyang Lin.
\newblock Qwen2.5-vl technical report, 2025.
\newblock URL \url{https://arxiv.org/abs/2502.13923}.

\bibitem[Ben-Kish et~al.(2023)Ben-Kish, Yanuka, Alper, Giryes, and Averbuch-Elor]{ben2023mocha}
Assaf Ben-Kish, Moran Yanuka, Morris Alper, Raja Giryes, and Hadar Averbuch-Elor.
\newblock Mocha: Multi-objective reinforcement mitigating caption hallucinations.
\newblock \emph{arXiv preprint arXiv:2312.03631}, 2, 2023.

\bibitem[Beyer et~al.(2024)Beyer, Steiner, Pinto, Kolesnikov, Wang, Salz, Neumann, Alabdulmohsin, Tschannen, Bugliarello, et~al.]{beyer2024paligemma}
Lucas Beyer, Andreas Steiner, Andr{\'e}~Susano Pinto, Alexander Kolesnikov, Xiao Wang, Daniel Salz, Maxim Neumann, Ibrahim Alabdulmohsin, Michael Tschannen, Emanuele Bugliarello, et~al.
\newblock Paligemma: A versatile 3b vlm for transfer.
\newblock \emph{arXiv preprint arXiv:2407.07726}, 2024.

\bibitem[Biten et~al.(2019)Biten, Tito, Mafla, Gomez, Rusiñol, Jawahar, Valveny, and Karatzas]{STVQA}
Ali~Furkan Biten, Rubèn Tito, Andrés Mafla, Lluis Gomez, Marçal Rusiñol, C.V. Jawahar, Ernest Valveny, and Dimosthenis Karatzas.
\newblock Scene text visual question answering.
\newblock In \emph{2019 IEEE/CVF International Conference on Computer Vision (ICCV)}, pp.\  4290--4300, 2019.
\newblock \doi{10.1109/ICCV.2019.00439}.

\bibitem[Bizzoni et~al.(2020)Bizzoni, Juzek, Espa{\~n}a-Bonet, Chowdhury, van Genabith, and Teich]{bizzoni2020human}
Yuri Bizzoni, Tom~S Juzek, Cristina Espa{\~n}a-Bonet, Koel~Dutta Chowdhury, Josef van Genabith, and Elke Teich.
\newblock How human is machine translationese? comparing human and machine translations of text and speech.
\newblock In \emph{Proceedings of the 17th International conference on spoken language translation}, pp.\  280--290, 2020.

\bibitem[Changpinyo et~al.(2022)Changpinyo, Xue, Yarom, Thapliyal, Szpektor, Amelot, Chen, and Soricut]{changpinyo2022maxm}
Soravit Changpinyo, Linting Xue, Michal Yarom, Ashish~V Thapliyal, Idan Szpektor, Julien Amelot, Xi~Chen, and Radu Soricut.
\newblock Maxm: Towards multilingual visual question answering.
\newblock \emph{arXiv preprint arXiv:2209.05401}, 2022.

\bibitem[Chen et~al.(2023)Chen, Djolonga, Padlewski, Mustafa, Changpinyo, Wu, Ruiz, Goodman, Wang, Tay, et~al.]{chen2023pali}
Xi~Chen, Josip Djolonga, Piotr Padlewski, Basil Mustafa, Soravit Changpinyo, Jialin Wu, Carlos~Riquelme Ruiz, Sebastian Goodman, Xiao Wang, Yi~Tay, et~al.
\newblock Pali-x: On scaling up a multilingual vision and language model.
\newblock \emph{arXiv preprint arXiv:2305.18565}, 2023.

\bibitem[Chen et~al.(2024)Chen, Wang, Tian, Ye, Gao, Cui, Tong, Hu, Luo, Ma, et~al.]{chen2024far}
Zhe Chen, Weiyun Wang, Hao Tian, Shenglong Ye, Zhangwei Gao, Erfei Cui, Wenwen Tong, Kongzhi Hu, Jiapeng Luo, Zheng Ma, et~al.
\newblock How far are we to gpt-4v? closing the gap to commercial multimodal models with open-source suites.
\newblock \emph{Science China Information Sciences}, 67\penalty0 (12):\penalty0 220101, 2024.

\bibitem[Chen et~al.(2021)Chen, Chen, Smiley, Shah, Borova, Langdon, Moussa, Beane, Huang, Routledge, et~al.]{chen2021finqa}
Zhiyu Chen, Wenhu Chen, Charese Smiley, Sameena Shah, Iana Borova, Dylan Langdon, Reema Moussa, Matt Beane, Ting-Hao Huang, Bryan Routledge, et~al.
\newblock Finqa: A dataset of numerical reasoning over financial data.
\newblock \emph{arXiv preprint arXiv:2109.00122}, 2021.

\bibitem[Cohere et~al.(2025)Cohere, Aakanksha, Ahmadian, Ahmed, Alammar, Alnumay, Althammer, Arkhangorodsky, Aryabumi, Aumiller, Avalos, Aviv, Bae, Baji, Barbet, Bartolo, Bebensee, Beladia, Beller-Morales, Bérard, Berneshawi, Bialas, Blunsom, Bobkin, Bongale, Braun, Brunet, Cahyawijaya, Cairuz, Campos, Cao, Cao, Castagné, Cendrero, Currie, Chandak, Chang, Chatziveroglou, Chen, Cheng, Chevalier, Chiu, Cho, Choi, Choi, Chung, Cirik, Cismaru, Clavier, Conklin, Crawhall-Stein, Crouse, Cruz-Salinas, Cyrus, D'souza, Dalla-Torre, Dang, Darling, Domingues, Dash, Debugne, Dehaze, Desai, Devassy, Dholakia, Duffy, Edalati, Eldeib, Elkady, Elsharkawy, Ergün, Ermis, Fadaee, Fan, Fayoux, Flet-Berliac, Frosst, Gallé, Galuba, Garg, Geist, Azar, Goldfarb-Tarrant, Goldsack, Gomez, Gonzaga, Govindarajan, Govindassamy, Grinsztajn, Gritsch, Gu, Guo, Haefeli, Hajjar, Hawes, He, Hofstätter, Hong, Hooker, Hosking, Howe, Hu, Huang, Jain, Jain, Jakobi, Jenkins, Jordan, Joshi, Jung, Kalyanpur, Kamalakara, Kedrzycki, Keskin, Kim,
  Kim, Ko, Kocmi, Kozakov, Kryściński, Jain, Teru, Land, Lasby, Lasche, Lee, Lewis, Li, Li, Lin, Locatelli, Luong, Ma, Mach, Machado, Magbitang, Lopez, Mann, Marchisio, Markham, Matton, McKinney, McLoughlin, Mokry, Morisot, Moulder, Moynehan, Mozes, Muppalla, Murakhovska, Nagarajan, Nandula, Nasir, Nehra, Netto-Rosen, Ohashi, Owers-Bardsley, Ozuzu, Padilla, Park, Passaglia, Pekmez, Penstone, Piktus, Ploeg, Poulton, Qi, Raghvendra, Ramos, Ranjan, Richemond, Robert-Michon, Rodriguez, Roy, Ruis, Rust, Sachan, Salamanca, Saravanakumar, Satyakam, Sebag, Sen, Sepehri, Seshadri, Shen, Sherborne, Shi, Shivaprasad, Shmyhlo, Shrinivason, Shteinbuk, Shukayev, Simard, Snyder, Spataru, Spooner, Starostina, Strub, Su, Sun, Talupuru, Tarassov, Tommasone, Tracey, Trend, Tumer, Üstün, Venkitesh, Venuto, Verga, Voisin, Wang, Wang, Wang, Wen, White, Willman, Winkels, Xia, Xie, Xu, Yang, Yi-Chern, Zhang, Zhao, and Zhao]{cohere2025commandaenterprisereadylarge}
Team Cohere, Aakanksha, Arash Ahmadian, Marwan Ahmed, Jay Alammar, Yazeed Alnumay, Sophia Althammer, Arkady Arkhangorodsky, Viraat Aryabumi, Dennis Aumiller, Raphaël Avalos, Zahara Aviv, Sammie Bae, Saurabh Baji, Alexandre Barbet, Max Bartolo, Björn Bebensee, Neeral Beladia, Walter Beller-Morales, Alexandre Bérard, Andrew Berneshawi, Anna Bialas, Phil Blunsom, Matt Bobkin, Adi Bongale, Sam Braun, Maxime Brunet, Samuel Cahyawijaya, David Cairuz, Jon~Ander Campos, Cassie Cao, Kris Cao, Roman Castagné, Julián Cendrero, Leila~Chan Currie, Yash Chandak, Diane Chang, Giannis Chatziveroglou, Hongyu Chen, Claire Cheng, Alexis Chevalier, Justin~T. Chiu, Eugene Cho, Eugene Choi, Eujeong Choi, Tim Chung, Volkan Cirik, Ana Cismaru, Pierre Clavier, Henry Conklin, Lucas Crawhall-Stein, Devon Crouse, Andres~Felipe Cruz-Salinas, Ben Cyrus, Daniel D'souza, Hugo Dalla-Torre, John Dang, William Darling, Omar~Darwiche Domingues, Saurabh Dash, Antoine Debugne, Théo Dehaze, Shaan Desai, Joan Devassy, Rishit Dholakia, Kyle
  Duffy, Ali Edalati, Ace Eldeib, Abdullah Elkady, Sarah Elsharkawy, Irem Ergün, Beyza Ermis, Marzieh Fadaee, Boyu Fan, Lucas Fayoux, Yannis Flet-Berliac, Nick Frosst, Matthias Gallé, Wojciech Galuba, Utsav Garg, Matthieu Geist, Mohammad~Gheshlaghi Azar, Seraphina Goldfarb-Tarrant, Tomas Goldsack, Aidan Gomez, Victor~Machado Gonzaga, Nithya Govindarajan, Manoj Govindassamy, Nathan Grinsztajn, Nikolas Gritsch, Patrick Gu, Shangmin Guo, Kilian Haefeli, Rod Hajjar, Tim Hawes, Jingyi He, Sebastian Hofstätter, Sungjin Hong, Sara Hooker, Tom Hosking, Stephanie Howe, Eric Hu, Renjie Huang, Hemant Jain, Ritika Jain, Nick Jakobi, Madeline Jenkins, JJ~Jordan, Dhruti Joshi, Jason Jung, Trushant Kalyanpur, Siddhartha~Rao Kamalakara, Julia Kedrzycki, Gokce Keskin, Edward Kim, Joon Kim, Wei-Yin Ko, Tom Kocmi, Michael Kozakov, Wojciech Kryściński, Arnav~Kumar Jain, Komal~Kumar Teru, Sander Land, Michael Lasby, Olivia Lasche, Justin Lee, Patrick Lewis, Jeffrey Li, Jonathan Li, Hangyu Lin, Acyr Locatelli, Kevin Luong,
  Raymond Ma, Lukas Mach, Marina Machado, Joanne Magbitang, Brenda~Malacara Lopez, Aryan Mann, Kelly Marchisio, Olivia Markham, Alexandre Matton, Alex McKinney, Dominic McLoughlin, Jozef Mokry, Adrien Morisot, Autumn Moulder, Harry Moynehan, Maximilian Mozes, Vivek Muppalla, Lidiya Murakhovska, Hemangani Nagarajan, Alekhya Nandula, Hisham Nasir, Shauna Nehra, Josh Netto-Rosen, Daniel Ohashi, James Owers-Bardsley, Jason Ozuzu, Dennis Padilla, Gloria Park, Sam Passaglia, Jeremy Pekmez, Laura Penstone, Aleksandra Piktus, Case Ploeg, Andrew Poulton, Youran Qi, Shubha Raghvendra, Miguel Ramos, Ekagra Ranjan, Pierre Richemond, Cécile Robert-Michon, Aurélien Rodriguez, Sudip Roy, Laura Ruis, Louise Rust, Anubhav Sachan, Alejandro Salamanca, Kailash~Karthik Saravanakumar, Isha Satyakam, Alice~Schoenauer Sebag, Priyanka Sen, Sholeh Sepehri, Preethi Seshadri, Ye~Shen, Tom Sherborne, Sylvie~Chang Shi, Sanal Shivaprasad, Vladyslav Shmyhlo, Anirudh Shrinivason, Inna Shteinbuk, Amir Shukayev, Mathieu Simard, Ella Snyder,
  Ava Spataru, Victoria Spooner, Trisha Starostina, Florian Strub, Yixuan Su, Jimin Sun, Dwarak Talupuru, Eugene Tarassov, Elena Tommasone, Jennifer Tracey, Billy Trend, Evren Tumer, Ahmet Üstün, Bharat Venkitesh, David Venuto, Pat Verga, Maxime Voisin, Alex Wang, Donglu Wang, Shijian Wang, Edmond Wen, Naomi White, Jesse Willman, Marysia Winkels, Chen Xia, Jessica Xie, Minjie Xu, Bowen Yang, Tan Yi-Chern, Ivan Zhang, Zhenyu Zhao, and Zhoujie Zhao.
\newblock Command a: An enterprise-ready large language model, 2025.
\newblock URL \url{https://arxiv.org/abs/2504.00698}.

\bibitem[Costa-Juss{\`a} et~al.(2022)Costa-Juss{\`a}, Cross, {\c{C}}elebi, Elbayad, Heafield, Heffernan, Kalbassi, Lam, Licht, Maillard, et~al.]{costa2022no}
Marta~R Costa-Juss{\`a}, James Cross, Onur {\c{C}}elebi, Maha Elbayad, Kenneth Heafield, Kevin Heffernan, Elahe Kalbassi, Janice Lam, Daniel Licht, Jean Maillard, et~al.
\newblock No language left behind: Scaling human-centered machine translation.
\newblock \emph{arXiv preprint arXiv:2207.04672}, 2022.

\bibitem[Dai et~al.(2024)Dai, Lee, Wang, Yang, Liu, Barker, Rintamaki, Shoeybi, Catanzaro, and Ping]{dai2024nvlm}
Wenliang Dai, Nayeon Lee, Boxin Wang, Zhuolin Yang, Zihan Liu, Jon Barker, Tuomas Rintamaki, Mohammad Shoeybi, Bryan Catanzaro, and Wei Ping.
\newblock Nvlm: Open frontier-class multimodal llms.
\newblock \emph{arXiv preprint arXiv:2409.11402}, 2024.

\bibitem[Dang et~al.(2024)Dang, Singh, D'souza, Ahmadian, Salamanca, Smith, Peppin, Hong, Govindassamy, Zhao, et~al.]{dang2024aya}
John Dang, Shivalika Singh, Daniel D'souza, Arash Ahmadian, Alejandro Salamanca, Madeline Smith, Aidan Peppin, Sungjin Hong, Manoj Govindassamy, Terrence Zhao, et~al.
\newblock Aya expanse: Combining research breakthroughs for a new multilingual frontier.
\newblock \emph{arXiv preprint arXiv:2412.04261}, 2024.

\bibitem[Deitke et~al.(2024)Deitke, Clark, Lee, Tripathi, Yang, Park, Salehi, Muennighoff, Lo, Soldaini, et~al.]{deitke2024molmo}
Matt Deitke, Christopher Clark, Sangho Lee, Rohun Tripathi, Yue Yang, Jae~Sung Park, Mohammadreza Salehi, Niklas Muennighoff, Kyle Lo, Luca Soldaini, et~al.
\newblock Molmo and pixmo: Open weights and open data for state-of-the-art multimodal models.
\newblock \emph{arXiv preprint arXiv:2409.17146}, 2024.

\bibitem[Ding et~al.(2023)Ding, Luo, Chung, and Han]{ding2023vqa}
Yihao Ding, Siwen Luo, Hyunsuk Chung, and Soyeon~Caren Han.
\newblock Vqa: A new dataset for real-world vqa on pdf documents.
\newblock In \emph{Joint European Conference on Machine Learning and Knowledge Discovery in Databases}, pp.\  585--601. Springer, 2023.

\bibitem[Ermis et~al.(2024)Ermis, Pozzobon, Hooker, and Lewis]{ermis-etal-2024-one}
Beyza Ermis, Luiza Pozzobon, Sara Hooker, and Patrick Lewis.
\newblock From one to many: Expanding the scope of toxicity mitigation in language models.
\newblock In Lun-Wei Ku, Andre Martins, and Vivek Srikumar (eds.), \emph{Findings of the Association for Computational Linguistics: ACL 2024}, pp.\  15041--15058, Bangkok, Thailand, August 2024. Association for Computational Linguistics.
\newblock \doi{10.18653/v1/2024.findings-acl.893}.
\newblock URL \url{https://aclanthology.org/2024.findings-acl.893/}.

\bibitem[Fang et~al.(2024)Fang, Zhu, Lu, Wang, Molchanov, Kautz, Cho, Pavone, Han, and Yin]{fang2024vila}
Yunhao Fang, Ligeng Zhu, Yao Lu, Yan Wang, Pavlo Molchanov, Jan Kautz, Jang~Hyun Cho, Marco Pavone, Song Han, and Hongxu Yin.
\newblock Vila$^2$: Vila augmented vila, 2024.
\newblock URL \url{https://arxiv.org/abs/2407.17453}.

\bibitem[Frankle et~al.(2020)Frankle, Dziugaite, Roy, and Carbin]{frankle2020linear}
Jonathan Frankle, Gintare~Karolina Dziugaite, Daniel Roy, and Michael Carbin.
\newblock Linear mode connectivity and the lottery ticket hypothesis.
\newblock In \emph{International Conference on Machine Learning}, pp.\  3259--3269. PMLR, 2020.

\bibitem[Gadre et~al.(2023)Gadre, Ilharco, Fang, Hayase, Smyrnis, Nguyen, Marten, Wortsman, Ghosh, Zhang, et~al.]{Nguyen2023}
Samir~Yitzhak Gadre, Gabriel Ilharco, Alex Fang, Jonathan Hayase, Georgios Smyrnis, Thao Nguyen, Ryan Marten, Mitchell Wortsman, Dhruba Ghosh, Jieyu Zhang, et~al.
\newblock Datacomp: In search of the next generation of multimodal datasets.
\newblock \emph{Advances in Neural Information Processing Systems}, 36:\penalty0 27092--27112, 2023.

\bibitem[Geigle et~al.(2023)Geigle, Jain, Timofte, and Glava{\v{s}}]{geigle2023mblip}
Gregor Geigle, Abhay Jain, Radu Timofte, and Goran Glava{\v{s}}.
\newblock mblip: Efficient bootstrapping of multilingual vision-llms.
\newblock \emph{arXiv preprint arXiv:2307.06930}, 2023.

\bibitem[Goddard et~al.(2024)Goddard, Siriwardhana, Ehghaghi, Meyers, Karpukhin, Benedict, McQuade, and Solawetz]{goddard2024arcee}
Charles Goddard, Shamane Siriwardhana, Malikeh Ehghaghi, Luke Meyers, Vladimir Karpukhin, Brian Benedict, Mark McQuade, and Jacob Solawetz.
\newblock Arcee’s mergekit: A toolkit for merging large language models.
\newblock In \emph{Proceedings of the 2024 Conference on Empirical Methods in Natural Language Processing: Industry Track}, pp.\  477--485, 2024.

\bibitem[Goyal et~al.(2017)Goyal, Khot, Summers-Stay, Batra, and Parikh]{goyal2017making}
Yash Goyal, Tejas Khot, Douglas Summers-Stay, Dhruv Batra, and Devi Parikh.
\newblock Making the v in vqa matter: Elevating the role of image understanding in visual question answering.
\newblock In \emph{Proceedings of the IEEE conference on computer vision and pattern recognition}, pp.\  6904--6913, 2017.

\bibitem[Grattafiori et~al.(2024)Grattafiori, Dubey, Jauhri, Pandey, Kadian, Al-Dahle, Letman, Mathur, Schelten, Vaughan, et~al.]{grattafiori2024llama}
Aaron Grattafiori, Abhimanyu Dubey, Abhinav Jauhri, Abhinav Pandey, Abhishek Kadian, Ahmad Al-Dahle, Aiesha Letman, Akhil Mathur, Alan Schelten, Alex Vaughan, et~al.
\newblock The llama 3 herd of models.
\newblock \emph{arXiv preprint arXiv:2407.21783}, 2024.

\bibitem[Gunjal et~al.(2023)Gunjal, Yin, and Bas]{gunjal2024detecting}
Anisha Gunjal, Jihan Yin, and Erhan Bas.
\newblock Detecting and preventing hallucinations in large vision language models.
\newblock \emph{arXiv preprint arXiv:2308.06394}, 2023.

\bibitem[Guo et~al.(2024)Guo, Zheng, Bai, Li, Wang, Zhu, Li, Neubig, Chen, and Yue]{guo2024mammoth}
Jarvis Guo, Tuney Zheng, Yuelin Bai, Bo~Li, Yubo Wang, King Zhu, Yizhi Li, Graham Neubig, Wenhu Chen, and Xiang Yue.
\newblock Mammoth-vl: Eliciting multimodal reasoning with instruction tuning at scale.
\newblock \emph{arXiv preprint arXiv:2412.05237}, 2024.

\bibitem[Guzm{\'a}n et~al.(2019)Guzm{\'a}n, Chen, Ott, Pino, Lample, Koehn, Chaudhary, and Ranzato]{guzmán2019flores}
Francisco Guzm{\'a}n, Peng-Jen Chen, Myle Ott, Juan Pino, Guillaume Lample, Philipp Koehn, Vishrav Chaudhary, and Marc'Aurelio Ranzato.
\newblock The flores evaluation datasets for low-resource machine translation: Nepali-english and sinhala-english.
\newblock \emph{arXiv preprint arXiv:1902.01382}, 2019.

\bibitem[Hartung et~al.(2023)Hartung, Herygers, Kurlekar, Zakaria, Volkan, Gr{\"o}ttrup, and Georges]{hartung2023measuring}
Kai Hartung, Aaricia Herygers, Shubham~Vijay Kurlekar, Khabbab Zakaria, Taylan Volkan, S{\"o}ren Gr{\"o}ttrup, and Munir Georges.
\newblock Measuring sentiment bias in machine translation.
\newblock In \emph{International Conference on Text, Speech, and Dialogue}, pp.\  82--93. Springer, 2023.

\bibitem[Hendy et~al.(2023)Hendy, Abdelrehim, Sharaf, Raunak, Gabr, Matsushita, Kim, Afify, and Awadalla]{hendy2023good}
Amr Hendy, Mohamed Abdelrehim, Amr Sharaf, Vikas Raunak, Mohamed Gabr, Hitokazu Matsushita, Young~Jin Kim, Mohamed Afify, and Hany~Hassan Awadalla.
\newblock How good are gpt models at machine translation? a comprehensive evaluation.
\newblock \emph{arXiv preprint arXiv:2302.09210}, 2023.

\bibitem[Hooker(2024)]{hooker2024limitationscomputethresholdsgovernance}
Sara Hooker.
\newblock On the limitations of compute thresholds as a governance strategy, 2024.
\newblock URL \url{https://arxiv.org/abs/2407.05694}.

\bibitem[Hsiao et~al.(2022)Hsiao, Zubach, Baechler, Carbune, Lin, Wang, Sunkara, Zhu, and Chen]{hsiao2022screenqa}
Yu-Chung Hsiao, Fedir Zubach, Gilles Baechler, Victor Carbune, Jason Lin, Maria Wang, Srinivas Sunkara, Yun Zhu, and Jindong Chen.
\newblock Screenqa: Large-scale question-answer pairs over mobile app screenshots.
\newblock \emph{arXiv preprint arXiv:2209.08199}, 2022.

\bibitem[Hu et~al.(2022)Hu, Shen, Wallis, Allen-Zhu, Li, Wang, Wang, Chen, et~al.]{hu2022lora}
Edward~J Hu, Yelong Shen, Phillip Wallis, Zeyuan Allen-Zhu, Yuanzhi Li, Shean Wang, Lu~Wang, Weizhu Chen, et~al.
\newblock Lora: Low-rank adaptation of large language models.
\newblock \emph{ICLR}, 1\penalty0 (2):\penalty0 3, 2022.

\bibitem[Ilharco et~al.(2022)Ilharco, Ribeiro, Wortsman, Gururangan, Schmidt, Hajishirzi, and Farhadi]{ilharco2022editing}
Gabriel Ilharco, Marco~Tulio Ribeiro, Mitchell Wortsman, Suchin Gururangan, Ludwig Schmidt, Hannaneh Hajishirzi, and Ali Farhadi.
\newblock Editing models with task arithmetic.
\newblock \emph{arXiv preprint arXiv:2212.04089}, 2022.

\bibitem[Izmailov et~al.(2018)Izmailov, Podoprikhin, Garipov, Vetrov, and Wilson]{izmailov2018averaging}
Pavel Izmailov, Dmitrii Podoprikhin, Timur Garipov, Dmitry Vetrov, and Andrew~Gordon Wilson.
\newblock Averaging weights leads to wider optima and better generalization.
\newblock \emph{arXiv preprint arXiv:1803.05407}, 2018.

\bibitem[Jain et~al.(2021)Jain, Guo, Srinivasan, Chen, Kudugunta, Jia, Yang, and Baldridge]{jain2021mural}
Aashi Jain, Mandy Guo, Krishna Srinivasan, Ting Chen, Sneha Kudugunta, Chao Jia, Yinfei Yang, and Jason Baldridge.
\newblock Mural: multimodal, multitask retrieval across languages.
\newblock \emph{arXiv preprint arXiv:2109.05125}, 2021.

\bibitem[Kafle et~al.(2018)Kafle, Price, Cohen, and Kanan]{DVQA}
Kushal Kafle, Brian Price, Scott Cohen, and Christopher Kanan.
\newblock Dvqa: Understanding data visualizations via question answering.
\newblock In \emph{Proceedings of the IEEE conference on computer vision and pattern recognition}, pp.\  5648--5656, 2018.

\bibitem[Kahou et~al.(2017)Kahou, Michalski, Atkinson, K{\'a}d{\'a}r, Trischler, and Bengio]{FigureQA}
Samira~Ebrahimi Kahou, Vincent Michalski, Adam Atkinson, {\'A}kos K{\'a}d{\'a}r, Adam Trischler, and Yoshua Bengio.
\newblock Figureqa: An annotated figure dataset for visual reasoning.
\newblock \emph{arXiv preprint arXiv:1710.07300}, 2017.

\bibitem[Kaplan et~al.(2020)Kaplan, McCandlish, Henighan, Brown, Chess, Child, Gray, Radford, Wu, and Amodei]{kaplan2020scaling}
Jared Kaplan, Sam McCandlish, Tom Henighan, Tom~B Brown, Benjamin Chess, Rewon Child, Scott Gray, Alec Radford, Jeffrey Wu, and Dario Amodei.
\newblock Scaling laws for neural language models.
\newblock \emph{arXiv preprint arXiv:2001.08361}, 2020.

\bibitem[Kembhavi et~al.(2016)Kembhavi, Salvato, Kolve, Seo, Hajishirzi, and Farhadi]{kembhavi2016diagram}
Aniruddha Kembhavi, Mike Salvato, Eric Kolve, Minjoon Seo, Hannaneh Hajishirzi, and Ali Farhadi.
\newblock A diagram is worth a dozen images.
\newblock In \emph{Computer Vision--ECCV 2016: 14th European Conference, Amsterdam, The Netherlands, October 11--14, 2016, Proceedings, Part IV 14}, pp.\  235--251. Springer, 2016.

\bibitem[Kembhavi et~al.(2017)Kembhavi, Seo, Schwenk, Choi, Farhadi, and Hajishirzi]{TQA}
Aniruddha Kembhavi, Minjoon Seo, Dustin Schwenk, Jonghyun Choi, Ali Farhadi, and Hannaneh Hajishirzi.
\newblock Are you smarter than a sixth grader? textbook question answering for multimodal machine comprehension.
\newblock In \emph{2017 IEEE Conference on Computer Vision and Pattern Recognition (CVPR)}, pp.\  5376--5384, 2017.
\newblock \doi{10.1109/CVPR.2017.571}.

\bibitem[Krishna et~al.(2017)Krishna, Zhu, Groth, Johnson, Hata, Kravitz, Chen, Kalantidis, Li, Shamma, et~al.]{krishna2017visual}
Ranjay Krishna, Yuke Zhu, Oliver Groth, Justin Johnson, Kenji Hata, Joshua Kravitz, Stephanie Chen, Yannis Kalantidis, Li-Jia Li, David~A Shamma, et~al.
\newblock Visual genome: Connecting language and vision using crowdsourced dense image annotations.
\newblock \emph{International journal of computer vision}, 123:\penalty0 32--73, 2017.

\bibitem[Kuznetsova et~al.(2020)Kuznetsova, Rom, Alldrin, Uijlings, Krasin, Pont-Tuset, Kamali, Popov, Malloci, Kolesnikov, et~al.]{kuznetsova2020open}
Alina Kuznetsova, Hassan Rom, Neil Alldrin, Jasper Uijlings, Ivan Krasin, Jordi Pont-Tuset, Shahab Kamali, Stefan Popov, Matteo Malloci, Alexander Kolesnikov, et~al.
\newblock The open images dataset v4: Unified image classification, object detection, and visual relationship detection at scale.
\newblock \emph{International journal of computer vision}, 128\penalty0 (7):\penalty0 1956--1981, 2020.

\bibitem[Lambert et~al.(2024)Lambert, Morrison, Pyatkin, Huang, Ivison, Brahman, Miranda, Liu, Dziri, Lyu, et~al.]{lambert2024t}
Nathan Lambert, Jacob Morrison, Valentina Pyatkin, Shengyi Huang, Hamish Ivison, Faeze Brahman, Lester James~V Miranda, Alisa Liu, Nouha Dziri, Shane Lyu, et~al.
\newblock T$\backslash$" ulu 3: Pushing frontiers in open language model post-training.
\newblock \emph{arXiv preprint arXiv:2411.15124}, 2024.

\bibitem[Lauren{\c{c}}on et~al.(2024{\natexlab{a}})Lauren{\c{c}}on, Marafioti, Sanh, and Tronchon]{laurenccon2024building}
Hugo Lauren{\c{c}}on, Andr{\'e}s Marafioti, Victor Sanh, and L{\'e}o Tronchon.
\newblock Building and better understanding vision-language models: insights and future directions.
\newblock In \emph{Workshop on Responsibly Building the Next Generation of Multimodal Foundational Models}, 2024{\natexlab{a}}.

\bibitem[Lauren{\c{c}}on et~al.(2024{\natexlab{b}})Lauren{\c{c}}on, Tronchon, Cord, and Sanh]{laurenccon2024matters}
Hugo Lauren{\c{c}}on, L{\'e}o Tronchon, Matthieu Cord, and Victor Sanh.
\newblock What matters when building vision-language models?
\newblock \emph{Advances in Neural Information Processing Systems}, 37:\penalty0 87874--87907, 2024{\natexlab{b}}.

\bibitem[Li et~al.(2023{\natexlab{a}})Li, Zhang, Chen, Wang, Pu, Yang, Li, and Liu]{li2023mimic}
Bo~Li, Yuanhan Zhang, Liangyu Chen, Jinghao Wang, Fanyi Pu, Jingkang Yang, Chunyuan Li, and Ziwei Liu.
\newblock Mimic-it: Multi-modal in-context instruction tuning.
\newblock \emph{arXiv preprint arXiv:2306.05425}, 2023{\natexlab{a}}.

\bibitem[Li et~al.(2025)Li, Lin, Chen, and Lee]{li2025transferring}
Chen-An Li, Tzu-Han Lin, Yun-Nung Chen, and Hung-yi Lee.
\newblock Transferring textual preferences to vision-language understanding through model merging.
\newblock \emph{arXiv preprint arXiv:2502.13487}, 2025.

\bibitem[Li et~al.(2023{\natexlab{b}})Li, Yin, Li, Chen, Wang, Ren, Li, Yang, Xu, Sun, et~al.]{li2023m}
Lei Li, Yuwei Yin, Shicheng Li, Liang Chen, Peiyi Wang, Shuhuai Ren, Mukai Li, Yazheng Yang, Jingjing Xu, Xu~Sun, et~al.
\newblock M3it: A large-scale dataset towards multi-modal multilingual instruction tuning.
\newblock \emph{arXiv preprint arXiv:2306.04387}, 2023{\natexlab{b}}.

\bibitem[Li et~al.(2024)Li, Chiang, Frick, Dunlap, Wu, Zhu, Gonzalez, and Stoica]{li2024crowdsourced}
Tianle Li, Wei-Lin Chiang, Evan Frick, Lisa Dunlap, Tianhao Wu, Banghua Zhu, Joseph~E Gonzalez, and Ion Stoica.
\newblock From crowdsourced data to high-quality benchmarks: Arena-hard and benchbuilder pipeline.
\newblock \emph{arXiv preprint arXiv:2406.11939}, 2024.

\bibitem[Li et~al.(2023{\natexlab{c}})Li, Du, Zhou, Wang, Zhao, and Wen]{li2023evaluating}
Yifan Li, Yifan Du, Kun Zhou, Jinpeng Wang, Wayne~Xin Zhao, and Ji-Rong Wen.
\newblock Evaluating object hallucination in large vision-language models.
\newblock \emph{arXiv preprint arXiv:2305.10355}, 2023{\natexlab{c}}.

\bibitem[Lin et~al.(2014)Lin, Maire, Belongie, Hays, Perona, Ramanan, Doll{\'a}r, and Zitnick]{lin2014microsoft}
Tsung-Yi Lin, Michael Maire, Serge Belongie, James Hays, Pietro Perona, Deva Ramanan, Piotr Doll{\'a}r, and C~Lawrence Zitnick.
\newblock Microsoft coco: Common objects in context.
\newblock In \emph{Computer vision--ECCV 2014: 13th European conference, zurich, Switzerland, September 6-12, 2014, proceedings, part v 13}, pp.\  740--755. Springer, 2014.

\bibitem[Liu et~al.(2023{\natexlab{a}})Liu, Emerson, and Collier]{liu2023vsr}
Fangyu Liu, Guy Emerson, and Nigel Collier.
\newblock Visual spatial reasoning.
\newblock \emph{Transactions of the Association for Computational Linguistics}, 11:\penalty0 635--651, 2023{\natexlab{a}}.

\bibitem[Liu et~al.(2023{\natexlab{b}})Liu, Lin, Li, Wang, Yacoob, and Wang]{liu2023mitigating}
Fuxiao Liu, Kevin Lin, Linjie Li, Jianfeng Wang, Yaser Yacoob, and Lijuan Wang.
\newblock Mitigating hallucination in large multi-modal models via robust instruction tuning.
\newblock \emph{arXiv preprint arXiv:2306.14565}, 2023{\natexlab{b}}.

\bibitem[Liu et~al.(2023{\natexlab{c}})Liu, Li, Wu, and Lee]{liu2023visual}
Haotian Liu, Chunyuan Li, Qingyang Wu, and Yong~Jae Lee.
\newblock Visual instruction tuning.
\newblock \emph{Advances in neural information processing systems}, 36:\penalty0 34892--34916, 2023{\natexlab{c}}.

\bibitem[Liu et~al.(2024)Liu, Li, Li, and Lee]{liu2024improved}
Haotian Liu, Chunyuan Li, Yuheng Li, and Yong~Jae Lee.
\newblock Improved baselines with visual instruction tuning.
\newblock In \emph{Proceedings of the IEEE/CVF Conference on Computer Vision and Pattern Recognition}, pp.\  26296--26306, 2024.

\bibitem[Lu et~al.(2021)Lu, Gong, Jiang, Qiu, Huang, Liang, and Zhu]{lu2021inter}
Pan Lu, Ran Gong, Shibiao Jiang, Liang Qiu, Siyuan Huang, Xiaodan Liang, and Song-Chun Zhu.
\newblock Inter-gps: Interpretable geometry problem solving with formal language and symbolic reasoning.
\newblock \emph{arXiv preprint arXiv:2105.04165}, 2021.

\bibitem[Lu et~al.(2023)Lu, Qiu, Chang, Wu, Zhu, Rajpurohit, Clark, and Kalyan]{TabMWP}
Pan Lu, Liang Qiu, Kai-Wei Chang, Ying~Nian Wu, Song-Chun Zhu, Tanmay Rajpurohit, Peter Clark, and Ashwin Kalyan.
\newblock Dynamic prompt learning via policy gradient for semi-structured mathematical reasoning.
\newblock In \emph{International Conference on Learning Representations (ICLR)}, 2023.

\bibitem[Lu et~al.(2024)Lu, Jiang, Chen, Wang, Choi, and Lin]{lu2024wildvision}
Yujie Lu, Dongfu Jiang, Wenhu Chen, William~Yang Wang, Yejin Choi, and Bill~Yuchen Lin.
\newblock Wildvision: Evaluating vision-language models in the wild with human preferences.
\newblock \emph{arXiv preprint arXiv:2406.11069}, 2024.

\bibitem[Maaz et~al.(2024)Maaz, Rasheed, Shaker, Khan, Cholakal, Anwer, Baldwin, Felsberg, and Khan]{maaz2024palo}
Muhammad Maaz, Hanoona Rasheed, Abdelrahman Shaker, Salman Khan, Hisham Cholakal, Rao~M Anwer, Tim Baldwin, Michael Felsberg, and Fahad~S Khan.
\newblock Palo: A polyglot large multimodal model for 5b people.
\newblock \emph{arXiv preprint arXiv:2402.14818}, 2024.

\bibitem[Marafioti et~al.(2025)Marafioti, Zohar, Farr{\'e}, Noyan, Bakouch, Cuenca, Zakka, Allal, Lozhkov, Tazi, et~al.]{marafioti2025smolvlm}
Andr{\'e}s Marafioti, Orr Zohar, Miquel Farr{\'e}, Merve Noyan, Elie Bakouch, Pedro Cuenca, Cyril Zakka, Loubna~Ben Allal, Anton Lozhkov, Nouamane Tazi, et~al.
\newblock Smolvlm: Redefining small and efficient multimodal models.
\newblock \emph{arXiv preprint arXiv:2504.05299}, 2025.

\bibitem[Marino et~al.(2019)Marino, Rastegari, Farhadi, and Mottaghi]{marino2019ok}
Kenneth Marino, Mohammad Rastegari, Ali Farhadi, and Roozbeh Mottaghi.
\newblock Ok-vqa: A visual question answering benchmark requiring external knowledge.
\newblock In \emph{Proceedings of the IEEE/cvf conference on computer vision and pattern recognition}, pp.\  3195--3204, 2019.

\bibitem[Masry et~al.(2022)Masry, Long, Tan, Joty, and Hoque]{masry2022chartqa}
Ahmed Masry, Do~Xuan Long, Jia~Qing Tan, Shafiq Joty, and Enamul Hoque.
\newblock Chartqa: A benchmark for question answering about charts with visual and logical reasoning.
\newblock \emph{arXiv preprint arXiv:2203.10244}, 2022.

\bibitem[Matena \& Raffel(2022)Matena and Raffel]{matena2022merging}
Michael~S Matena and Colin~A Raffel.
\newblock Merging models with fisher-weighted averaging.
\newblock \emph{Advances in Neural Information Processing Systems}, 35:\penalty0 17703--17716, 2022.

\bibitem[Mathew et~al.(2021)Mathew, Karatzas, and Jawahar]{DocVQA}
Minesh Mathew, Dimosthenis Karatzas, and C.~V. Jawahar.
\newblock Docvqa: A dataset for vqa on document images.
\newblock In \emph{2021 IEEE Winter Conference on Applications of Computer Vision (WACV)}, pp.\  2199--2208, 2021.
\newblock \doi{10.1109/WACV48630.2021.00225}.

\bibitem[McCarthy \& Jarvis(2010)McCarthy and Jarvis]{mccarthy2010mtld}
Philip~M. McCarthy and Scott Jarvis.
\newblock Mtld, vocd-d, and hd-d: A validation study of sophisticated approaches to lexical diversity assessment.
\newblock \emph{Behavior Research Methods}, 42\penalty0 (2):\penalty0 381--392, 2010.
\newblock \doi{10.3758/BRM.42.2.381}.

\bibitem[McKinzie et~al.(2024)McKinzie, Gan, Fauconnier, Dodge, Zhang, Dufter, Shah, Du, Peng, Belyi, et~al.]{mckinzie2024mm1}
Brandon McKinzie, Zhe Gan, Jean-Philippe Fauconnier, Sam Dodge, Bowen Zhang, Philipp Dufter, Dhruti Shah, Xianzhi Du, Futang Peng, Anton Belyi, et~al.
\newblock Mm1: methods, analysis and insights from multimodal llm pre-training.
\newblock In \emph{European Conference on Computer Vision}, pp.\  304--323. Springer, 2024.

\bibitem[Muennighoff et~al.(2022)Muennighoff, Scao, Wang, Schmid, Bawden, Fan, Chaudhary, Gall{\'e}, et~al.]{muennighoff2022crosslingual}
Niklas Muennighoff, Teven~Le Scao, Yacine~Jernite Wang, Philipp Schmid, Rachel Bawden, Angela Fan, Vishrav Chaudhary, Matthias Gall{\'e}, et~al.
\newblock Crosslingual generalization through multitask finetuning.
\newblock In \emph{Proceedings of the 2022 Conference on Empirical Methods in Natural Language Processing (EMNLP)}, 2022.

\bibitem[Nguyen et~al.(2024)Nguyen, Chaudhary, Wang, Dabre, Elbayad, Fan, et~al.]{nguyen2024diverse}
Toan~Q Nguyen, Vishrav Chaudhary, Xian Wang, Raj Dabre, Maha Elbayad, Angela Fan, et~al.
\newblock Diverse multilingual pretraining for vision-language models.
\newblock \emph{arXiv preprint arXiv:2402.13673}, 2024.

\bibitem[Ni et~al.(2021)Ni, Huang, Su, Cui, Bharti, Wang, Zhang, and Duan]{Ni_2021_CVPR}
Minheng Ni, Haoyang Huang, Lin Su, Edward Cui, Taroon Bharti, Lijuan Wang, Dongdong Zhang, and Nan Duan.
\newblock M3p: Learning universal representations via multitask multilingual multimodal pre-training.
\newblock In \emph{Proceedings of the IEEE/CVF Conference on Computer Vision and Pattern Recognition (CVPR)}, pp.\  3977--3986, June 2021.

\bibitem[{OpenAI}(2024)]{openai2024gpt4o}
{OpenAI}.
\newblock Gpt-4o system card.
\newblock \url{https://arxiv.org/abs/2410.21276}, October 2024.
\newblock Accessed: 2025-04-17.

\bibitem[Ploeger et~al.(2024)Ploeger, Lai, van Noord, and Toral]{ploeger2024tailored}
Esther Ploeger, Huiyuan Lai, Rik van Noord, and Antonio Toral.
\newblock Towards tailored recovery of lexical diversity in literary machine translation.
\newblock \emph{arXiv preprint arXiv:2408.17308}, 2024.

\bibitem[Pont-Tuset et~al.(2020)Pont-Tuset, Uijlings, Changpinyo, Soricut, and Ferrari]{LocalizedNarratives}
Jordi Pont-Tuset, Jasper Uijlings, Soravit Changpinyo, Radu Soricut, and Vittorio Ferrari.
\newblock Connecting vision and language with localized narratives, 2020.

\bibitem[Pozzobon et~al.(2023)Pozzobon, Ermis, Lewis, and Hooker]{pozzobon2023goodtrieveradaptivetoxicitymitigation}
Luiza Pozzobon, Beyza Ermis, Patrick Lewis, and Sara Hooker.
\newblock Goodtriever: Adaptive toxicity mitigation with retrieval-augmented models, 2023.
\newblock URL \url{https://arxiv.org/abs/2310.07589}.

\bibitem[Pudjiati et~al.(2022)Pudjiati, Lustyantie, Iskandar, and Fitria]{pudjiati2022post}
Danti Pudjiati, Ninuk Lustyantie, Ifan Iskandar, and Tira~Nur Fitria.
\newblock Post-editing of machine translation: Creating a better translation of cultural specific terms.
\newblock \emph{Language Circle: Journal of Language and Literature}, 17\penalty0 (1):\penalty0 61--73, 2022.

\bibitem[Radford et~al.(2021)Radford, Kim, Hallacy, Ramesh, Goh, Agarwal, Sastry, Askell, Mishkin, Clark, et~al.]{radford2021learning}
Alec Radford, Jong~Wook Kim, Chris Hallacy, Aditya Ramesh, Gabriel Goh, Sandhini Agarwal, Girish Sastry, Amanda Askell, Pamela Mishkin, Jack Clark, et~al.
\newblock Learning transferable visual models from natural language supervision.
\newblock In \emph{International conference on machine learning}, pp.\  8748--8763. PmLR, 2021.

\bibitem[Ranaldi \& Pucci(2023)Ranaldi and Pucci]{ranaldi2023does}
Leonardo Ranaldi and Giulia Pucci.
\newblock Does the english matter? elicit cross-lingual abilities of large language models.
\newblock In \emph{Proceedings of the 3rd Workshop on Multi-lingual Representation Learning (MRL)}, pp.\  173--183, 2023.

\bibitem[Raunak et~al.(2023)Raunak, Sharaf, Wang, Awadallah, and Menezes]{raunak2023leveraging}
Vikas Raunak, Amr Sharaf, Yiren Wang, Hany~Hassan Awadallah, and Arul Menezes.
\newblock Leveraging gpt-4 for automatic translation post-editing.
\newblock \emph{arXiv preprint arXiv:2305.14878}, 2023.

\bibitem[Rei et~al.(2020)Rei, Stewart, Farinha, and Lavie]{rei2020comet}
Ricardo Rei, Craig Stewart, Ana~C Farinha, and Alon Lavie.
\newblock Comet: A neural framework for mt evaluation.
\newblock \emph{arXiv preprint arXiv:2009.09025}, 2020.

\bibitem[Rei et~al.(2023)Rei, Guerreiro, Pombal, van Stigt, Treviso, Coheur, de~Souza, and Martins]{rei2023scaling}
Ricardo Rei, Nuno~M Guerreiro, Jos{\'e} Pombal, Daan van Stigt, Marcos Treviso, Luisa Coheur, Jos{\'e}~GC de~Souza, and Andr{\'e}~FT Martins.
\newblock Scaling up cometkiwi: Unbabel-ist 2023 submission for the quality estimation shared task.
\newblock \emph{arXiv preprint arXiv:2309.11925}, 2023.

\bibitem[Rohrbach et~al.(2018)Rohrbach, Hendricks, Burns, Darrell, and Saenko]{rohrbach2018object}
Anna Rohrbach, Lisa~Anne Hendricks, Kaylee Burns, Trevor Darrell, and Kate Saenko.
\newblock Object hallucination in image captioning.
\newblock \emph{arXiv preprint arXiv:1809.02156}, 2018.

\bibitem[Romanou et~al.(2024)Romanou, Foroutan, Sotnikova, Chen, Nelaturu, Singh, Maheshwary, Altomare, Haggag, Amayuelas, et~al.]{romanou2024include}
Angelika Romanou, Negar Foroutan, Anna Sotnikova, Zeming Chen, Sree~Harsha Nelaturu, Shivalika Singh, Rishabh Maheshwary, Micol Altomare, Mohamed~A Haggag, Alfonso Amayuelas, et~al.
\newblock Include: Evaluating multilingual language understanding with regional knowledge.
\newblock \emph{arXiv preprint arXiv:2411.19799}, 2024.

\bibitem[Romero et~al.(2024)Romero, Lyu, Wibowo, Lynn, Hamed, Kishore, Mandal, Dragonetti, Abzaliev, Tonja, et~al.]{romero2024cvqa}
David Romero, Chenyang Lyu, Haryo~Akbarianto Wibowo, Teresa Lynn, Injy Hamed, Aditya~Nanda Kishore, Aishik Mandal, Alina Dragonetti, Artem Abzaliev, Atnafu~Lambebo Tonja, et~al.
\newblock Cvqa: Culturally-diverse multilingual visual question answering benchmark.
\newblock \emph{arXiv preprint arXiv:2406.05967}, 2024.

\bibitem[Salazar et~al.(2025)Salazar, Burda, Islam, Moakhar, Singh, Farestam, Romanou, Boiko, Khullar, Zhang, et~al.]{salazar2025kaleidoscope}
Israfel Salazar, Manuel~Fern{\'a}ndez Burda, Shayekh~Bin Islam, Arshia~Soltani Moakhar, Shivalika Singh, Fabian Farestam, Angelika Romanou, Danylo Boiko, Dipika Khullar, Mike Zhang, et~al.
\newblock Kaleidoscope: In-language exams for massively multilingual vision evaluation.
\newblock \emph{arXiv preprint arXiv:2504.07072}, 2025.

\bibitem[Savoldi et~al.(2021)Savoldi, Gaido, Bentivogli, Negri, and Turchi]{savoldi2021gender}
Beatrice Savoldi, Marco Gaido, Luisa Bentivogli, Matteo Negri, and Marco Turchi.
\newblock Gender bias in machine translation.
\newblock \emph{Transactions of the Association for Computational Linguistics}, 9:\penalty0 845--874, 2021.

\bibitem[Schwenk et~al.(2022)Schwenk, Khandelwal, Clark, Marino, and Mottaghi]{schwenk2022okvqa}
Dustin Schwenk, Apoorv Khandelwal, Christopher Clark, Kenneth Marino, and Roozbeh Mottaghi.
\newblock A-okvqa: A benchmark for visual question answering using world knowledge.
\newblock In \emph{European conference on computer vision}, pp.\  146--162. Springer, 2022.

\bibitem[Shaham et~al.(2024)Shaham, Efrat, Kwiatkowski, Gupta, Tran, Xiong, and Subramani]{shaham2024pinch}
Uri Shaham, Avia Efrat, Tom Kwiatkowski, Raghav Gupta, Chau Tran, Caiming Xiong, and Nishant Subramani.
\newblock Just a pinch of multilinguality improves instruction tuning.
\newblock In \emph{Findings of the Association for Computational Linguistics (ACL)}, 2024.

\bibitem[Shazeer(2020)]{shazeer2020glu}
Noam Shazeer.
\newblock Glu variants improve transformer.
\newblock \emph{arXiv preprint arXiv:2002.05202}, 2020.

\bibitem[Shen(2022)]{shen2022lexicalrichness}
Lucas Shen.
\newblock Lexicalrichness: A small module to compute textual lexical richness, 2022.
\newblock URL \url{https://github.com/LSYS/lexicalrichness}.

\bibitem[Shi et~al.(2022)Shi, Suzgun, Freitag, Wang, Srivats, Vosoughi, Chung, Tay, Ruder, Zhou, et~al.]{shi2022language}
Freda Shi, Mirac Suzgun, Markus Freitag, Xuezhi Wang, Suraj Srivats, Soroush Vosoughi, Hyung~Won Chung, Yi~Tay, Sebastian Ruder, Denny Zhou, et~al.
\newblock Language models are multilingual chain-of-thought reasoners.
\newblock \emph{arXiv preprint arXiv:2210.03057}, 2022.

\bibitem[Singh et~al.(2019)Singh, Natarajan, Shah, Jiang, Chen, Batra, Parikh, and Rohrbach]{singh2019towards}
Amanpreet Singh, Vivek Natarajan, Meet Shah, Yu~Jiang, Xinlei Chen, Dhruv Batra, Devi Parikh, and Marcus Rohrbach.
\newblock Towards vqa models that can read.
\newblock In \emph{Proceedings of the IEEE/CVF conference on computer vision and pattern recognition}, pp.\  8317--8326, 2019.

\bibitem[Singh et~al.(2024{\natexlab{a}})Singh, Romanou, Fourrier, Adelani, Ngui, Vila-Suero, Limkonchotiwat, Marchisio, Leong, Susanto, et~al.]{singh2024global}
Shivalika Singh, Angelika Romanou, Cl{\'e}mentine Fourrier, David~I Adelani, Jian~Gang Ngui, Daniel Vila-Suero, Peerat Limkonchotiwat, Kelly Marchisio, Wei~Qi Leong, Yosephine Susanto, et~al.
\newblock Global mmlu: Understanding and addressing cultural and linguistic biases in multilingual evaluation.
\newblock \emph{arXiv preprint arXiv:2412.03304}, 2024{\natexlab{a}}.

\bibitem[Singh et~al.(2024{\natexlab{b}})Singh, Vargus, Dsouza, Karlsson, Mahendiran, Ko, Shandilya, Patel, Mataciunas, OMahony, Zhang, Hettiarachchi, Wilson, Machado, Moura, Krzemiński, Fadaei, Ergün, Okoh, Alaagib, Mudannayake, Alyafeai, Chien, Ruder, Guthikonda, Alghamdi, Gehrmann, Muennighoff, Bartolo, Kreutzer, Üstün, Fadaee, and Hooker]{singh2024aya}
Shivalika Singh, Freddie Vargus, Daniel Dsouza, Börje~F. Karlsson, Abinaya Mahendiran, Wei-Yin Ko, Herumb Shandilya, Jay Patel, Deividas Mataciunas, Laura OMahony, Mike Zhang, Ramith Hettiarachchi, Joseph Wilson, Marina Machado, Luisa~Souza Moura, Dominik Krzemiński, Hakimeh Fadaei, Irem Ergün, Ifeoma Okoh, Aisha Alaagib, Oshan Mudannayake, Zaid Alyafeai, Vu~Minh Chien, Sebastian Ruder, Surya Guthikonda, Emad~A. Alghamdi, Sebastian Gehrmann, Niklas Muennighoff, Max Bartolo, Julia Kreutzer, Ahmet Üstün, Marzieh Fadaee, and Sara Hooker.
\newblock Aya dataset: An open-access collection for multilingual instruction tuning, 2024{\natexlab{b}}.

\bibitem[Song et~al.(2025)Song, Ou, Kong, Li, Neubig, and Yue]{song2025visualpuzzles}
Yueqi Song, Tianyue Ou, Yibo Kong, Zecheng Li, Graham Neubig, and Xiang Yue.
\newblock Visualpuzzles: Decoupling multimodal reasoning evaluation from domain knowledge.
\newblock \emph{arXiv preprint arXiv:2504.10342}, 2025.

\bibitem[Tanaka et~al.(2023)Tanaka, Nishida, Nishida, Hasegawa, Saito, and Saito]{tanaka2023slidevqa}
Ryota Tanaka, Kyosuke Nishida, Kosuke Nishida, Taku Hasegawa, Itsumi Saito, and Kuniko Saito.
\newblock Slidevqa: A dataset for document visual question answering on multiple images.
\newblock \emph{arXiv preprint arXiv:2301.04883}, 2023.

\bibitem[Tang et~al.(2024)Tang, Liu, Ye, Lu, Wei, Lin, Li, Mahmood, Feng, Zhao, et~al.]{tang2024mtvqa}
Jingqun Tang, Qi~Liu, Yongjie Ye, Jinghui Lu, Shu Wei, Chunhui Lin, Wanqing Li, Mohamad Fitri Faiz~Bin Mahmood, Hao Feng, Zhen Zhao, et~al.
\newblock Mtvqa: Benchmarking multilingual text-centric visual question answering.
\newblock \emph{arXiv preprint arXiv:2405.11985}, 2024.

\bibitem[Team(2024{\natexlab{a}})]{team2024chameleon}
Chameleon Team.
\newblock Chameleon: Mixed-modal early-fusion foundation models.
\newblock \emph{arXiv preprint arXiv:2405.09818}, 2024{\natexlab{a}}.

\bibitem[Team(2024{\natexlab{b}})]{geminiteam2024gemini15unlockingmultimodal}
Gemini Team.
\newblock Gemini 1.5: Unlocking multimodal understanding across millions of tokens of context, 2024{\natexlab{b}}.
\newblock URL \url{https://arxiv.org/abs/2403.05530}.

\bibitem[Team et~al.(2024)Team, Georgiev, Lei, Burnell, Bai, Gulati, Tanzer, Vincent, Pan, Wang, et~al.]{team2024gemini}
Gemini Team, Petko Georgiev, Ving~Ian Lei, Ryan Burnell, Libin Bai, Anmol Gulati, Garrett Tanzer, Damien Vincent, Zhufeng Pan, Shibo Wang, et~al.
\newblock Gemini 1.5: Unlocking multimodal understanding across millions of tokens of context.
\newblock \emph{arXiv preprint arXiv:2403.05530}, 2024.

\bibitem[Team et~al.(2025)Team, Du, Gao, Xing, Jiang, Chen, Li, Xiao, Du, Liao, Tang, Wang, Zhang, Yuan, Lu, Tang, Sung, Wei, Lai, Guo, Zhu, Ding, Hu, Yang, Zhang, Yao, Zhao, Lu, Li, Yu, Gao, Zheng, Yuan, Chen, Guo, Su, Wang, Zhao, Zhang, Liu, Yan, Wu, Shi, Ye, Yu, Dong, Zhang, Ma, Pan, Gong, Liu, Ma, Wei, Cao, Huang, Jiang, Gao, Xiong, He, Huang, Wu, He, Wei, Jia, Wu, Xu, Zu, Zhou, Pan, Charles, Li, Hu, Liu, Chen, Wang, Liu, Qin, Liu, Yang, Bao, Du, Wu, Wang, Zhou, Wang, Li, Zhu, Zhang, Wang, Yang, Huang, Huang, Xu, and Yang]{kimiteam2025kimik15scalingreinforcement}
Kimi Team, Angang Du, Bofei Gao, Bowei Xing, Changjiu Jiang, Cheng Chen, Cheng Li, Chenjun Xiao, Chenzhuang Du, Chonghua Liao, Chuning Tang, Congcong Wang, Dehao Zhang, Enming Yuan, Enzhe Lu, Fengxiang Tang, Flood Sung, Guangda Wei, Guokun Lai, Haiqing Guo, Han Zhu, Hao Ding, Hao Hu, Hao Yang, Hao Zhang, Haotian Yao, Haotian Zhao, Haoyu Lu, Haoze Li, Haozhen Yu, Hongcheng Gao, Huabin Zheng, Huan Yuan, Jia Chen, Jianhang Guo, Jianlin Su, Jianzhou Wang, Jie Zhao, Jin Zhang, Jingyuan Liu, Junjie Yan, Junyan Wu, Lidong Shi, Ling Ye, Longhui Yu, Mengnan Dong, Neo Zhang, Ningchen Ma, Qiwei Pan, Qucheng Gong, Shaowei Liu, Shengling Ma, Shupeng Wei, Sihan Cao, Siying Huang, Tao Jiang, Weihao Gao, Weimin Xiong, Weiran He, Weixiao Huang, Wenhao Wu, Wenyang He, Xianghui Wei, Xianqing Jia, Xingzhe Wu, Xinran Xu, Xinxing Zu, Xinyu Zhou, Xuehai Pan, Y.~Charles, Yang Li, Yangyang Hu, Yangyang Liu, Yanru Chen, Yejie Wang, Yibo Liu, Yidao Qin, Yifeng Liu, Ying Yang, Yiping Bao, Yulun Du, Yuxin Wu, Yuzhi Wang, Zaida Zhou,
  Zhaoji Wang, Zhaowei Li, Zhen Zhu, Zheng Zhang, Zhexu Wang, Zhilin Yang, Zhiqi Huang, Zihao Huang, Ziyao Xu, and Zonghan Yang.
\newblock Kimi k1.5: Scaling reinforcement learning with llms, 2025.
\newblock URL \url{https://arxiv.org/abs/2501.12599}.

\bibitem[Tong et~al.(2024)Tong, Brown, Wu, Woo, IYER, Akula, Yang, Yang, Middepogu, Wang, et~al.]{tong2024cambrian}
Peter Tong, Ellis Brown, Penghao Wu, Sanghyun Woo, Adithya Jairam~Vedagiri IYER, Sai~Charitha Akula, Shusheng Yang, Jihan Yang, Manoj Middepogu, Ziteng Wang, et~al.
\newblock Cambrian-1: A fully open, vision-centric exploration of multimodal llms.
\newblock \emph{Advances in Neural Information Processing Systems}, 37:\penalty0 87310--87356, 2024.

\bibitem[Tschannen et~al.(2025)Tschannen, Gritsenko, Wang, Naeem, Alabdulmohsin, Parthasarathy, Evans, Beyer, Xia, Mustafa, et~al.]{tschannen2025siglip}
Michael Tschannen, Alexey Gritsenko, Xiao Wang, Muhammad~Ferjad Naeem, Ibrahim Alabdulmohsin, Nikhil Parthasarathy, Talfan Evans, Lucas Beyer, Ye~Xia, Basil Mustafa, et~al.
\newblock Siglip 2: Multilingual vision-language encoders with improved semantic understanding, localization, and dense features.
\newblock \emph{arXiv preprint arXiv:2502.14786}, 2025.

\bibitem[{\"U}st{\"u}n et~al.(2024){\"U}st{\"u}n, Aryabumi, Yong, Ko, D'souza, Onilude, Bhandari, Singh, Ooi, Kayid, et~al.]{ustun2024aya}
Ahmet {\"U}st{\"u}n, Viraat Aryabumi, Zheng-Xin Yong, Wei-Yin Ko, Daniel D'souza, Gbemileke Onilude, Neel Bhandari, Shivalika Singh, Hui-Lee Ooi, Amr Kayid, et~al.
\newblock Aya model: An instruction finetuned open-access multilingual language model.
\newblock \emph{arXiv preprint arXiv:2402.07827}, 2024.

\bibitem[Vanmassenhove et~al.(2021)Vanmassenhove, Shterionov, and Gwilliam]{vanmassenhove2021machine}
Eva Vanmassenhove, Dimitar Shterionov, and Matthew Gwilliam.
\newblock Machine translationese: Effects of algorithmic bias on linguistic complexity in machine translation.
\newblock \emph{arXiv preprint arXiv:2102.00287}, 2021.

\bibitem[Wang et~al.(2021)Wang, Li, Zhou, Chen, Grossman, and Li]{wang2021screen2words}
Bryan Wang, Gang Li, Xin Zhou, Zhourong Chen, Tovi Grossman, and Yang Li.
\newblock Screen2words: Automatic mobile ui summarization with multimodal learning.
\newblock In \emph{The 34th Annual ACM Symposium on User Interface Software and Technology}, pp.\  498--510, 2021.

\bibitem[Wang et~al.(2024{\natexlab{a}})Wang, Zhou, Huang, Xu, Zhang, Poon, and Chen]{wang2024mdpo}
Fei Wang, Wenxuan Zhou, James~Y Huang, Nan Xu, Sheng Zhang, Hoifung Poon, and Muhao Chen.
\newblock mdpo: Conditional preference optimization for multimodal large language models.
\newblock \emph{arXiv preprint arXiv:2406.11839}, 2024{\natexlab{a}}.

\bibitem[Wang et~al.(2022)Wang, Rubinstein, and Cohn]{wang2022measuring}
Jun Wang, Benjamin Rubinstein, and Trevor Cohn.
\newblock Measuring and mitigating name biases in neural machine translation.
\newblock In \emph{Proceedings of the 60th Annual Meeting of the Association for Computational Linguistics (Volume 1: Long Papers)}, pp.\  2576--2590, 2022.

\bibitem[Wang et~al.(2024{\natexlab{b}})Wang, Bai, Tan, Wang, Fan, Bai, Chen, Liu, Wang, Ge, et~al.]{wang2024qwen2}
Peng Wang, Shuai Bai, Sinan Tan, Shijie Wang, Zhihao Fan, Jinze Bai, Keqin Chen, Xuejing Liu, Jialin Wang, Wenbin Ge, et~al.
\newblock Qwen2-vl: Enhancing vision-language model's perception of the world at any resolution.
\newblock \emph{arXiv preprint arXiv:2409.12191}, 2024{\natexlab{b}}.

\bibitem[Wang et~al.(2024{\natexlab{c}})Wang, Xia, He, Chen, Liu, Zhu, Liang, Wu, Liu, Malladi, et~al.]{wang2024charxiv}
Zirui Wang, Mengzhou Xia, Luxi He, Howard Chen, Yitao Liu, Richard Zhu, Kaiqu Liang, Xindi Wu, Haotian Liu, Sadhika Malladi, et~al.
\newblock Charxiv: Charting gaps in realistic chart understanding in multimodal llms.
\newblock \emph{Advances in Neural Information Processing Systems}, 37:\penalty0 113569--113697, 2024{\natexlab{c}}.

\bibitem[Wendler(2023)]{RenderedText}
Christoph Wendler.
\newblock wendlerc/renderedtext, 2023.
\newblock URL \url{https://huggingface.co/datasets/wendlerc/RenderedText}.

\bibitem[Wortsman et~al.(2022)Wortsman, Ilharco, Gadre, Roelofs, Gontijo-Lopes, Morcos, Namkoong, Farhadi, Carmon, Kornblith, et~al.]{wortsman2022model}
Mitchell Wortsman, Gabriel Ilharco, Samir~Ya Gadre, Rebecca Roelofs, Raphael Gontijo-Lopes, Ari~S Morcos, Hongseok Namkoong, Ali Farhadi, Yair Carmon, Simon Kornblith, et~al.
\newblock Model soups: averaging weights of multiple fine-tuned models improves accuracy without increasing inference time.
\newblock In \emph{International conference on machine learning}, pp.\  23965--23998. PMLR, 2022.

\bibitem[xAI(2024)]{xai2024realworldqa}
xAI.
\newblock Realworldqa dataset, 2024.
\newblock URL \url{https://huggingface.co/datasets/xai-org/RealworldQA}.
\newblock Accessed on May 4, 2025.

\bibitem[Yadav et~al.(2023)Yadav, Tam, Choshen, Raffel, and Bansal]{yadav2023ties}
Prateek Yadav, Derek Tam, Leshem Choshen, Colin~A Raffel, and Mohit Bansal.
\newblock Ties-merging: Resolving interference when merging models.
\newblock \emph{Advances in Neural Information Processing Systems}, 36:\penalty0 7093--7115, 2023.

\bibitem[Yasunaga et~al.(2025)Yasunaga, Zettlemoyer, and Ghazvininejad]{yasunaga2025multimodal}
Michihiro Yasunaga, Luke Zettlemoyer, and Marjan Ghazvininejad.
\newblock Multimodal rewardbench: Holistic evaluation of reward models for vision language models, 2025.
\newblock URL \url{https://arxiv.org/abs/2502.14191}.

\bibitem[Yue et~al.(2024{\natexlab{a}})Yue, Ni, Zhang, Zheng, Liu, Zhang, Stevens, Jiang, Ren, Sun, et~al.]{yue2024mmmu}
Xiang Yue, Yuansheng Ni, Kai Zhang, Tianyu Zheng, Ruoqi Liu, Ge~Zhang, Samuel Stevens, Dongfu Jiang, Weiming Ren, Yuxuan Sun, et~al.
\newblock Mmmu: A massive multi-discipline multimodal understanding and reasoning benchmark for expert agi.
\newblock In \emph{Proceedings of the IEEE/CVF Conference on Computer Vision and Pattern Recognition}, pp.\  9556--9567, 2024{\natexlab{a}}.

\bibitem[Yue et~al.(2024{\natexlab{b}})Yue, Song, Asai, Khanuja, Kantharuban, Kim, de~Dieu~Nyandwi, Sutawika, Ramamoorthy, and Neubig]{yue2024pangea}
Xiang Yue, Yueqi Song, Akari Asai, Simran Khanuja, Anjali Kantharuban, Seungone Kim, Jean de~Dieu~Nyandwi, Lintang Sutawika, Sathyanarayanan Ramamoorthy, and Graham Neubig.
\newblock Pangea: A fully open multilingual multimodal llm for 39 languages.
\newblock In \emph{The Thirteenth International Conference on Learning Representations}, 2024{\natexlab{b}}.

\bibitem[Zadouri et~al.(2023)Zadouri, {\"U}st{\"u}n, Ahmadian, Ermi{\c{s}}, Locatelli, and Hooker]{zadouri2023pushing}
Ted Zadouri, Ahmet {\"U}st{\"u}n, Arash Ahmadian, Beyza Ermi{\c{s}}, Acyr Locatelli, and Sara Hooker.
\newblock Pushing mixture of experts to the limit: Extremely parameter efficient moe for instruction tuning.
\newblock \emph{arXiv preprint arXiv:2309.05444}, 2023.

\bibitem[Zeng et~al.(2023)Zeng, Zhou, Luo, Cheng, and Zhang]{zeng2023crossviewlanguagemodelingunified}
Yan Zeng, Wangchunshu Zhou, Ao~Luo, Ziming Cheng, and Xinsong Zhang.
\newblock Cross-view language modeling: Towards unified cross-lingual cross-modal pre-training, 2023.
\newblock URL \url{https://arxiv.org/abs/2206.00621}.

\bibitem[Zhai et~al.(2023)Zhai, Mustafa, Kolesnikov, and Beyer]{zhai2023sigmoid}
Xiaohua Zhai, Basil Mustafa, Alexander Kolesnikov, and Lucas Beyer.
\newblock Sigmoid loss for language image pre-training.
\newblock In \emph{Proceedings of the IEEE/CVF international conference on computer vision}, pp.\  11975--11986, 2023.

\bibitem[Zhao et~al.(2022)Zhao, Li, Li, and Zhang]{Multihiertt}
Yilun Zhao, Yunxiang Li, Chenying Li, and Rui Zhang.
\newblock {M}ulti{H}iertt: Numerical reasoning over multi hierarchical tabular and textual data.
\newblock In \emph{Proceedings of the 60th Annual Meeting of the Association for Computational Linguistics (Volume 1: Long Papers)}, pp.\  6588--6600, Dublin, Ireland, May 2022. Association for Computational Linguistics.
\newblock URL \url{https://aclanthology.org/2022.acl-long.454}.

\bibitem[Zhou et~al.(2023)Zhou, Lu, Mishra, Brahma, Basu, Luan, Zhou, and Hou]{zhou2023instruction}
Jeffrey Zhou, Tianjian Lu, Swaroop Mishra, Siddhartha Brahma, Sujoy Basu, Yi~Luan, Denny Zhou, and Le~Hou.
\newblock Instruction-following evaluation for large language models.
\newblock \emph{arXiv preprint arXiv:2311.07911}, 2023.

\bibitem[Zhu et~al.(2025)Zhu, Song, Shen, Zhao, Yang, Zhang, and Wu]{zhu2025remedy}
Didi Zhu, Yibing Song, Tao Shen, Ziyu Zhao, Jinluan Yang, Min Zhang, and Chao Wu.
\newblock Remedy: Recipe merging dynamics in large vision-language models.
\newblock In \emph{The Thirteenth International Conference on Learning Representations}, 2025.

\bibitem[Zhu et~al.(2021)Zhu, Lei, Huang, Wang, Zhang, Lv, Feng, and Chua]{TAT-QA}
Fengbin Zhu, Wenqiang Lei, Youcheng Huang, Chao Wang, Shuo Zhang, Jiancheng Lv, Fuli Feng, and Tat-Seng Chua.
\newblock {TAT}-{QA}: A question answering benchmark on a hybrid of tabular and textual content in finance.
\newblock In \emph{Proceedings of the 59th Annual Meeting of the Association for Computational Linguistics and the 11th International Joint Conference on Natural Language Processing (Volume 1: Long Papers)}, pp.\  3277--3287, Online, August 2021. Association for Computational Linguistics.
\newblock \doi{10.18653/v1/2021.acl-long.254}.
\newblock URL \url{https://aclanthology.org/2021.acl-long.254}.

\bibitem[Zhu et~al.(2023)Zhu, Liu, Dong, Xu, Huang, Kong, Chen, and Li]{zhu2023multilingual}
Wenhao Zhu, Hongyi Liu, Qingxiu Dong, Jingjing Xu, Shujian Huang, Lingpeng Kong, Jiajun Chen, and Lei Li.
\newblock Multilingual machine translation with large language models: Empirical results and analysis.
\newblock \emph{arXiv preprint arXiv:2304.04675}, 2023.

\end{thebibliography}

\newpage
\appendix
\section{Evaluation Details}
\label{app:eval-details}

\subsection{AyaVisionBench}
\label{app:ayavisionbench}

To create this dataset, we first sourced images from the test splits of various datasets included in Cauldron \citep{laurenccon2024matters}, a large-scale collection of 50 high-quality vision datasets. By exclusively selecting images from the test splits, we ensured that none had been seen during model training. 
Following the original task categories defined in Cauldron, we randomly sampled 15 images from each of 9 distinct tasks, resulting in a total of 135 unseen images. For each image, we generated a corresponding question that required explicit visual understanding to answer. These questions were initially generated synthetically and then manually reviewed for clarity, relevance, and dependence on the visual content.

\begin{figure}[H]
    \centering
    \makebox[\textwidth][c]{
        \includegraphics[width=\linewidth]{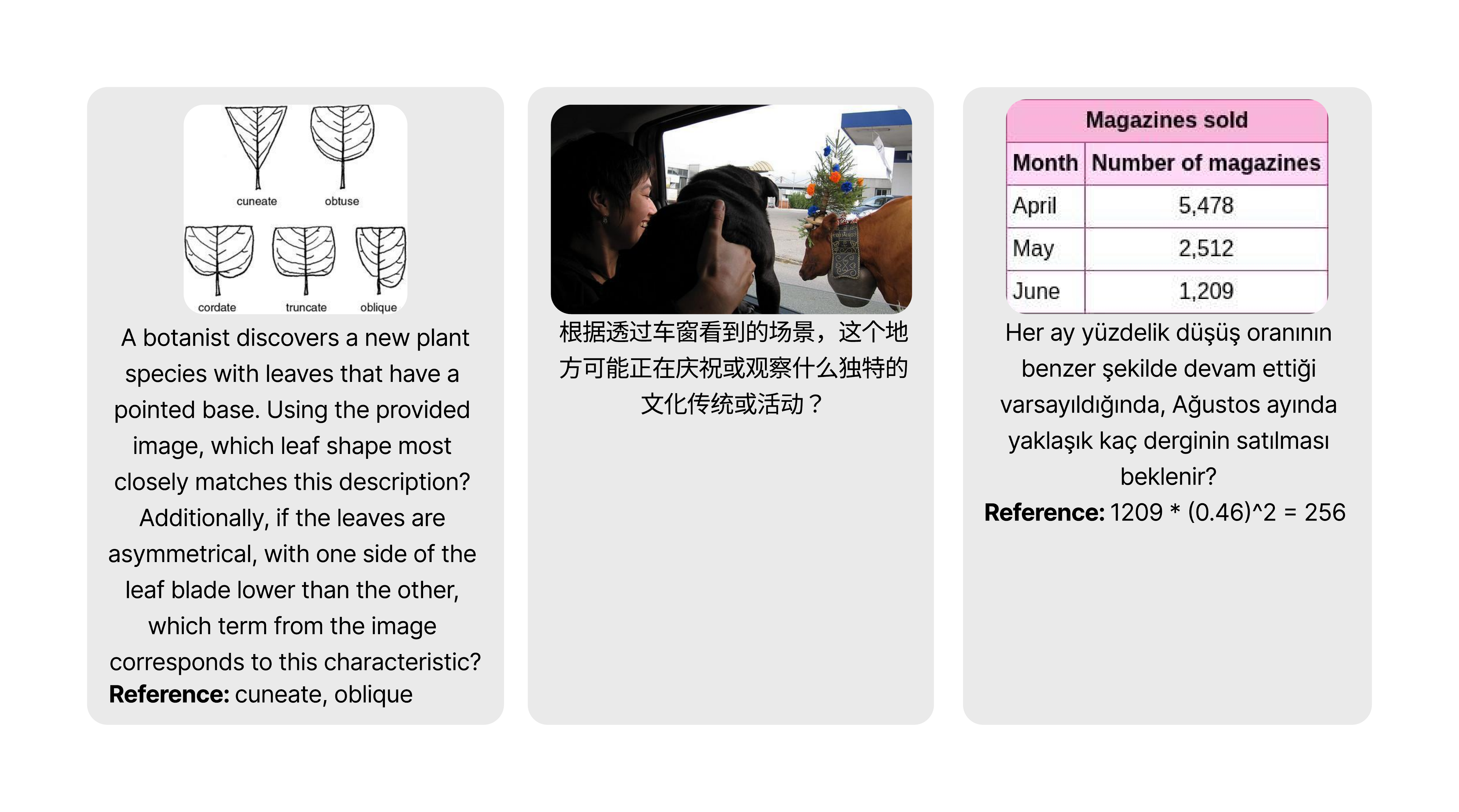}
    }
    \caption{\textbf{Three sample entries from AyaVisionBench, illustrating a range of languages and image task types.} From left to right: English (TQA \citep{TQA}), Chinese (VSR \citep{liu2023vsr}), and Turkish (TabMWP \citep{TabMWP}). All images are sourced from the test sets of the respective datasets.}
    \label{fig:ayavisionbench_examples}
\end{figure}

To support multilingual evaluation, each question was translated into 22 additional languages using Google Translate\footnote{\url{https://cloud.google.com/translate?hl=en}}, covering all 23 languages supported by Aya Vision. All translations were subsequently verified by human annotators to ensure fidelity and naturalness.
During human annotation, annotators were also asked to validate the prompts and provide reference answers for questions with deterministic answers. These reference answers are included in the benchmark.
The resulting dataset, \textbf{AyaVisionBench}, offers a diverse and challenging benchmark for evaluating vision-language models in multilingual and open-ended contexts. Representative examples are shown in \Cref{fig:ayavisionbench_examples}.

\subsection{Multimodal Acedemic Benchmarks}
\begin{itemize}
    \item \textbf{xMMMU} \citep{yue2024pangea}, a machine-translated version of 300 questions from the MMMU validation set into 6 languages to measure the multimodal understanding and reasoning.
    \item \textbf{MaXM} \citep{changpinyo2022maxm} evaluates vision-language models on multilingual VQA tasks in 7 languages.
    \item \textbf{CVQA} \citep{romero2024cvqa} is a large-scale, multilingual VQA dataset to test models' understanding of cultural nuances in 31 languages.
    \item \textbf{MTVQA} \citep{tang2024mtvqa} evaluates multilingual multimodal models on text-centric scene understanding in 9 languages.
    \item \textbf{Kaleidoscope} \citep{salazar2025kaleidoscope} consists of 20,911 multimodal multiple-choice questions in 18 languages, designed to evaluate the reasoning and knowledge of vision-language models across diverse subjects and cultures.
\end{itemize}

\subsection{Text-Only Benchmarks}
\begin{itemize}
    \item \textbf{m-ArenaHard} \citep{li2024crowdsourced} following \citep{dang2024aya}, we use multilingual ArenaHard to measure the win-rates against other models across 23 languages to understand the impact of multimodal training on the model's text-only capabilities. We use \texttt{gpt-4o-2024-11-20} \citep{openai2024gpt4o} as the judge.
    \item \textbf{MGSM} \citep{shi2022language} evaluates the reasoning abilities of large language models with 250 grade-school math problems in 10 languages 
    \item \textbf{Global MMLU-Lite} \citep{singh2024global} is a multilingual MMLU test set spanning 42 languages 
    \item \textbf{FLORES} \citep{guzmán2019flores} is an evaluation benchmark for machine translation in low-resource languages. 
    \item \textbf{IFEval} \citep{zhou2023instruction} is a benchmark designed to assess the ability of large language models to follow verifiable instructions.
\end{itemize}

\section{Training Hyerparameters}
{\small
\definecolor{highlightcolor}{RGB}{211, 227, 252}
\renewcommand{\arraystretch}{1.2}
\setlength{\tabcolsep}{8pt}
\begin{longtable}{lcc}
\caption{Training Hyper-parameters for Aya Vision-8B and Aya Vision-32B models}
\label{tab:train-hyperparams} \\
\toprule
 \textbf{Aya Vision} & \textbf{8B} & \textbf{32B} \\
\midrule
\rowcolor{highlightcolor}\multicolumn{3}{l}{\textbf{Vision Encoder}} \\
Params      & 400M       &  400M\\
Dim         & 1152       &  1152\\
MLP Dim     & 4304       &  4304 \\
Act.        & GELU       &  GELU\\
Heads       & 16         &  16\\
KV Heads    & 16         &  16\\
Layers      & 27         &  27\\
Image Size  & 364$\times$364 & 512$\times$512 \\
Patch Size  & 14         &  16 \\
\midrule
\rowcolor{highlightcolor}\multicolumn{3}{l}{\textbf{Vision-Language Connector}} \\
Params      & 190M       & 428M \\
Downsample Factor   & 2  & 2 \\
MLP Dim     & 14336      & 24676 \\
Act.        & SwiGLU     & SwiGLU \\
\midrule
\rowcolor{highlightcolor}\multicolumn{3}{l}{\textbf{LLM}} \\
Params      & 8B       & 32.3B \\
Embed       & 256k     & 256k \\
Dim         & 4096       & 8192 \\
MLP Dim     & 14336      & 24676 \\
Act.        & SwiGLU     & SwiGLU \\
Heads       & 32         & 64 \\
KV Heads    & 8          & 8 \\
Layers      & 32         & 40 \\
Theta       & 50k         & 4M \\
\midrule
\rowcolor{highlightcolor}\multicolumn{3}{l}{\textbf{Alignment}} \\
Warmup  & 200       & 200 \\
Peak LR      & 1e-4       & 1e-3 \\
Cosine Decay& 10\%       & 10\% \\
Optimizer & AdamW        & AdamW \\
Betas       & 0.9, 0.95  & 0.9, 0.95 \\
Batch Size  & 128        & 128 \\
Steps       & 9.7k      & 19k \\
\midrule
\rowcolor{highlightcolor}\multicolumn{3}{l}{\textbf{SFT}} \\
Warmup LLM  & 200        & 200 \\
Peak LR      & 1e-4       & 5e-4 \\
Cosine Decay& 10\%       & 10\% \\
Betas       & 0.9, 0.95  & 0.9, 0.95 \\
Batch Size  & 128        & 128 \\
Steps       & 31k        & 31k \\
\bottomrule
\end{longtable}
}

\section{Additional Ablations}
\label{app:add-ablations}
\subsection{Stronger Vision Encoder Improves VQA Performance}
\label{sec:siglip}
\begin{figure}[H]
    \centering
    \includegraphics[width=0.3\linewidth]{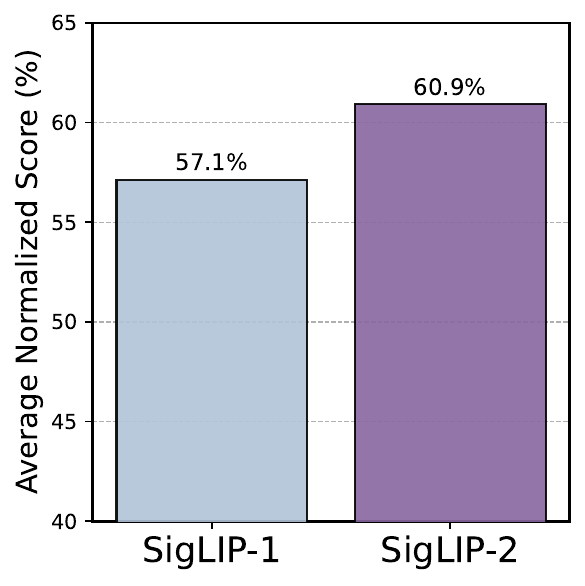}
    \caption{\textbf{Improvement by switching to SigLIP-2.} We report the average of VQA evaluations listed in \S~\ref{sec:siglip}.}
    \label{fig:siglip-ablation}
\end{figure}

With the recent releases of better vision encoders, we ask \textit{how do these gains translate to downstream multimodal performance?} We design an experiment by training a variant of Aya Vision-8B with the original SigLIP encoder instead of SigLIP-2 with the same resolution and patch size. Interestingly, we observe no visible impact on the multimodal win-rates; however, switching to SigLIP-2 provides substantial improvements in multimodal academic benchmarks like CVQA\citep{romero2024cvqa}, TextVQA \citep{singh2019towards}, DocVQA \citep{DocVQA}, ChartQA \citep{masry2022chartqa}, OKVQA \citep{marino2019ok} and RealWorldQA \citep{xai2024realworldqa} -- with an average improvement of $4\%$ as shown in Figure \ref{fig:siglip-ablation}. 

\newpage

\section{Recaptioning Templates}
\label{app:recaption-templates}

\definecolor{highlightcolor}{RGB}{240, 243, 250}
\definecolor{framecolor}{RGB}{83, 109, 222}
\begin{tcolorbox}[
  colback=highlightcolor!90, 
  colframe=framecolor, 
  boxrule=0.8pt, 
  title=General Visual Question Answering,
  label=recaption-template-vqa
]
\textbf{System Prompt:} \\
You are an advanced multimodal AI chatbot with strong visual question answering capabilities. \\
\textbf{User Prompt: } \\
Here is a question-answer pair for the given image: \\
\textit{Question:} \\
\{instruction\} \\
\textit{Reference Answer:} \\
\{answer\} \\ 
\textit{Task Description: }\\
Analyze all provided image and fully understand the question, paying attention to every detail and context within the image. \\
The reference answer is the correct answer to the question. \\
Your task is to generate a more comprehensive, natural and human-preferred response to the question. \\
Enhance the response by adding additional visual context, mentioning relevant information, or providing detailed explanations. \\
If the question is multiple-choice, the response should mention the letter/number of the selected choice. \\
Also, ensure that the final result in the response is consistent with the reference answer. \\
But, do not explicitly mention there is a reference answer in the response. \\ 
The response should stand independently as a complete and well-organized new answer to the question. \\
\\
Enclose the new answer within <answer> </answer> tags.
\end{tcolorbox}

\begin{tcolorbox}[
  colback=highlightcolor!90, 
  colframe=framecolor, 
  boxrule=0.8pt, 
  title=Captioning,
  label=recaption-template-captioning
]
\textbf{System Prompt:} \\
You are an advanced multimodal AI chatbot with strong image captioning capabilities. \\
\textbf{User Prompt: } \\
Here is an image captioning instruction along with the original caption for the provided image. \\
\textit{Instruction:}\\
\{instruction\}\\
\textit{Original Caption: }\\
\{answer\}\\
\textit{Task Description:}\\
Examine the image carefully, paying attention to every detail and context within the image.\\
Your task is to rewrite the original caption to be more detailed, descriptive, comprehensive, and human-preferred. \\
Ensure that the new caption accurately reflects the content and context of the image while following the given instruction. \\
Since this is an image captioning task, do not include any information that is not directly visible in the image.\\
Do not explicitly mention there is an original caption in the response. \\
Ensure the response stands independently as a complete and well-organized new caption. \\

Enclose the new caption within <answer> </answer> tags.
\end{tcolorbox}

\begin{tcolorbox}[
  colback=highlightcolor!90, 
  colframe=framecolor, 
  boxrule=0.8pt, 
  title=\parbox{\linewidth}{OCR, document understanding, text transcription},
  label=recaption-template-ocr
]
\textbf{System Prompt:} \\
You are an advanced multimodal AI chatbot with strong text-rich image understanding capabilities. \\
\textbf{User Prompt: } \\
Here is a question-answer pair based on the provided document, screenshot or scanned image. \\
\textit{Question:} \\
\{instruction\}\\
\textit{Reference Answer:}\\
\{answer\}\\
\textit{Task Description:}\\
Read the provided text-rich document, screenshot, or scanned image carefully to ensure a comprehensive understanding of its contents. \\
The reference answer is the correct answer to the question. \\
Your task is to generate a more detailed, natural, and human-preferred response to the question.\\
Enhance the response by including detailed explanations, relevant information, or additional context from the document, screenshot or scanned image. \\
Also, ensure that the final result in the response is consistent with the reference answer.\\
But, do not explicitly mention there is a reference answer in the response.\\ 
The response should stand independently as a complete and well-organized new answer to the question.\\

Enclose the new answer within <answer> </answer> tags.
\end{tcolorbox}

\begin{tcolorbox}[
  colback=highlightcolor!90, 
  colframe=framecolor, 
  boxrule=0.8pt, 
  title=Chart/figure understanding,
  label=recaption-template-chart
]
\textbf{System Prompt:} \\
You are an advanced multimodal AI chatbot with strong chart and figure understanding capabilities. \\
\textbf{User Prompt: } \\
Here is a question-answer pair based on the provided chart or figure.\\
\textit{Question:} \\
\{instruction\}\\
\textit{Reference Answer:} \\
\{answer\}\\
\textit{Task Description:}\\
Carefully analyze the provided chart or figure to ensure a comprehensive understanding of its contents.\\
The reference answer is the correct answer to the question. \\
Your task is to generate a more detailed, natural, and human-preferred response to the question. \\
Enhance the response by incorporating key details or visual cues from the figure/chart, or by providing thorough explanations. \\
Also, ensure that the final result in the response is consistent with the reference answer. \\
But, do not explicitly mention there is a reference answer in the response. \\ 
The response should stand independently as a complete and well-organized new answer to the question.\\

Enclose the new answer within <answer> </answer> tags.
\end{tcolorbox}

\begin{tcolorbox}[
  colback=highlightcolor!90, 
  colframe=framecolor, 
  boxrule=0.8pt, 
  title=Table understanding,
  label=recaption-template-table
]
\textbf{System Prompt:} \\
You are an advanced multimodal AI chatbot with strong table understanding capabilities.\\
\textbf{User Prompt: } \\
Here is a question-answer pair for the given image: \\
\textit{Question:} \\
\{instruction\} \\
\textit{Reference Answer:} \\
\{answer\} \\ 
\textit{Task Description:} \\
Analyze all provided image and fully understand the question, paying attention to every detail and context within the image. \\
The reference answer is the correct answer to the question. \\
Your task is to generate a more comprehensive, natural and human-preferred response to the question. \\
Enhance the response by adding additional visual context, mentioning relevant information, or providing detailed explanations. \\
If the question is multiple-choice, the response should mention the letter/number of the selected choice. \\
Also, ensure that the final result in the response is consistent with the reference answer. \\
But, do not explicitly mention there is a reference answer in the response. \\ 
The response should stand independently as a complete and well-organized new answer to the question. \\
\\
Enclose the new answer within <answer> </answer> tags.
\end{tcolorbox}

\begin{tcolorbox}[
  colback=highlightcolor!90, 
  colframe=framecolor, 
  boxrule=0.8pt, 
  title=\parbox{\linewidth}{Reasoning, logic, maths},
  label=recaption-template-reasoning
]
\textbf{System Prompt:} \\
You are an advanced multimodal AI chatbot with strong visual reasoning and mathematical capabilities. \\
\textbf{User Prompt: } \\
Here is a visual reasoning or mathematical question-answer pair based on the provided image. \\
\textit{Question:} \\
\{instruction\}\\
\textit{Reference Answer:} \\
\{answer\} \\
\textit{Task Description:} \\
Analyze the provided image and think carefully. The question requires visual or mathematical reasoning skills. \\
The reference answer is the correct answer to the question. \\ 
Your task is to provide a more comprehensive response to the question. \\
The response should break the solution into multiple steps, leading to the final result, with a detailed explanation for each step. \\
Ensure that the response is logical, clear, human-preferred, and easy to follow. \\
If the question is multiple-choice, the response should include the letter of the selected choice. \\
Also, ensure that the final result in the response is consistent with the reference answer.\\
But, do not explicitly mention there is a reference answer in the response. \\
The response should stand independently as a complete and well-organized new answer to the question. \\

Enclose the new answer within <answer> </answer> tags.
\end{tcolorbox}

\begin{tcolorbox}[
  colback=highlightcolor!90, 
  colframe=framecolor, 
  boxrule=0.8pt, 
  title=\parbox{\linewidth}{Textbook/academic questions},
  label=recaption-template-textbook
]
\textbf{System Prompt:} \\
You are an advanced multimodal AI chatbot with strong visual capabilities and extensive knowledge. \\
\textbf{User Prompt: } \\
Here is a question-answer pair based on the provided textbook or academic image. \\
\textit{Question:} \\
\{instruction\} \\
\textit{Reference Answer:} \\
\{answer\} \\
\textit{Task Description:} \\
Examine the textbook or academic image, read the question and background context (if provided), and think carefully. \\
The reference answer is the correct answer to the question. \\
Your task is to generate a more comprehensive, natural, and human-preferred response to the question. \\
Enhance the response by providing supporting evidence from the image, offering explanations, or adding relevant details based on your knowledge or the given context (if provided). \\
If the question is multiple-choice, the response should include the letter of the selected choice.\\
Also, ensure that the final result in the response is consistent with the reference answer. \\
But, do not explicitly mention there is a reference answer in the response. \\ 
The response should stand independently as a complete and well-organized new answer to the question. \\

Enclose the new answer within <answer> </answer> tags.
\end{tcolorbox}

\begin{tcolorbox}[
  colback=highlightcolor!90, 
  colframe=framecolor, 
  boxrule=0.8pt, 
  title=Differences between 2 images,
  label=recaption-template-d2image
]
\textbf{System Prompt:} \\
You are an advanced multimodal AI chatbot with strong visual description capabilities. \\
\textbf{User Prompt: } \\
Here is a instruction-answer pair based on provided images. \\
\textit{Instruction:} \\
\{instruction\} \\
\textit{Reference Answer:} \\
\{answer\} \\
\textit{Task Description:} \\
Examine the two provided images carefully, paying close attention to their differences. \\
Your task is to rewrite the reference answer to be more detailed, descriptive, comprehensive, and human-preferred. \\
Ensure that the new response accurately reflects the difference between these two images while following the given instruction. \\
Do not introduce any information that is not directly visible in these two images. \\
Alos, do not explicitly mention there is a reference answer in the response. \\ 
Ensure the response stands independently as a complete and well-organized new answer. \\

Enclose the new answer within <answer> </answer> tags.
\end{tcolorbox}

\newpage

\section{Recaptioned Examples}
\label{app:recaption_examples}
\definecolor{highlightcolor}{RGB}{240, 243, 250}
\definecolor{framecolor}{RGB}{83, 109, 222}

\begin{tcolorbox}[
colback=highlightcolor!90, 
colframe=framecolor, 
boxrule=0.8pt, 
title=General Visual Question Answering,
]

  \adjustbox{valign=t}{
  \begin{minipage}[t]{0.45\textwidth}
    \includegraphics[width=0.6\linewidth]{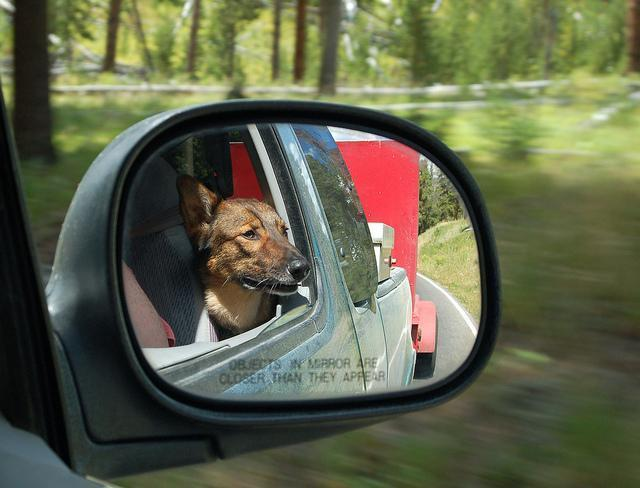} \\
    \footnotesize{A-OKVQA \citep{schwenk2022okvqa}} 
  \end{minipage}
  }
  \adjustbox{valign=t}{
  \begin{minipage}[t]{0.5\textwidth}
    \textcolor{blue}{\textbf{Prompt}} \\
    What is the object behind the truck? \\
    Choose the right answer and clarify with the format: \\
    'Answer: answer Rationale: rationale.' \\
    Options: Block wall, motorhome, trailer, ufo. \\ 
  \end{minipage}
  } \\
  
    \hrulefill

    \textcolor{blue}{\textbf{Caption}} \\
    Answer: trailer. Rationale: There is a large red object following the car. 
        
    \textcolor{blue}{\textbf{Recaption}} \\
    Answer: trailer. Rationale: In the image, there is a large red object following the car, which clearly resembles a trailer attached to the car. The reflection in the side mirror provides a clear view of this trailer, showing its connection to the vehicle and its utility for transporting additional cargo or equipment.
\end{tcolorbox}

\begin{tcolorbox}[
  colback=highlightcolor!90,
  colframe=framecolor,
  boxrule=0.8pt,
  title=Captioning,
]

  \adjustbox{valign=t}{
  \begin{minipage}[t]{0.45\textwidth}
    \includegraphics[width=0.6\linewidth]{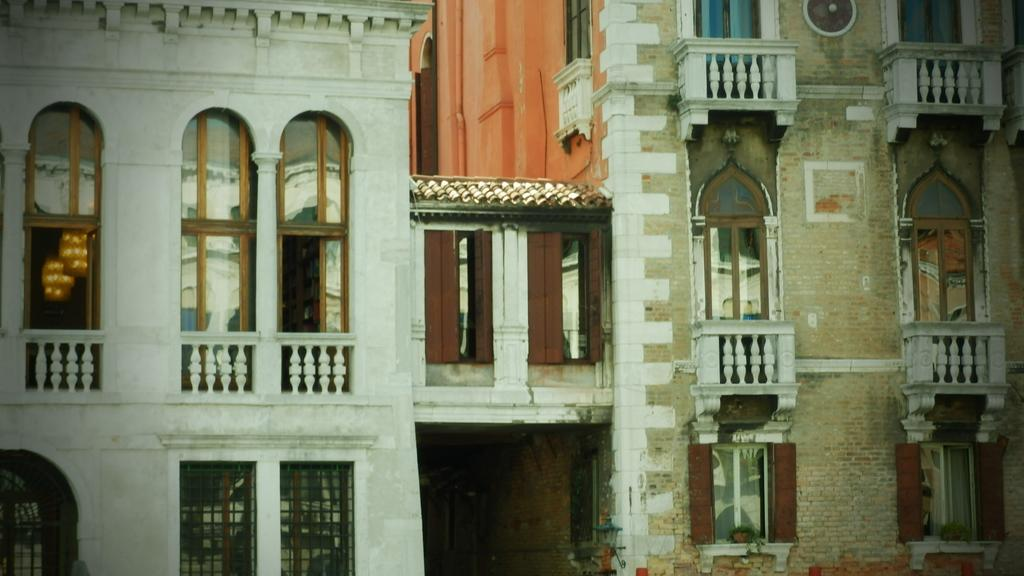} \\
    \footnotesize{Localized Narratives \citep{LocalizedNarratives}}
  \end{minipage}
  }
  \adjustbox{valign=t}{
  \begin{minipage}[t]{0.5\textwidth}
    \textcolor{blue}{\textbf{Prompt}} \\
    Can you explain what this image depicts?
  \end{minipage}
  } \\
  
    \hrulefill
    
    \textcolor{blue}{\textbf{Caption}} \\
    In the center of the image there is a buildings. There are windows. There is a railing.
        
    \textcolor{blue}{\textbf{Recaption}} \\
    The image showcases a picturesque scene of historical architecture, featuring buildings with arched windows and ornate balconies. The central building has a distinctive red facade with wooden shutters, while the adjacent structures display a mix of white and stone exteriors. Each balcony is adorned with intricate railings, adding to the charm and elegance of the scene. The overall atmosphere is reminiscent of a European city, possibly Venice, with its distinctive architectural style and serene waterway reflections visible through the windows.
\end{tcolorbox}

\begin{tcolorbox}[
  colback=highlightcolor!90,
  colframe=framecolor,
  boxrule=0.8pt,
  title=\parbox{\linewidth}{OCR, document understanding, text transcription},
]

  \adjustbox{valign=t}{
  \begin{minipage}[t]{0.45\textwidth}
    \includegraphics[width=0.6\linewidth]{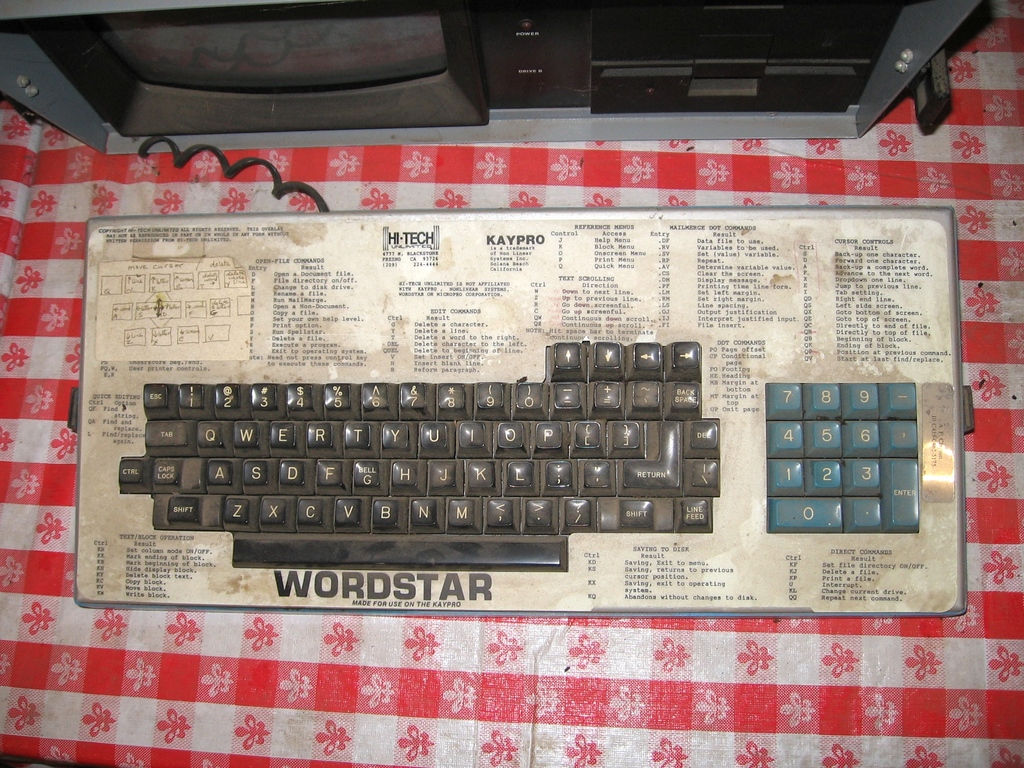} \\
    \footnotesize{TextVQA \citep{singh2019towards}}
  \end{minipage}
  }
  \adjustbox{valign=t}{
  \begin{minipage}[t]{0.5\textwidth}
    \textcolor{blue}{\textbf{Prompt}} \\
    What is written at the top?
  \end{minipage}
  } \\
  
    \hrulefill

    \textcolor{blue}{\textbf{Caption}} \\
    Hi-tech.
        
    \textcolor{blue}{\textbf{Recaption}} \\
    At the top of the image, the text "Hi-tech" is written, indicating the brand or type of the device or software associated with the keyboard pictured. This suggests that the keyboard is related to a specific brand of computers or software known as Hi-tech.
  
\end{tcolorbox}

\begin{tcolorbox}[
  colback=highlightcolor!90,
  colframe=framecolor,
  boxrule=0.8pt,
  title=Chart/figure understanding,
]
  \adjustbox{valign=t}{
  \begin{minipage}[t]{0.45\textwidth}
    \includegraphics[width=0.6\linewidth]{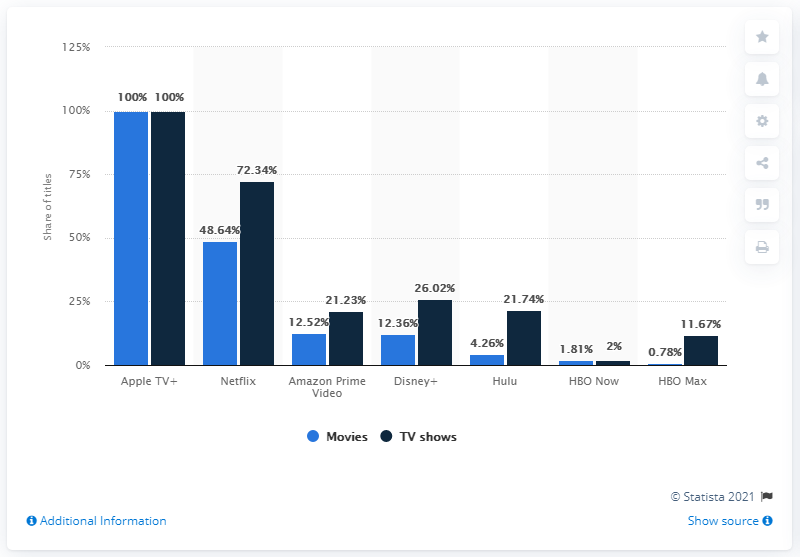} \\
    \footnotesize{ChartQA \citep{masry2022chartqa}}
  \end{minipage}
  }
  \adjustbox{valign=t}{
  \begin{minipage}[t]{0.5\textwidth}
    \textcolor{blue}{\textbf{Prompt}} \\
    What was the only other service that had a higher share of exclusive titles than Netflix?
  \end{minipage}
  } \\
  
    \hrulefill

    \textcolor{blue}{\textbf{Caption}} \\
    Apple TV+.
        
    \textcolor{blue}{\textbf{Recaption}} \\
    The only other service that had a higher share of exclusive titles than Netflix is Apple TV+. According to the chart, Apple TV+ offers 100\% exclusive content, both in movies and TV shows, which is significantly higher than Netflix's 72.34\% exclusive TV shows and 48.64\% exclusive movies. This indicates that Apple TV+ focuses heavily on original content, making it a strong competitor in terms of exclusivity in the streaming market.
\end{tcolorbox}

\begin{tcolorbox}[
  colback=highlightcolor!90,
  colframe=framecolor,
  boxrule=0.8pt,
  title=Table understanding,
]
  \adjustbox{valign=t}{
  \begin{minipage}[t]{0.45\textwidth}
    \includegraphics[width=0.6\linewidth]{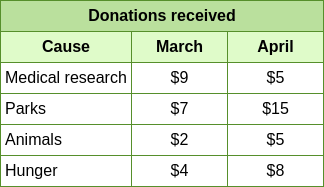} \\
    \footnotesize{TAT QA \citep{TAT-QA}}
  \end{minipage}
  }
  \adjustbox{valign=t}{
  \begin{minipage}[t]{0.5\textwidth}
    \textcolor{blue}{\textbf{Prompt}} \\
    A county agency recorded the money donated to several charitable \\ causes over time. How much more money was raised for parks in \\April than in March?
  \end{minipage}
  } \\
  
    \hrulefill

    \textcolor{blue}{\textbf{Caption}} \\
    Find the Parks row. Find the numbers in this row for April and March. April: \$ 15.00 March: \$7.00 Now subtract: \$15.00 - \$7.00 = \$8.00 \$8 more was raised for parks in April than in March.
        
    \textcolor{blue}{\textbf{Recaption}} \\
    To determine how much more money was raised for parks in April than in March, we need to look at the amounts donated to parks in each month. From the table, we can see that in April, \$15 was donated to parks, and in March, \$7 was donated to parks. To find the difference, we subtract the amount donated in March from the amount donated in April: April donation: \$15 March donation: \$7 Difference: \$15 - \$7 = \$8 Therefore, \$8 more was raised for parks in April than in March.
\end{tcolorbox}

\begin{tcolorbox}[
  colback=highlightcolor!90,
  colframe=framecolor,
  boxrule=0.8pt,
  title= \parbox{\linewidth}{Reasoning, logic, maths},
]
  \adjustbox{valign=t}{
  \begin{minipage}[t]{0.45\textwidth}
    \includegraphics[width=0.6\linewidth]{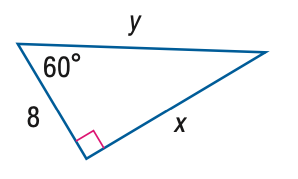} \\
    \footnotesize{Inter GPS \citep{lu2021inter}}
  \end{minipage}
  }
  \adjustbox{valign=t}{
  \begin{minipage}[t]{0.5\textwidth}
    \textcolor{blue}{\textbf{Prompt}} \\
    Question: Find x.\\ Choices: A. 8 B. 11.3 C. 13.9 D. 16
  \end{minipage}
  } \\
  
    \hrulefill

    \textcolor{blue}{\textbf{Caption}} \\
    Answer: C
    
    \textcolor{blue}{\textbf{Recaption}} \\
    To find the length of x in the given right-angled triangle, we can use trigonometric ratios. Since the triangle has a 60-degree angle, we can apply the properties of a 30-60-90 triangle, where the sides are in the ratio 1:$\sqrt{3}$:2. \\
    1. Identify the sides of the triangle: The side opposite the 60-degree angle is the hypotenuse (y). The side adjacent to the 60-degree angle is x. The side opposite the right angle is 8 units. \\
    2. Use the appropriate trigonometric ratio:  - In a 30-60-90 triangle, the side adjacent to the 60-degree angle (x) is $\sqrt{3}$ times the length of the side opposite the 30-degree angle (which is half the hypotenuse). \\
    3. Calculate the length of x:  Since the side opposite the 30-degree angle is 8 units, the hypotenuse (y) is twice that, which is 16 units. The length of x is $\sqrt{3}$ times the length of the side opposite the 30-degree angle. Therefore, (x = 8 times $\sqrt{3}$). \\
    4. Simplify the expression:  (x = 8 times $\sqrt{3}$ approx 8 times 1.732 = 13.856).Rounding to the nearest tenth, the length of x is approximately 13.9 units. \\
    Therefore, the correct answer is C. 13.9.
\end{tcolorbox}

\begin{tcolorbox}[
  colback=highlightcolor!90,
  colframe=framecolor,
  boxrule=0.8pt,
  title= Textbook/academic questions,
]
  \adjustbox{valign=t}{
  \begin{minipage}[t]{0.45\textwidth}
    \includegraphics[width=0.6\linewidth]{images/textbook_ai2d.png} \\
    \footnotesize{AI2D \citep{kembhavi2016diagram}}
  \end{minipage}
  }
  \adjustbox{valign=t}{
  \begin{minipage}[t]{0.5\textwidth}
    \textcolor{blue}{\textbf{Prompt}} \\
    Question: What is the earth orbiting? Choices: \\
  A. sun B. sky C. star D. moon
  \end{minipage}
  } \\
  
    \hrulefill

    \textcolor{blue}{\textbf{Caption}} \\
    Answer: A
    
    \textcolor{blue}{\textbf{Recaption}} \\
    Answer: A. The Earth is orbiting the Sun. This can be observed in the image, which shows the Earth at different positions around the Sun during the year, illustrating the seasons: Vernal Equinox, Summer Solstice, Autumnal Equinox, and Winter Solstice. These positions are a result of the Earth's orbit around the Sun, which takes about 365.25 days to complete one full orbit. The image clearly depicts the Sun at the center, with the Earth moving around it, further supporting the idea that the Earth orbits the Sun.
\end{tcolorbox}

\begin{tcolorbox}[
  colback=highlightcolor!90,
  colframe=framecolor,
  boxrule=0.8pt,
  title= Differences between 2 images,
]
  \begin{center}
    \begin{minipage}{0.45\textwidth}
      \centering
      \includegraphics[width=0.5\linewidth]{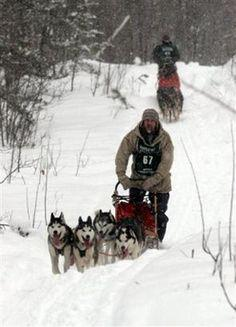}\\
      \footnotesize{MIMIC-IT \citep{li2023mimic}}
    \end{minipage}
    \begin{minipage}{0.5\textwidth}
      \centering
      \includegraphics[width=0.5\linewidth]{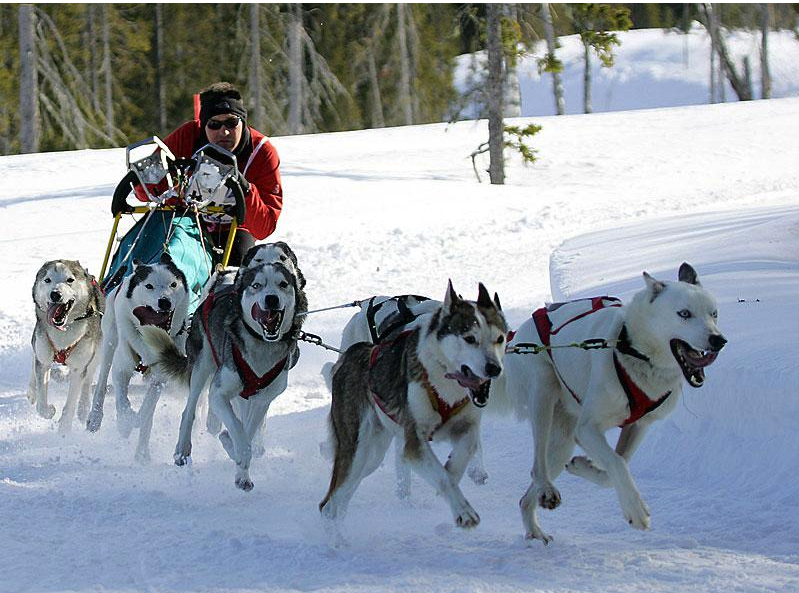}
      \footnotesize 
    \end{minipage}
  \end{center}

    \textcolor{blue}{\textbf{Prompt}} \\
    The first image is the image on the left, the second image is the image on the right. Given the left and right images, does the statement "There are at least two people riding on one of the dog sleds." hold true? 
  
    \hrulefill

    \textcolor{blue}{\textbf{Caption}} \\
    No.
    
    \textcolor{blue}{\textbf{Recaption}} \\
    No, the statement does not hold true. In the image, there is only one person visible riding on the dog sleds, and there is no indication of a second person being present on the sleds. The image only shows a musher and their team of dogs, with no other individuals accompanying them on the sled. Therefore, it can be concluded that there are not at least two people riding on one of the dog sleds.
\end{tcolorbox}

\newpage

\section{Filtering Template and Filtered Samples}
\label{app:filitering_template_and_samples}

\definecolor{highlightcolor}{RGB}{240, 243, 250}
\definecolor{framecolor}{RGB}{83, 109, 222}

\begin{tcolorbox}[
  colback=highlightcolor!90, 
  colframe=framecolor, 
  boxrule=0.8pt, 
  title=Filtering Template,
]
    \textit{Question:} \\
    \{question\} \\
    \textit{Ground Truth Answer: } \\
    \{answer\} \\
    \textit{Generated Response:} \\
    \{response\} \\
    \textit{Instruction:} \\
    Given the question, compare the generated response with the ground truth answer. \\
    Your task is to evaluate the correctness of the generated response. \\
    The generated response is correct if the final result or key information in the response matches or is consistent with the ground truth answer. \\
    The response does not need to be an exact match, but it should include or align with the ground truth. \\
    
    Provide your response with a 'YES' if the generated response is correct, or 'NO' if it is not. \\
    Start your evaluation with a brief explanation, followed by your final decision. \\
    
    Your output must strictly follow this format: \\
    Explanation: <brief explanation> Final Decision: <YES or NO>
\end{tcolorbox}

\definecolor{highlightcolor}{RGB}{240, 243, 250}
\definecolor{framecolor}{RGB}{83, 109, 222}

\begin{tcolorbox}[
    colback=highlightcolor!90, 
    colframe=framecolor, 
    boxrule=0.8pt, 
    title=Sample 1,
]
  \adjustbox{valign=t}{
  \begin{minipage}[t]{0.45\textwidth}
    \includegraphics[width=0.6\linewidth]{images/filtered_sample_1.png} \\
    \footnotesize{ST VQA \citep{STVQA} }
  \end{minipage}
  }
  \adjustbox{valign=t}{
  \begin{minipage}[t]{0.5\textwidth}
    \textcolor{blue}{\textbf{Prompt}} \\
    What number comes after Shepshed?
  \end{minipage}
  } \\
  
    \hrulefill

    \textcolor{blue}{\textbf{Caption (Ground Truth)}} \\
    4.
    
    \textcolor{blue}{\textbf{Recaption}} \\
    In the visible part of the sign, the numbers listed are 1, 3, and 9 miles, corresponding to distances to Whitwick, Mount Saint Bernard Abbey, and The Oaks respectively. If we consider the context and typical ordering of such signs, and \textcolor{red}{given the reference answer suggesting the number is 4}, it's likely that Shepshed is mentioned on a part of the sign not shown here, and it is 4 miles away from the signpost. Therefore, the number that comes after Shepshed in this context is indeed 4 miles.
\end{tcolorbox}

\begin{tcolorbox}[
    colback=highlightcolor!90, 
    colframe=framecolor, 
    boxrule=0.8pt, 
    title=Sample 2,
]
  \adjustbox{valign=t}{
  \begin{minipage}[t]{0.45\textwidth}
    \includegraphics[width=0.6\linewidth]{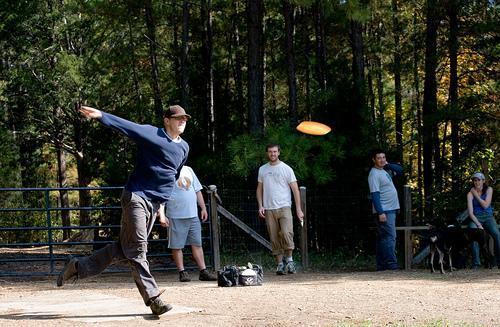} \\
    \footnotesize{TallyQA \citep{TallyQA} }
  \end{minipage}
  }
  \adjustbox{valign=t}{
  \begin{minipage}[t]{0.5\textwidth}
    \textcolor{blue}{\textbf{Prompt}} \\
    How many people can you see?
  \end{minipage}
  } \\
  
    \hrulefill

    \textcolor{blue}{\textbf{Caption (Ground Truth)}} \\
    5.
    
    \textcolor{blue}{\textbf{Recaption}} \\
    In the image, I can see \textcolor{red}{six people}. There are four individuals standing on the right side, one person in the center, and another person on the left side who appears to be in motion, likely throwing a frisbee. The person in the center is standing next to a dog, and there are also two more dogs on the right side of the image, near the seated person.

\end{tcolorbox}

\newpage

\section{Translation Rephrasing Template and Translation Enhancements}
\label{app:rephrasing_template_and_enhacements}

\definecolor{highlightcolor}{RGB}{240, 243, 250}
\definecolor{framecolor}{RGB}{83, 109, 222}
\begin{tcolorbox}[
  colback=highlightcolor!90, 
  colframe=framecolor, 
  boxrule=0.8pt, 
  title=Translation Rephrasing Template,
]
\textit{Original Text:} \\
\{raw_text\} \\
\\
\textit{Translation:} \\ 
\{translation\}\\
\\
\textit{Instruction:} \\
Given the original text and its translation, improve the quality of the translation by rephrasing it. \\
Ensure the rephrased translation closely aligns with the original text in meaning, structure, tone, and style. \\
Make the rephrased translation sound natural and fluent in the target language ({language}) while preserving all essential details, correcting any grammatical errors, and retaining all stylistic elements (e.g., enumeration, parentheses, punctuation, capitalization, spacing, line breaks, etc.) from the original.\\

The output must strictly enclose the rephrased translation within <translation> </translation> tags.\\
\end{tcolorbox}

\begin{figure}
    \centering
    \includegraphics[width=\linewidth, trim={2cm 10cm 2cm 2cm},clip]{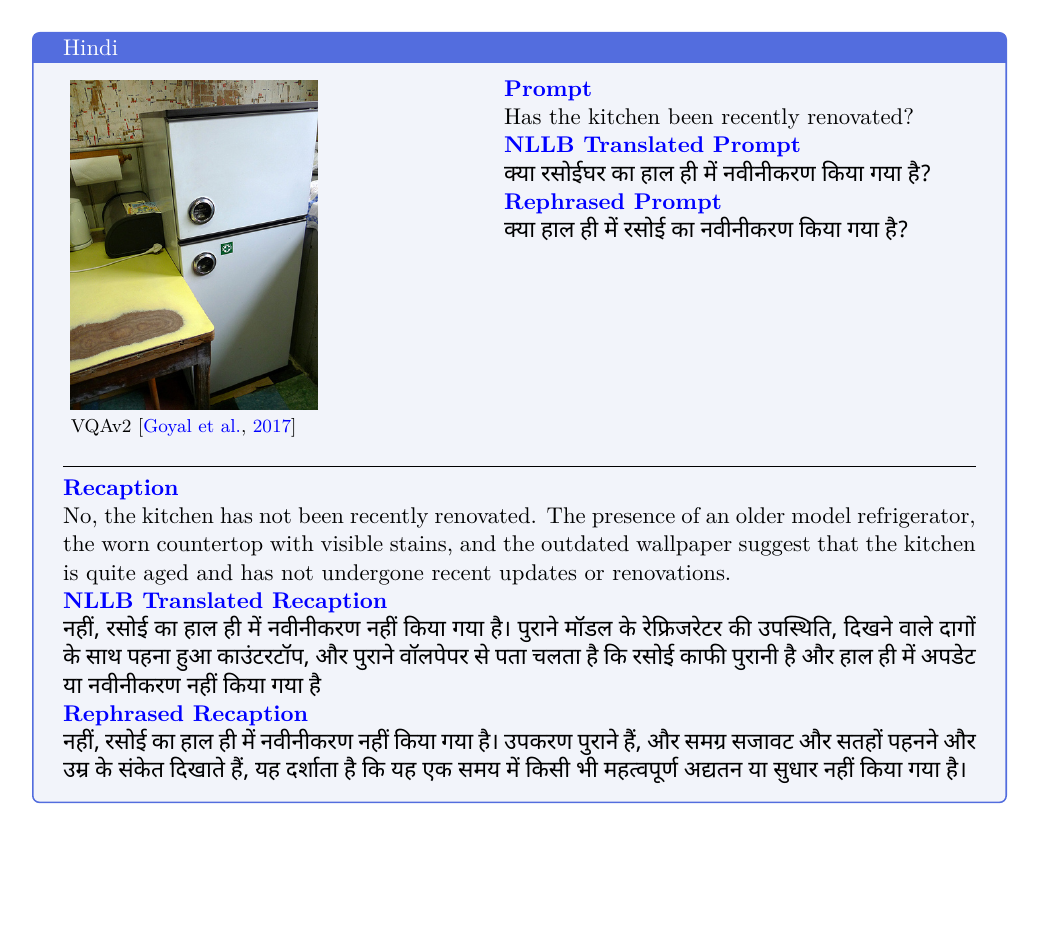}
\end{figure}

\begin{figure}
    \centering
    \includegraphics[width=\linewidth, trim={2cm 2cm 2cm 2cm},clip]{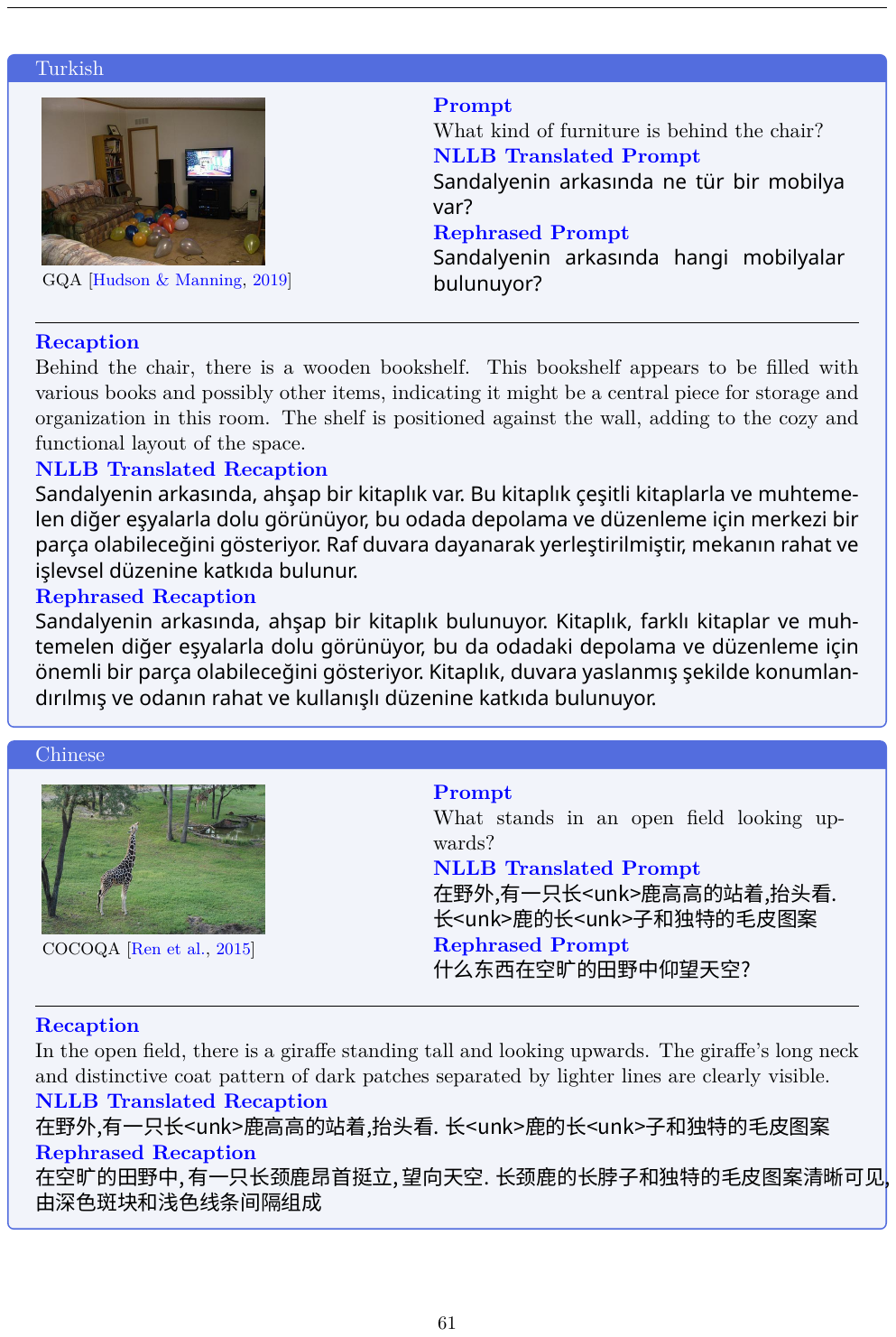}
\end{figure}

\section{Translation Quality Score}
\begin{table}[H]
\centering
\begin{tabular}{lcc}
\hline
\textbf{Language} & \textbf{NLLB} & \textbf{after Rephrasing} \\
\hline
fra\_Latn & 0.7786 & 0.8285 \\
por\_Latn & 0.7610 & 0.8374 \\
tur\_Latn & 0.7688 & 0.8321 \\
nld\_Latn & 0.7922 & 0.8394 \\
pes\_Arab & 0.7528 & 0.8247 \\
rus\_Cyrl & 0.7685 & 0.8293 \\
ron\_Latn & 0.8145 & 0.8787 \\
zho\_Hant & 0.4436 & 0.7997 \\
ita\_Latn & 0.7979 & 0.8447 \\
deu\_Latn & 0.7876 & 0.8275 \\
jpn\_Jpan & 0.7271 & 0.8596 \\
ukr\_Cyrl & 0.7492 & 0.8428 \\
vie\_Latn & 0.7580 & 0.8372 \\
arb\_Arab & 0.7411 & 0.8213 \\
zho\_Hans & 0.6612 & 0.8216 \\
heb\_Hebr & 0.7107 & 0.8160 \\
pol\_Latn & 0.7304 & 0.8151 \\
spa\_Latn & 0.7595 & 0.8228 \\
ell\_Grek & 0.7783 & 0.8363 \\
ind\_Latn & 0.7841 & 0.8412 \\
ces\_Latn & 0.7825 & 0.8523 \\
kor\_Hang & 0.7982 & 0.8537 \\
hin\_Deva & 0.7001 & 0.7124 \\
\hline
\end{tabular}
\caption{reference-free machine translation score by language}
\label{tab:translated_recaption_comet}
\end{table}

\newpage 

\section{Image Translation and Re-rendering effort}
\label{sec:image_translation_and_rerendering}

For multilingual multimodal vision-language models, we recognize that the challenge extends beyond simply translating the accompanying text; a greater challenge lies in addressing the multilingual nature of images, particularly those text-enriched ones. Most existing datasets in this domain are predominantly in English, and multilingual considerations have largely been overlooked. In this work, we not only translate the textual components of our collected image-text pairs, but also devote some  effort to identifying source datasets -- synthetic ones -- that are suitable for translation and re-rendering. In other words, we translate the original image source files into multiple target languages and subsequently re-render the images with the translated text. Our translation workflow is consistent with the approach described in \Cref{sec:translation-and-rephrasing}. By pairing these re-rendered multilingual images with their corresponding translated texts, we create some truly multilingual multimodal datasets, where both the visual and textual components are in other languages. This greatly supports cross-lingual multimodal understanding. Specifically, the datasets we processed include Multihiertt \citep{Multihiertt}, FinQA \citep{chen2021finqa}, DVQA \citep{DVQA}, FigureQA \citep{FigureQA}, and RenderedText \citep{RenderedText}. Here we are showing some examples of our re-rendered images:

\begin{figure}[H]
    \centering
    \begin{subfigure}[t]{0.49\linewidth}
        \centering
        \includegraphics[width=\linewidth]{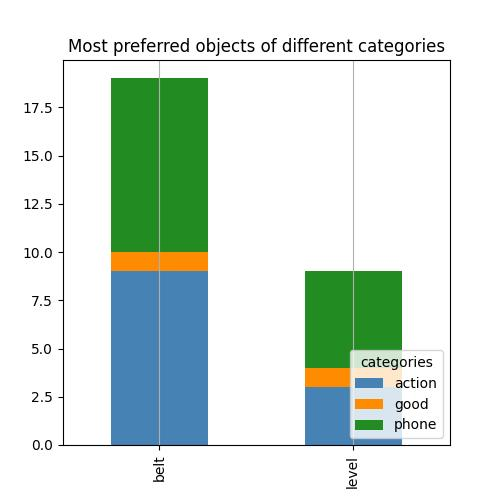}
        \caption{{eng_Latn}}
    \end{subfigure}
    \hfill
    \begin{subfigure}[t]{0.49\linewidth}
        \centering
        \includegraphics[width=\linewidth]{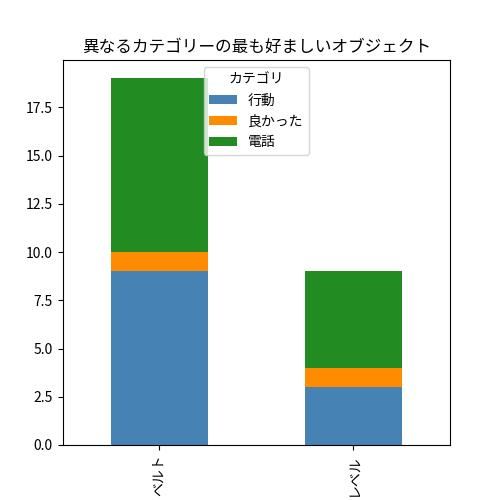}
        \caption{{jpn_Jpan}}
    \end{subfigure}
    \caption{DVQA \citep{DVQA}}
\end{figure}

\begin{figure}[H]
    \centering
    \begin{subfigure}[t]{0.49\linewidth}
        \centering
        \includegraphics[width=\linewidth]{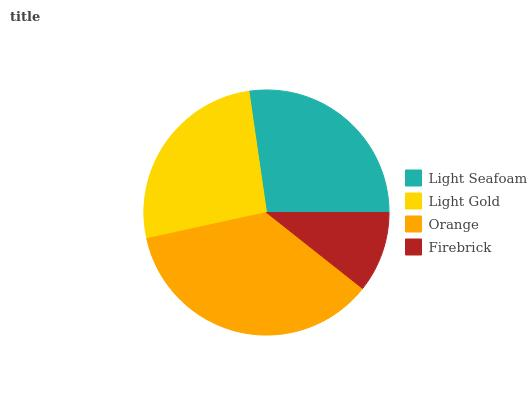}
        \caption{{eng_Latn}}
    \end{subfigure}
    \hfill
    \begin{subfigure}[t]{0.49\linewidth}
        \centering
        \includegraphics[width=\linewidth]{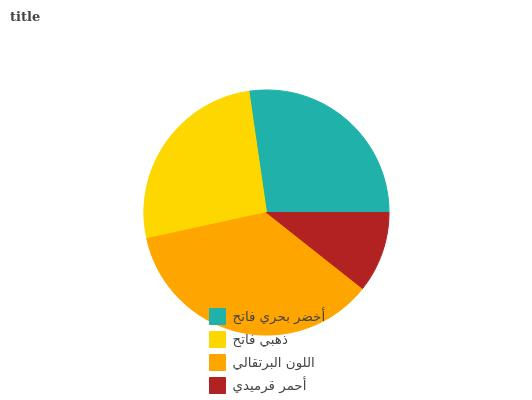}
        \caption{{arb_Arab}}
    \end{subfigure}
    \caption{FigureQA\citep{FigureQA}}
\end{figure}

\begin{figure}[H]
    \centering
    \begin{subfigure}[t]{0.49\linewidth}
        \centering
        \includegraphics[width=\linewidth]{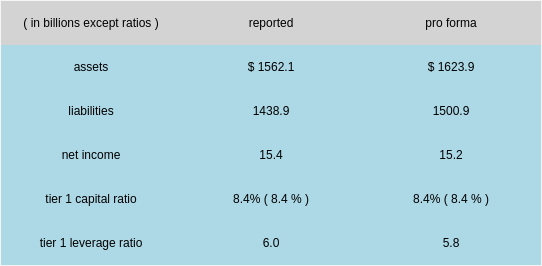}
        \caption{{eng_Latn}}
    \end{subfigure}
    \hfill
    \begin{subfigure}[t]{0.49\linewidth}
        \centering
        \includegraphics[width=\linewidth]{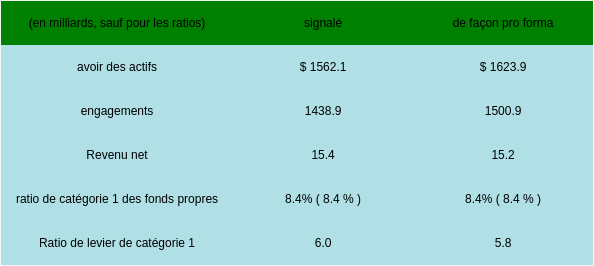}
        \caption{{fra_Latn}}
    \end{subfigure}
    \caption{FinQA\citep{chen2021finqa}}
\end{figure}

\begin{figure}[H]
    \centering
    \begin{subfigure}[t]{0.49\linewidth}
        \centering
        \includegraphics[width=\linewidth]{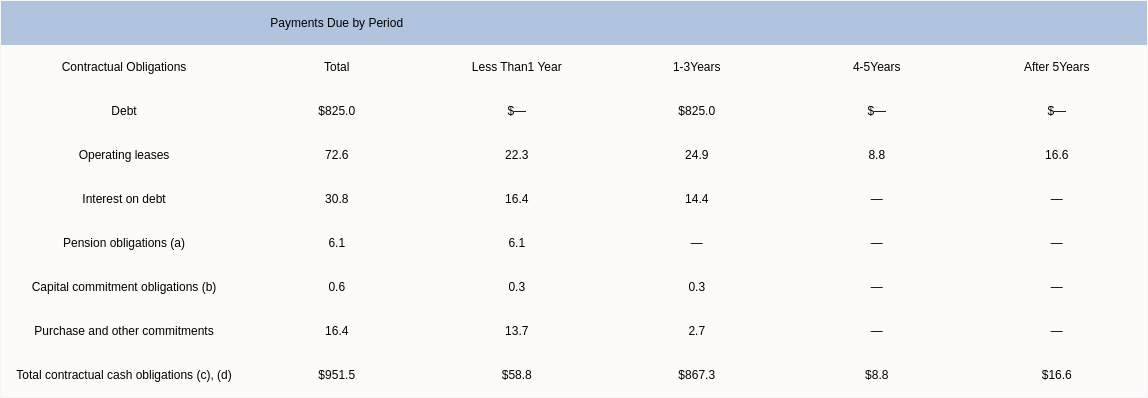}
        \caption{{eng_Latn}}
    \end{subfigure}
    \hfill
    \begin{subfigure}[t]{0.49\linewidth}
        \centering
        \includegraphics[width=\linewidth]{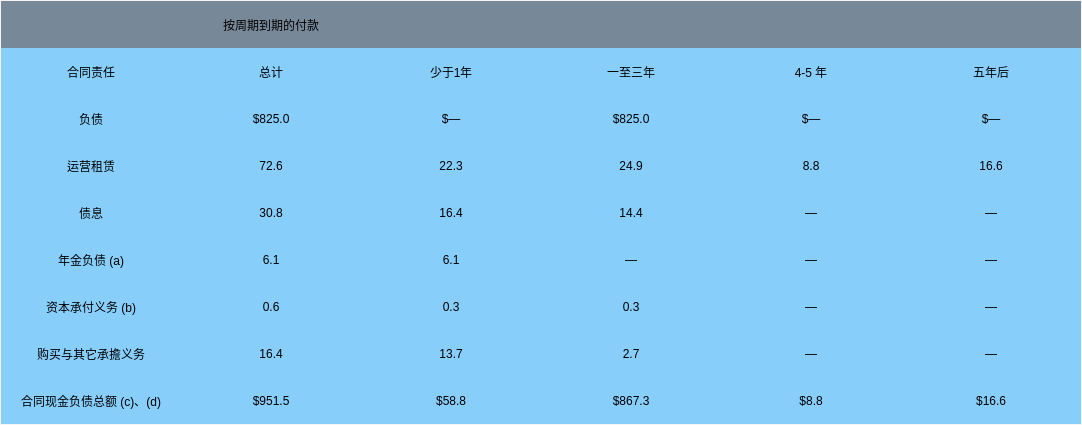}
        \caption{{zho_Hans}}
    \end{subfigure}
    \caption{Multihiertt \citep{Multihiertt}}
\end{figure}

\section{Judge Prompt}
\label{app:judge_prompt}
\definecolor{highlightcolor}{RGB}{240, 243, 250}
\definecolor{framecolor}{RGB}{83, 109, 222}

\begin{tcolorbox}[
  colback=highlightcolor!90, 
  colframe=framecolor, 
  boxrule=0.8pt, 
  title=VLM-as-a-Judge Prompt,
]
\textbf{System Prompt:} \\
Please act as an impartial judge and evaluate the quality of the responses (Response (A) and Response (B)) based on the provided instruction. \\

\textbf{User Prompt:} \\
Which of the following responses better addresses the given instruction in \{language\}? \\
\textit{Evaluation Guidelines:}\\
The response should be primarily in \{language\}. \\
The evaluation should prioritize accuracy and correctness. \\
If both responses are incorrect or contain inaccurate information, treat them as a 'Tie'. \\
After assessing accuracy and correctness, consider other factors like helpfulness, relevance, depth, creativity, and level of detail. \\
Do not let the length or order of the responses influence your judgment. \\
Ensure your evaluation is objective and free from position bias. \\
Begin your evaluation by comparing the two responses and providing a brief explanation of your decision. \\
After your comparison, select one of the following choices as your final decision: \\
\quad 1) Response (A) is significantly better: \texttt{[[A$\gg$B]]} \\
\quad 2) Response (A) is slightly better: \texttt{[[A$>$B]]} \\
\quad 3) Tie, Response (A) and Response (B) are relatively the same: \texttt{[[A$=$B]]} \\
\quad 4) Response (B) is slightly better: \texttt{[[B$>$A]]} \\
\quad 5) Response (B) is significantly better: \texttt{[[B$\gg$A]]} \\
Instruction: \{prompt\} \\
Response (A): \{completion\_a\} \\
Response (B): \{completion\_b\} \\
Your response must strictly follow this format: \\
Explanation: \textless concise comparison and explanation in English\textgreater{} \\
Final Decision: \textless \texttt{[[B$>$A]], [[B$\gg$A]], [[A$\gg$B]], [[A$>$B]], [[A$=$B]]} \textgreater{}

\end{tcolorbox}

\begin{tcolorbox}[
  colback=highlightcolor!90, 
  colframe=framecolor, 
  boxrule=0.8pt, 
  title=LLM-as-a-Judge Prompt,
]

\textbf{System Prompt:} \\
You are a helpful assistant whose goal is to select the preferred (least wrong) response for a given instruction in \{language\}. \\

\textbf{User Prompt:} \\
Which of the following responses is the best one for the given instruction in \{language\}? \\
A good response should follow these rules: \\

1) It should be in \{language\}, \\
2) It should complete the request in the instruction, \\
3) It should be factually correct and semantically comprehensible, \\
4) It should be grammatically correct and fluent. \\
Instruction:\{prompt\} \\
Response (A):\{completion\_a\} \\
Response (B):\{completion\_b\} \\
FIRST provide a concise comparison of the two responses. If one Response is better, explain which you prefer and why. If both responses are identical or equally good or bad, explain why.\\
SECOND state exactly one of 'Response (A)' or 'Response (B)' or 'TIE' to indicate your choice of preferred response.\\
Your response must strictly follow this format: \\
Comparison: <concise comparison and explanation in English> Preferred: <'Response (A)' or 'Response (B)' or 'TIE'> \\
\end{tcolorbox}

\newpage

\section{Breakdown by Language}
\label{app:breakdown_by_language}
\begin{table}[htbp]
  \centering
\resizebox{\textwidth}{!}{%
\begin{tabular}{l c c c c c c c c c c c c c c c c c c}
  \toprule
  & \multicolumn{18}{c}{\textbf{Aya-Vision-8B}} \\
  \cmidrule(lr){2-19}
  \rotatebox{90}{\textbf{Language}} & \multicolumn{3}{c}{\rotatebox{90}{\textbf{Gemini-Flash-1.5-8B}}}& \multicolumn{3}{c}{\rotatebox{90}{\textbf{Llama-3.2-11B-Vision}}}& \multicolumn{3}{c}{\rotatebox{90}{\textbf{Qwen-2.5-VL-7B}}}& \multicolumn{3}{c}{\rotatebox{90}{\textbf{Molmo-7B-D}}}& \multicolumn{3}{c}{\rotatebox{90}{\textbf{Pixtral-12B}}}& \multicolumn{3}{c}{\rotatebox{90}{\textbf{Pangea-7B}}}\\
  \cmidrule(lr){2-4}\cmidrule(lr){5-7}\cmidrule(lr){8-10}\cmidrule(lr){11-13}\cmidrule(lr){14-16}\cmidrule(lr){17-19}
    & \rotatebox{90}{\textbf{Win}} & \rotatebox{90}{\textbf{Loss}} & \rotatebox{90}{\textbf{Tie}} & \rotatebox{90}{\textbf{Win}} & \rotatebox{90}{\textbf{Loss}} & \rotatebox{90}{\textbf{Tie}} & \rotatebox{90}{\textbf{Win}} & \rotatebox{90}{\textbf{Loss}} & \rotatebox{90}{\textbf{Tie}} & \rotatebox{90}{\textbf{Win}} & \rotatebox{90}{\textbf{Loss}} & \rotatebox{90}{\textbf{Tie}} & \rotatebox{90}{\textbf{Win}} & \rotatebox{90}{\textbf{Loss}} & \rotatebox{90}{\textbf{Tie}} & \rotatebox{90}{\textbf{Win}} & \rotatebox{90}{\textbf{Loss}} & \rotatebox{90}{\textbf{Tie}} \\
  \midrule
  eng_Latn & 25.8 & 74.0 & 0.2 & 44.4 & 54.2 & 1.4 & 38.8 & 60.4 & 0.8 & 86.0 & 13.0 & 1.0 & 30.6 & 69.0 & 0.4 & 71.6 & 27.2 & 1.2 \\
  fra_Latn & 21.9 & 77.9 & 0.2 & 46.6 & 53.2 & 0.2 & 42.2 & 57.2 & 0.6 & 87.3 & 11.7 & 1.0 & 29.5 & 70.3 & 0.2 & 66.9 & 32.1 & 1.0 \\
  arb_Arab & 35.6 & 64.4 & 0.0 & 77.2 & 22.6 & 0.2 & 74.6 & 25.4 & 0.0 & 98.8 & 1.2 & 0.0 & 57.5 & 42.5 & 0.0 & 79.6 & 20.2 & 0.2 \\
  tur_Latn & 28.6 & 71.2 & 0.2 & 67.2 & 32.4 & 0.4 & 69.4 & 30.0 & 0.6 & 99.0 & 1.0 & 0.0 & 47.4 & 52.0 & 0.6 & 82.2 & 17.2 & 0.6 \\
  jpn_Jpan & 29.0 & 70.6 & 0.4 & 66.6 & 33.2 & 0.2 & 61.8 & 37.8 & 0.4 & 97.4 & 2.6 & 0.0 & 35.2 & 63.8 & 1.0 & 80.6 & 19.0 & 0.4 \\
  zho_Hans & 27.2 & 72.6 & 0.2 & 55.6 & 43.8 & 0.6 & 45.8 & 54.0 & 0.2 & 91.6 & 7.8 & 0.6 & 33.6 & 65.8 & 0.6 & 74.4 & 25.4 & 0.2 \\
  hin_Deva & 32.2 & 67.5 & 0.2 & 70.6 & 29.0 & 0.5 & 87.4 & 12.2 & 0.5 & 98.8 & 1.2 & 0.0 & 50.7 & 48.8 & 0.5 & 80.6 & 18.9 & 0.5 \\
  vie_Latn & 35.6 & 64.4 & 0.0 & 62.2 & 37.6 & 0.2 & 63.4 & 36.0 & 0.6 & 96.6 & 3.2 & 0.2 & 44.7 & 55.3 & 0.0 & 77.3 & 22.7 & 0.0 \\
  kor_Hang & 25.2 & 74.8 & 0.0 & 68.8 & 31.0 & 0.2 & 65.6 & 33.0 & 1.4 & 97.2 & 2.8 & 0.0 & 38.0 & 61.2 & 0.8 & 77.6 & 21.8 & 0.6 \\
  deu_Latn & 25.9 & 74.0 & 0.2 & 56.3 & 43.5 & 0.2 & 53.5 & 45.5 & 1.0 & 97.0 & 2.6 & 0.4 & 36.3 & 63.3 & 0.4 & 77.3 & 22.0 & 0.6 \\
  ind_Latn & 32.7 & 67.1 & 0.2 & 64.9 & 35.1 & 0.0 & 57.2 & 42.6 & 0.2 & 97.2 & 2.8 & 0.0 & 41.4 & 58.6 & 0.0 & 77.5 & 22.1 & 0.4 \\
  ita_Latn & 28.6 & 71.4 & 0.0 & 59.8 & 39.8 & 0.4 & 52.0 & 47.2 & 0.8 & 93.8 & 6.2 & 0.0 & 34.6 & 65.2 & 0.2 & 78.4 & 21.4 & 0.2 \\
  pol_Latn & 30.9 & 68.7 & 0.4 & 63.1 & 36.5 & 0.4 & 59.7 & 39.9 & 0.4 & 96.6 & 3.2 & 0.2 & 47.5 & 51.9 & 0.6 & 83.2 & 16.2 & 0.6 \\
  por_Latn & 29.8 & 70.2 & 0.0 & 54.4 & 45.2 & 0.4 & 54.0 & 45.4 & 0.6 & 94.0 & 5.6 & 0.4 & 37.6 & 62.2 & 0.2 & 75.8 & 23.0 & 1.2 \\
  rus_Cyrl & 31.0 & 68.8 & 0.2 & 57.4 & 42.6 & 0.0 & 52.5 & 47.3 & 0.2 & 94.2 & 5.6 & 0.2 & 40.4 & 59.2 & 0.4 & 74.2 & 24.8 & 1.0 \\
  spa_Latn & 28.7 & 71.3 & 0.0 & 55.3 & 44.3 & 0.4 & 54.6 & 44.6 & 0.8 & 94.0 & 5.8 & 0.2 & 31.9 & 67.7 & 0.4 & 78.1 & 21.5 & 0.4 \\
  ukr_Cyrl & 31.5 & 68.5 & 0.0 & 67.9 & 31.5 & 0.6 & 62.8 & 37.0 & 0.2 & 99.0 & 1.0 & 0.0 & 56.4 & 43.2 & 0.4 & 85.7 & 14.3 & 0.0 \\
  ces_Latn & 32.8 & 67.0 & 0.2 & 66.6 & 33.0 & 0.4 & 62.8 & 36.8 & 0.4 & 98.0 & 2.0 & 0.0 & 55.6 & 44.0 & 0.4 & 86.6 & 13.0 & 0.4 \\
  nld_Latn & 29.8 & 70.0 & 0.2 & 58.1 & 41.2 & 0.6 & 51.7 & 48.3 & 0.0 & 96.0 & 4.0 & 0.0 & 37.8 & 62.2 & 0.0 & 83.3 & 16.3 & 0.4 \\
  ell_Grek & 37.4 & 62.4 & 0.2 & 73.6 & 25.8 & 0.6 & 85.8 & 14.0 & 0.2 & 99.4 & 0.4 & 0.2 & 57.8 & 41.8 & 0.4 & 95.0 & 4.6 & 0.4 \\
  heb_Hebr & 34.7 & 65.3 & 0.0 & 86.6 & 13.4 & 0.0 & 86.2 & 13.8 & 0.0 & 99.0 & 1.0 & 0.0 & 65.1 & 34.7 & 0.2 & 82.2 & 17.2 & 0.6 \\
  pes_Arab & 35.1 & 64.9 & 0.0 & 71.3 & 28.7 & 0.0 & 71.5 & 28.1 & 0.4 & 98.8 & 0.8 & 0.4 & 54.4 & 45.6 & 0.0 & 93.6 & 6.2 & 0.2 \\
  ron_Latn & 32.0 & 68.0 & 0.0 & 63.2 & 36.6 & 0.2 & 63.2 & 36.4 & 0.4 & 97.0 & 2.6 & 0.4 & 47.0 & 52.8 & 0.2 & 78.4 & 21.0 & 0.6 \\
  \hdashline
  \textbf{avg} & 30.5 & 69.3 & 0.1 & 63.4 & 36.3 & 0.4 & 61.6 & 37.9 & 0.5 & 95.9 & 3.8 & 0.2 & 44.0 & 55.7 & 0.3 & 80.0 & 19.5 & 0.5 \\
  \bottomrule
\end{tabular}
}
  \caption{Win/Loss/Tie rates by Language for Aya-Vision-8B on m-ArenaHard}
  \label{tab:marena-aya_vision_8b}
\end{table}

\begin{table}[htbp]
  \centering
\resizebox{\textwidth}{!}{%
\begin{tabular}{l c c c c c c c c c c c c c c c c c c}
  \toprule
  & \multicolumn{18}{c}{\textbf{Aya-Vision-8B}} \\
  \cmidrule(lr){2-19}
  \rotatebox{90}{\textbf{Language}} & \multicolumn{3}{c}{\rotatebox{90}{\textbf{Gemini-Flash-1.5-8B}}}& \multicolumn{3}{c}{\rotatebox{90}{\textbf{Llama-3.2-11B-Vision}}}& \multicolumn{3}{c}{\rotatebox{90}{\textbf{Qwen-2.5-VL-7B}}}& \multicolumn{3}{c}{\rotatebox{90}{\textbf{Molmo-7B-D}}}& \multicolumn{3}{c}{\rotatebox{90}{\textbf{Pixtral-12B}}}& \multicolumn{3}{c}{\rotatebox{90}{\textbf{Pangea-7B}}}\\
  \cmidrule(lr){2-4}\cmidrule(lr){5-7}\cmidrule(lr){8-10}\cmidrule(lr){11-13}\cmidrule(lr){14-16}\cmidrule(lr){17-19}
    & \rotatebox{90}{\textbf{Win}} & \rotatebox{90}{\textbf{Loss}} & \rotatebox{90}{\textbf{Tie}} & \rotatebox{90}{\textbf{Win}} & \rotatebox{90}{\textbf{Loss}} & \rotatebox{90}{\textbf{Tie}} & \rotatebox{90}{\textbf{Win}} & \rotatebox{90}{\textbf{Loss}} & \rotatebox{90}{\textbf{Tie}} & \rotatebox{90}{\textbf{Win}} & \rotatebox{90}{\textbf{Loss}} & \rotatebox{90}{\textbf{Tie}} & \rotatebox{90}{\textbf{Win}} & \rotatebox{90}{\textbf{Loss}} & \rotatebox{90}{\textbf{Tie}} & \rotatebox{90}{\textbf{Win}} & \rotatebox{90}{\textbf{Loss}} & \rotatebox{90}{\textbf{Tie}} \\
  \midrule
  eng_Latn & 27.6 & 56.7 & 15.7 & 50.8 & 30.6 & 18.7 & 31.3 & 48.5 & 20.1 & 48.3 & 33.0 & 18.6 & 33.6 & 56.7 & 9.7 & 56.0 & 26.9 & 17.2 \\
  fra_Latn & 61.2 & 31.3 & 7.5 & 69.4 & 19.4 & 11.2 & 49.2 & 40.3 & 10.4 & 67.8 & 23.7 & 8.5 & 38.1 & 51.5 & 10.4 & 70.9 & 17.9 & 11.2 \\
  arb_Arab & 70.9 & 19.4 & 9.7 & 79.8 & 9.0 & 11.2 & 61.9 & 30.6 & 7.5 & 83.9 & 7.6 & 8.5 & 58.2 & 36.6 & 5.2 & 66.4 & 20.9 & 12.7 \\
  tur_Latn & 53.4 & 38.4 & 8.3 & 75.9 & 18.1 & 6.0 & 56.4 & 38.4 & 5.3 & 85.5 & 4.3 & 10.3 & 52.6 & 42.1 & 5.3 & 69.9 & 16.5 & 13.5 \\
  jpn_Jpan & 47.0 & 44.0 & 9.0 & 67.2 & 21.6 & 11.2 & 45.5 & 49.2 & 5.2 & 72.9 & 13.6 & 13.6 & 42.5 & 47.0 & 10.4 & 65.7 & 18.7 & 15.7 \\
  zho_Hans & 52.2 & 35.1 & 12.7 & 66.4 & 19.4 & 14.2 & 35.8 & 55.2 & 9.0 & 79.7 & 10.2 & 10.2 & 40.3 & 44.8 & 14.9 & 59.7 & 23.1 & 17.2 \\
  hin_Deva & 58.2 & 35.1 & 6.7 & 79.8 & 14.2 & 6.0 & 69.4 & 21.6 & 9.0 & 85.6 & 6.8 & 7.6 & 45.5 & 50.0 & 4.5 & 68.7 & 21.6 & 9.7 \\
  vie_Latn & 56.0 & 36.6 & 7.5 & 65.7 & 23.9 & 10.4 & 58.2 & 35.1 & 6.7 & 79.7 & 13.6 & 6.8 & 48.5 & 46.3 & 5.2 & 72.4 & 20.9 & 6.7 \\
  kor_Hang & 56.0 & 32.8 & 11.2 & 73.9 & 18.7 & 7.5 & 54.5 & 32.1 & 13.4 & 79.7 & 8.5 & 11.9 & 42.5 & 47.0 & 10.4 & 76.1 & 14.2 & 9.7 \\
  deu_Latn & 48.1 & 42.1 & 9.8 & 66.2 & 24.1 & 9.8 & 42.9 & 47.4 & 9.8 & 77.8 & 12.0 & 10.3 & 33.8 & 58.6 & 7.5 & 69.2 & 21.1 & 9.8 \\
  spa_Latn & 53.7 & 37.3 & 9.0 & 70.2 & 19.4 & 10.4 & 37.3 & 50.0 & 12.7 & 65.2 & 20.3 & 14.4 & 37.3 & 50.0 & 12.7 & 64.9 & 23.9 & 11.2 \\
  ind_Latn & 58.2 & 31.3 & 10.4 & 74.6 & 18.7 & 6.7 & 59.7 & 35.1 & 5.2 & 78.8 & 16.1 & 5.1 & 59.7 & 35.1 & 5.2 & 65.7 & 25.4 & 9.0 \\
  ita_Latn & 61.2 & 29.9 & 9.0 & 71.6 & 18.7 & 9.7 & 47.0 & 39.5 & 13.4 & 72.9 & 15.2 & 11.9 & 47.0 & 39.5 & 13.4 & 66.4 & 23.1 & 10.4 \\
  pol_Latn & 58.2 & 36.6 & 5.2 & 74.6 & 20.1 & 5.2 & 47.8 & 44.8 & 7.5 & 87.3 & 4.2 & 8.5 & 47.8 & 44.8 & 7.5 & 72.4 & 16.4 & 11.2 \\
  por_Latn & 55.2 & 33.6 & 11.2 & 70.9 & 22.4 & 6.7 & 49.2 & 38.1 & 12.7 & 66.1 & 21.2 & 12.7 & 49.2 & 38.1 & 12.7 & 73.1 & 15.7 & 11.2 \\
  rus_Cyrl & 50.0 & 43.3 & 6.7 & 63.4 & 25.4 & 11.2 & 41.8 & 50.0 & 8.2 & 70.3 & 16.9 & 12.7 & 41.8 & 50.0 & 8.2 & 67.9 & 18.7 & 13.4 \\
  ukr_Cyrl & 57.5 & 32.1 & 10.4 & 73.9 & 17.9 & 8.2 & 55.2 & 35.8 & 9.0 & 83.9 & 8.5 & 7.6 & 55.2 & 35.8 & 9.0 & 74.6 & 16.4 & 9.0 \\
  ces_Latn & 51.5 & 41.0 & 7.5 & 78.4 & 17.2 & 4.5 & 51.5 & 41.0 & 7.5 & 88.1 & 6.8 & 5.1 & 51.5 & 41.0 & 7.5 & 76.1 & 12.7 & 11.2 \\
  nld_Latn & 53.0 & 35.8 & 11.2 & 67.9 & 20.9 & 11.2 & 55.2 & 32.1 & 12.7 & 79.7 & 12.7 & 7.6 & 55.2 & 32.1 & 12.7 & 69.4 & 18.7 & 11.9 \\
  ell_Grek & 64.9 & 30.6 & 4.5 & 83.6 & 11.9 & 4.5 & 67.2 & 25.4 & 7.5 & 94.9 & 2.5 & 2.5 & 67.2 & 25.4 & 7.5 & 83.6 & 8.2 & 8.2 \\
  heb_Hebr & 67.2 & 28.4 & 4.5 & 87.3 & 8.2 & 4.5 & 73.9 & 18.7 & 7.5 & 90.7 & 1.7 & 7.6 & 73.9 & 18.7 & 7.5 & 75.4 & 17.9 & 6.7 \\
  pes_Arab & 67.9 & 23.9 & 8.2 & 75.4 & 17.2 & 7.5 & 61.9 & 26.9 & 11.2 & 84.8 & 5.9 & 9.3 & 61.9 & 26.9 & 11.2 & 82.8 & 9.7 & 7.5 \\
  ron_Latn & 59.0 & 32.1 & 9.0 & 73.1 & 21.6 & 5.2 & 58.2 & 31.3 & 10.4 & 83.0 & 8.5 & 8.5 & 58.2 & 31.3 & 10.4 & 68.7 & 20.9 & 10.4 \\
  \hdashline
  \textbf{avg} & 56.0 & 35.1 & 8.9 & 72.1 & 19.1 & 8.8 & 52.7 & 37.7 & 9.6 & 78.5 & 11.9 & 9.6 & 49.6 & 41.3 & 9.1 & 70.3 & 18.7 & 11.0 \\
  \bottomrule
\end{tabular}
}
  \caption{Win/Loss/Tie rates by Language for Aya-Vision-8B on AyaVisionBench}
  \label{tab:aya_vision_bench-aya_vision_8b}
\end{table}

\begin{table}[htbp]
  \centering
\resizebox{\textwidth}{!}{%
\begin{tabular}{l c c c c c c c c c c c c c c c c c c}
  \toprule
  & \multicolumn{18}{c}{\textbf{Aya-Vision-8B}} \\
  \cmidrule(lr){2-19}
  \rotatebox{90}{\textbf{Language}}
    & \multicolumn{3}{c}{\rotatebox{90}{\textbf{Gemini-Flash-1.5-8B}}}
    & \multicolumn{3}{c}{\rotatebox{90}{\textbf{Llama-3.2-11B-Vision}}}
    & \multicolumn{3}{c}{\rotatebox{90}{\textbf{Qwen-2.5-VL-7B}}}
    & \multicolumn{3}{c}{\rotatebox{90}{\textbf{Molmo-7B-D}}}
    & \multicolumn{3}{c}{\rotatebox{90}{\textbf{Pixtral-12B}}}
    & \multicolumn{3}{c}{\rotatebox{90}{\textbf{Pangea-7B}}} \\
  \cmidrule(lr){2-4}
  \cmidrule(lr){5-7}
  \cmidrule(lr){8-10}
  \cmidrule(lr){11-13}
  \cmidrule(lr){14-16}
  \cmidrule(lr){17-19}
    & \rotatebox{90}{\textbf{Win}} & \rotatebox{90}{\textbf{Loss}} & \rotatebox{90}{\textbf{Tie}}
    & \rotatebox{90}{\textbf{Win}} & \rotatebox{90}{\textbf{Loss}} & \rotatebox{90}{\textbf{Tie}}
    & \rotatebox{90}{\textbf{Win}} & \rotatebox{90}{\textbf{Loss}} & \rotatebox{90}{\textbf{Tie}}
    & \rotatebox{90}{\textbf{Win}} & \rotatebox{90}{\textbf{Loss}} & \rotatebox{90}{\textbf{Tie}}
    & \rotatebox{90}{\textbf{Win}} & \rotatebox{90}{\textbf{Loss}} & \rotatebox{90}{\textbf{Tie}}
    & \rotatebox{90}{\textbf{Win}} & \rotatebox{90}{\textbf{Loss}} & \rotatebox{90}{\textbf{Tie}} \\
  \midrule
  eng_Latn & 42.2 & 53.4 &  4.4 & 59.8 & 37.4 &  2.8 & 37.4 & 58.4 &  4.2 & 59.0 & 35.0 &  6.0 & 46.2 & 49.0 &  4.8 & 59.0 & 35.0 &  6.0 \\
  fra_Latn & 61.2 & 36.6 &  3.6 & 74.4 & 22.0 &  3.6 & 49.2 & 49.4 &  3.4 & 69.8 & 26.2 &  4.0 & 49.8 & 45.2 &  5.0 & 70.9 & 17.9 & 11.2 \\
  arb_Arab & 70.9 & 19.4 &  9.7 & 84.8 & 13.0 &  2.2 & 61.9 & 30.6 &  7.5 & 72.0 & 22.6 &  5.4 & 67.8 & 29.2 &  3.0 & 72.0 & 22.6 &  5.4 \\
  tur_Latn & 63.6 & 32.4 &  4.0 & 83.0 & 14.4 &  2.6 & 56.4 & 38.4 &  5.3 & 85.5 &  4.3 & 10.3 & 52.6 & 42.1 &  5.3 & 69.9 & 16.5 & 13.5 \\
  jpn_Jpan & 63.2 & 33.2 &  3.6 & 81.7 & 13.5 &  4.8 & 47.1 & 48.3 &  4.6 & 73.2 & 20.9 &  5.8 & 53.7 & 41.3 &  5.0 & 73.2 & 20.9 &  5.8 \\
  zho_Hans & 65.6 & 29.8 &  4.6 & 77.2 & 18.0 &  4.8 & 46.6 & 49.6 &  3.8 & 79.7 & 28.4 &  5.2 & 51.4 & 44.6 &  4.0 & 66.4 & 28.4 &  5.2 \\
  hin_Deva & 69.7 & 26.8 &  3.4 & 83.2 & 15.0 &  1.8 & 78.3 & 18.5 &  3.2 & 85.6 &  6.8 &  7.6 & 45.5 & 50.0 &  4.5 & 68.7 & 21.6 &  9.7 \\
  vie_Latn & 70.5 & 26.1 &  3.4 & 78.0 & 19.4 &  2.6 & 59.3 & 37.7 &  3.0 & 79.7 & 13.6 &  6.8 & 48.5 & 46.3 &  5.2 & 78.2 & 17.2 &  4.6 \\
  kor_Hang & 66.0 & 29.6 &  4.4 & 86.2 & 10.4 &  3.4 & 54.5 & 32.1 & 13.4 & 79.7 &  8.5 & 11.9 & 42.5 & 47.0 & 10.4 & 76.1 & 14.2 &  9.7 \\
  deu_Latn & 57.8 & 39.6 &  2.6 & 75.0 & 20.6 &  4.4 & 42.9 & 47.4 &  9.8 & 77.8 & 12.0 & 10.3 & 33.8 & 58.7 &  7.5 & 69.2 & 21.1 &  9.8 \\
  spa_Latn & 53.7 & 37.3 &  9.0 & 71.1 & 25.1 &  3.8 & 37.3 & 50.0 & 12.7 & 65.3 & 20.3 & 14.4 & 37.3 & 50.0 & 12.7 & 64.9 & 23.9 & 11.2 \\
  ind_Latn & 58.2 & 31.3 & 10.5 & 78.2 & 17.6 &  4.2 & 59.0 & 35.8 &  5.2 & 89.4 &  7.2 &  3.4 & 56.6 & 35.2 &  8.2 & 65.8 & 27.2 &  7.0 \\
  ita_Latn & 62.0 & 33.2 &  4.8 & 73.8 & 22.2 &  4.0 & 49.4 & 45.8 &  4.8 & 84.8 & 10.8 &  4.4 & 53.4 & 41.4 &  5.2 & 71.4 & 23.2 &  5.4 \\
  pol_Latn & 62.7 & 32.5 &  4.8 & 80.2 & 16.2 &  3.6 & 56.5 & 40.1 &  3.4 & 90.0 &  5.4 &  4.6 & 63.1 & 34.1 &  2.8 & 77.8 & 18.6 &  3.6 \\
  por_Latn & 62.0 & 31.0 &  7.0 & 74.2 & 21.6 &  4.2 & 48.4 & 45.4 &  6.2 & 66.1 & 21.2 & 12.7 & 50.6 & 41.8 &  7.6 & 66.8 & 25.6 &  7.6 \\
  rus_Cyrl & 65.0 & 32.8 &  2.2 & 81.9 & 14.3 &  3.8 & 56.1 & 41.3 &  2.6 & 85.9 &  8.7 &  5.4 & 56.3 & 40.2 &  3.4 & 70.8 & 23.9 &  5.2 \\
  ukr_Cyrl & 62.5 & 34.3 &  3.2 & 82.4 & 13.2 &  4.4 & 58.3 & 37.1 &  4.6 & 92.6 &  4.6 &  2.8 & 69.9 & 25.9 &  4.2 & 80.2 & 16.2 &  3.6 \\
  ces_Latn & 63.4 & 30.0 &  6.6 & 79.2 & 15.0 &  5.8 & 60.0 & 36.4 &  3.6 & 88.0 &  6.8 &  5.4 & 63.8 & 30.8 &  5.4 & 80.4 & 14.6 &  5.0 \\
  nld_Latn & 63.0 & 33.6 &  3.4 & 77.8 & 17.6 &  4.6 & 52.8 & 43.0 &  4.2 & 91.0 &  6.0 &  3.0 & 57.0 & 37.8 &  5.2 & 76.8 & 18.8 &  4.4 \\
  ell_Grek & 75.2 & 22.0 &  2.8 & 84.4 & 12.6 &  3.0 & 73.8 & 23.2 &  3.0 & 95.2 &  3.2 &  1.6 & 75.0 & 20.8 &  4.2 & 90.0 &  7.4 &  2.6 \\
  heb_Hebr & 70.0 & 26.0 &  4.0 & 85.2 & 11.2 &  3.6 & 77.8 & 18.8 &  3.4 & 92.0 &  4.6 &  3.4 & 70.4 & 25.0 &  4.6 & 73.2 & 22.6 &  4.2 \\
  pes_Arab & 76.8 & 19.8 &  3.4 & 88.2 &  9.4 &  2.4 & 72.3 & 24.7 &  3.0 & 93.4 &  3.6 &  3.0 & 76.8 & 19.0 &  4.2 & 86.4 & 10.0 &  3.6 \\
  ron_Latn & 63.1 & 31.9 &  5.0 & 78.4 & 15.8 &  5.8 & 60.3 & 35.1 &  4.6 & 89.2 &  6.4 &  4.4 & 63.7 & 30.9 &  5.4 & 68.3 & 27.7 &  4.0 \\
  \hdashline
 \textbf{avg}
    & 64.5 & 31.4 &  4.0 & 79.1 & 17.2 &  3.8 & 57.5 & 38.4 &  4.1
    & 80.3 & 15.3 &  4.5 & 59.4 & 35.7 &  5.0 & 73.1 & 22.1 &  4.8 \\
  \bottomrule
\end{tabular}
}
  \caption{Win/Loss/Tie rates by Language for Aya-Vision-8B on m-WildVision}
  \label{tab:wildvision-aya_vision_8b}
\end{table}

\begin{table}[htbp]
  \centering
  \begin{tabular}{l c c c c c c c c c}
    \toprule
    & \multicolumn{9}{c}{\textbf{Aya-Vision-32B}} \\
    \cmidrule(lr){2-10}
    \textbf{Language} & \multicolumn{3}{c}{\textbf{Llama-3.2-90B-Vision}} & \multicolumn{3}{c}{\textbf{Molmo-72B}} & \multicolumn{3}{c}{\textbf{Qwen-2.5-VL-72B}} \\
    \cmidrule(lr){2-4}\cmidrule(lr){5-7}\cmidrule(lr){8-10}
    & \textbf{Win} & \textbf{Loss} & \textbf{Tie}& \textbf{Win} & \textbf{Loss} & \textbf{Tie}& \textbf{Win} & \textbf{Loss} & \textbf{Tie} \\
    \midrule
    eng\_Latn & 26.2 & 73.6 & 0.2 & 66.0 & 32.8 & 1.2 & 35.8 & 63.6 & 0.6 \\
    fra\_Latn & 39.6 & 60.4 & 0.0 & 72.2 & 27.6 & 0.2 & 46.8 & 52.8 & 0.4 \\
    hin\_Deva & 47.4 & 52.0 & 0.6 & 86.0 & 14.0 & 0.0 & 69.2 & 30.8 & 0.0 \\
    arb\_Arab & 54.2 & 45.2 & 0.6 & 81.4 & 18.6 & 0.0 & 59.6 & 40.4 & 0.0 \\
    tur\_Latn & 45.2 & 54.4 & 0.4 & 78.6 & 20.8 & 0.6 & 51.4 & 48.2 & 0.4 \\
    jpn\_Jpan & 47.2 & 52.4 & 0.4 & 84.2 & 15.8 & 0.0 & 54.8 & 44.6 & 0.6 \\
    zho\_Hans & 42.8 & 57.0 & 0.2 & 75.2 & 24.6 & 0.2 & 43.6 & 55.6 & 0.8 \\
    vie\_Latn & 41.8 & 58.0 & 0.2 & 77.0 & 22.6 & 0.4 & 55.0 & 44.8 & 0.2 \\
    kor\_Hang & 51.6 & 48.4 & 0.0 & 78.6 & 21.2 & 0.2 & 56.4 & 43.6 & 0.0 \\
    deu\_Latn & 40.4 & 59.6 & 0.0 & 78.6 & 21.0 & 0.4 & 47.4 & 51.8 & 0.8 \\
    ind\_Latn & 39.8 & 59.8 & 0.4 & 76.4 & 23.2 & 0.4 & 49.2 & 50.4 & 0.4 \\
    ita\_Latn & 41.0 & 59.0 & 0.0 & 75.2 & 24.2 & 0.6 & 38.2 & 61.2 & 0.6 \\
    pol\_Latn & 42.2 & 57.6 & 0.2 & 75.4 & 24.0 & 0.6 & 43.4 & 56.4 & 0.2 \\
    por\_Latn & 35.2 & 64.6 & 0.2 & 70.6 & 29.0 & 0.4 & 44.6 & 55.4 & 0.0 \\
    rus\_Cyrl & 40.0 & 60.0 & 0.0 & 66.8 & 33.0 & 0.2 & 47.6 & 52.0 & 0.4 \\
    spa\_Latn & 38.8 & 60.8 & 0.4 & 69.2 & 30.6 & 0.2 & 45.4 & 54.0 & 0.6 \\
    ukr\_Cyrl & 44.6 & 55.2 & 0.2 & 80.0 & 20.0 & 0.0 & 48.0 & 51.8 & 0.2 \\
    ces\_Latn & 45.6 & 54.2 & 0.2 & 75.6 & 24.4 & 0.0 & 53.0 & 47.0 & 0.0 \\
    nld\_Latn & 42.0 & 57.2 & 0.8 & 76.8 & 23.2 & 0.0 & 46.8 & 52.6 & 0.6 \\
    ell\_Grek & 46.2 & 53.6 & 0.2 & 84.2 & 15.4 & 0.4 & 62.4 & 37.2 & 0.4 \\
    heb\_Hebr & 51.2 & 48.6 & 0.2 & 85.8 & 14.0 & 0.2 & 63.4 & 36.6 & 0.0 \\
    pes\_Arab & 51.0 & 48.8 & 0.2 & 84.4 & 15.0 & 0.6 & 57.6 & 42.4 & 0.0 \\
    ron\_Latn & 40.4 & 59.2 & 0.4 & 78.8 & 21.0 & 0.2 & 51.6 & 48.2 & 0.2 \\
    \hdashline
   \textbf{avg} & 43.2 & 56.5 & 0.3 & 77.3 & 22.4 & 0.3 & 50.9 & 48.8 & 0.3 \\
    \bottomrule
  \end{tabular}
    \caption{Win/Loss/Tie rates by Language for Aya-Vision-32B on m-ArenaHard}
  \label{tab:marena-Aya_Vision-32B}
\end{table}

\begin{table}[htbp]
  \centering
  \begin{tabular}{l c c c c c c c c c}
    \toprule
    & \multicolumn{9}{c}{\textbf{Aya-Vision-32B}} \\
    \cmidrule(lr){2-10}
    \textbf{Language}
      & \multicolumn{3}{c}{\textbf{Llama-3.2-90B-Vision}}
      & \multicolumn{3}{c}{\textbf{Molmo-72B}}
      & \multicolumn{3}{c}{\textbf{Qwen-2.5-VL-72B}} \\
    \cmidrule(lr){2-4}\cmidrule(lr){5-7}\cmidrule(lr){8-10}
      & \textbf{Win} & \textbf{Loss} & \textbf{Tie}
      & \textbf{Win} & \textbf{Loss} & \textbf{Tie}
      & \textbf{Win} & \textbf{Loss} & \textbf{Tie} \\
    \midrule
    eng\_Latn   & 49.25 & 38.81 & 11.94 & 35.82 & 54.48 &  9.70 & 62.69 & 24.63 & 12.69 \\
    fra\_Latn   & 64.93 & 24.63 & 10.45 & 53.73 & 39.55 &  6.72 & 49.25 & 42.54 &  8.21 \\
    hin\_Deva   & 74.63 & 23.13 &  2.24 & 72.39 & 25.37 &  2.24 & 35.82 & 61.19 &  2.99 \\
    arb\_Arab   & 70.90 & 19.40 &  9.70 & 73.13 & 20.90 &  5.97 & 44.03 & 47.76 &  8.21 \\
    tur\_Latn   & 63.91 & 30.08 &  6.02 & 64.66 & 30.08 &  5.26 & 52.63 & 44.36 &  3.01 \\
    jpn\_Jpan   & 61.94 & 28.36 &  9.70 & 61.94 & 35.82 &  2.24 & 48.51 & 45.52 &  5.97 \\
    zho\_Hans   & 65.67 & 28.36 &  5.97 & 66.42 & 26.87 &  6.72 & 44.03 & 46.27 &  9.70 \\
    vie\_Latn   & 64.93 & 24.63 & 10.45 & 50.75 & 42.54 &  6.72 & 52.99 & 41.04 &  5.97 \\
    kor\_Hang   & 64.93 & 28.36 &  6.72 & 58.96 & 33.58 &  7.46 & 44.78 & 44.78 & 10.45 \\
    deu\_Latn   & 69.92 & 21.80 &  8.27 & 60.15 & 33.83 &  6.02 & 48.87 & 48.12 &  3.01 \\
    ind\_Latn   & 68.66 & 26.87 &  4.48 & 56.72 & 37.31 &  5.97 & 47.76 & 44.78 &  7.46 \\
    ita\_Latn   & 62.69 & 29.85 &  7.46 & 55.97 & 35.07 &  8.96 & 52.99 & 39.55 &  7.46 \\
    pol\_Latn   & 74.63 & 20.90 &  4.48 & 65.67 & 28.36 &  5.97 & 48.51 & 45.52 &  5.97 \\
    por\_Latn   & 52.99 & 41.79 &  5.22 & 51.49 & 42.54 &  5.97 & 54.48 & 36.57 &  8.96 \\
    rus\_Cyrl   & 60.45 & 29.10 & 10.45 & 50.75 & 40.30 &  8.96 & 50.75 & 41.04 &  8.21 \\
    spa\_Latn   & 61.19 & 29.85 &  8.96 & 52.99 & 37.31 &  9.70 & 50.75 & 43.28 &  5.97 \\
    ukr\_Cyrl   & 75.37 & 20.90 &  3.73 & 61.94 & 32.84 &  5.22 & 50.75 & 43.28 &  5.97 \\
    ces\_Latn   & 73.88 & 20.15 &  5.97 & 67.91 & 27.61 &  4.48 & 50.75 & 46.27 &  2.99 \\
    nld\_Latn   & 64.93 & 24.63 & 10.45 & 52.24 & 42.54 &  5.22 & 50.00 & 45.52 &  4.48 \\
    ell\_Grek   & 66.42 & 26.12 &  7.46 & 78.36 & 17.91 &  3.73 & 38.81 & 51.49 &  9.70 \\
    heb\_Hebr   & 68.66 & 24.63 &  6.72 & 68.66 & 26.87 &  4.48 & 42.54 & 51.49 &  5.97 \\
    pes\_Arab   & 70.90 & 23.88 &  5.22 & 78.36 & 18.66 &  2.99 & 46.27 & 50.00 &  3.73 \\
    ron\_Latn   & 64.18 & 31.34 &  4.48 & 68.66 & 26.87 &  4.48 & 47.01 & 45.52 &  7.46 \\
    \hdashline
    \textbf{avg}
      & 65.91 & 26.85 &  7.24 & 61.20 & 32.92 &  5.88 & 48.48 & 44.81 &  6.72 \\
    \bottomrule
  \end{tabular}
    \caption{Win/Loss/Tie rates by Language for Aya-Vision-32B on AyaVisionBench}
  \label{tab:aya-vision-bench-aya_vision_32b}
\end{table}

\begin{table}[htbp]
  \centering
  \begin{tabular}{l c c c c c c c c c}
    \toprule
    & \multicolumn{9}{c}{\textbf{Aya-Vision-32B}} \\
    \cmidrule(lr){2-10}
    \textbf{Language}
      & \multicolumn{3}{c}{\textbf{Qwen-2.5-VL-72B}}
      & \multicolumn{3}{c}{\textbf{Llama-3.2-90B-Vision}}
      & \multicolumn{3}{c}{\textbf{Molmo-72B}} \\
    \cmidrule(lr){2-4}\cmidrule(lr){5-7}\cmidrule(lr){8-10}
      & \textbf{Win} & \textbf{Loss} & \textbf{Tie}
      & \textbf{Win} & \textbf{Loss} & \textbf{Tie}
      & \textbf{Win} & \textbf{Loss} & \textbf{Tie} \\
    \midrule
    eng\_Latn   & 37.4 & 56.4 &  6.2 & 67.6 & 29.2 &  3.2 & 56.2 & 39.2 &  4.6 \\
    fra\_Latn   & 46.2 & 50.0 &  3.8 & 69.9 & 26.4 &  3.6 & 59.0 & 37.2 &  3.8 \\
    hin\_Deva   & 67.4 & 30.6 &  2.0 & 78.4 & 17.6 &  4.0 & 75.6 & 20.0 &  4.4 \\
    arb\_Arab   & 57.4 & 39.2 &  3.4 & 79.0 & 17.8 &  3.2 & 79.2 & 16.8 &  4.0 \\
    tur\_Latn   & 56.0 & 39.6 &  4.4 & 77.8 & 19.0 &  3.2 & 76.5 & 20.5 &  3.0 \\
    jpn\_Jpan   & 49.0 & 46.4 &  4.6 & 72.2 & 25.4 &  2.4 & 76.2 & 20.2 &  3.6 \\
    zho\_Hans   & 39.0 & 56.4 &  4.6 & 77.0 & 19.0 &  4.0 & 78.0 & 19.6 &  2.4 \\
    vie\_Latn   & 57.4 & 38.6 &  4.0 & 76.6 & 21.4 &  2.0 & 64.2 & 31.6 &  4.2 \\
    kor\_Hang   & 55.4 & 40.8 &  3.8 & 75.4 & 21.0 &  3.6 & 70.4 & 25.2 &  4.4 \\
    deu\_Latn   & 49.2 & 46.4 &  4.4 & 67.0 & 28.6 &  4.4 & 68.0 & 28.0 &  4.0 \\
    ind\_Latn   & 51.0 & 45.8 &  3.2 & 72.0 & 26.0 &  2.0 & 65.2 & 30.0 &  4.8 \\
    ita\_Latn   & 46.2 & 49.0 &  4.8 & 69.8 & 26.2 &  4.0 & 59.0 & 33.8 &  7.2 \\
    pol\_Latn   & 50.8 & 46.8 &  2.4 & 73.6 & 23.4 &  3.0 & 67.2 & 29.0 &  3.8 \\
    por\_Latn   & 49.2 & 45.8 &  5.0 & 68.2 & 26.8 &  5.0 & 61.2 & 33.6 &  5.2 \\
    rus\_Cyrl   & 50.2 & 47.2 &  2.6 & 73.2 & 23.6 &  3.2 & 60.3 & 36.3 &  3.4 \\
    spa\_Latn   & 48.6 & 46.6 &  4.8 & 65.2 & 30.6 &  4.2 & 57.0 & 37.8 &  5.2 \\
    ukr\_Cyrl   & 58.4 & 38.8 &  2.8 & 74.4 & 21.4 &  4.2 & 70.6 & 25.4 &  4.0 \\
    ces\_Latn   & 54.4 & 42.2 &  3.4 & 69.6 & 27.2 &  3.2 & 67.6 & 28.8 &  3.6 \\
    nld\_Latn   & 47.6 & 48.8 &  3.6 & 69.4 & 25.8 &  4.8 & 61.4 & 33.8 &  4.8 \\
    ell\_Grek  & 66.6 & 30.2 &  3.2 & 75.0 & 22.0 &  3.0 & 84.2 & 11.8 &  4.0 \\
    heb\_Hebr   & 66.0 & 30.6 &  3.4 & 74.2 & 22.8 &  3.0 & 74.0 & 22.4 &  3.6 \\
    pes\_Arab  & 64.4 & 30.8 &  4.8 & 80.6 & 16.6 &  2.8 & 77.6 & 18.4 &  4.0 \\
    ron\_Latn  & 58.0 & 39.2 &  2.8 & 73.6 & 24.4 &  2.0 & 74.6 & 21.8 &  3.6 \\
    \hdashline
    \textbf{avg}
      & 53.3 & 42.9 &  3.8
      & 73.0 & 23.6 &  3.4
      & 68.8 & 27.0 &  4.2 \\
    \bottomrule
  \end{tabular}
    \caption{Win/Loss/Tie rates by Language for Aya-Vision-32B on m-WildVision.}
  \label{tab:mwildvision-aya_vision_32b}
\end{table}

\definecolor{highlightcolor}{RGB}{211, 227, 252}

\begin{table}
    \centering
    \resizebox{1.0\textwidth}{!}{
    \begin{tabular}{lcccccccccccc}
    \toprule
     & eng_Latn & fra_Latn & heb_Hebr & hin_Deva & ron_Latn & tha_Thai & zho_Hans & \textbf{avg} \\
    \midrule
    Pangea-7B & 55.30 & 43.60 & 59.30 & 53.50 & 45.80 & 67.20 & 50.20 & 53.56  \\
    Molmo-7B-D & 68.09 & 54.17 & 34.29 & 31.92 & 30.28 & 53.73 & 46.21 & 45.53 \\
    Llama-3.2-11B-Vision & 56.03 & 45.08 & 31.07 & 45.00 & 38.38 & 42.16 & 20.22 & 39.71 \\
    Pixtral-12B & 57.20 & 43.56 & 40.00 & 55.38 & 41.20 & 55.97 & 29.24 & 46.08 \\
    Qwen-2.5-VL-7B & 57.98 & 52.65 & 54.29 & 54.62 & 44.72 & 67.16 & 51.62 & 54.72 \\
    \rowcolor{highlightcolor} Aya-Vision-8B & 57.59 & 54.92 & 58.57 & 66.92 & 54.93 & 33.21 & 56.32 & 54.64 \\
    \midrule
    Molmo-72B & 59.92 & 54.92 & 58.21 & 62.69 & 50.70 & 65.30 & 47.29 & 57.01 \\
    Llama-3.2-90B-Vision & 75.00 & 67.05 & 59.64 & 70.38 & 59.51 & 68.66 & 53.43 & 64.81 \\
    Qwen-2.5-VL-72B & 55.25 & 49.62 & 62.86 & 66.15 & 46.13 & 74.25 & 58.48 & 58.96 \\
     \rowcolor{highlightcolor}Aya-Vision-32B & 55.64 & 60.61 & 66.43 & 71.54 & 57.75 & 43.07 & 61.73 & 59.54 \\
    \bottomrule
    \end{tabular}
    }
    \caption{MaxM}
    \label{tab:maxm}
\end{table}
\definecolor{highlightcolor}{RGB}{211, 227, 252}

\begin{table}
    \centering
    \resizebox{1.0\textwidth}{!}{
    \begin{tabular}{lcccccccccccc}
    \toprule
     & fra_Latn & jpn_Jpan & ind_Latn & por_Latn & hin_Deva & arb_Arab & eng_Latn & \textbf{avg} \\
    \midrule
    Pangea-7B & 45.30 & 40.50 & 46.50 & 46.10 & 41.60 & 42.30 & 45.70 & 44.00 \\ 
    Molmo-7B-D & 38.90 & 37.10 & 38.90 & 38.10 & 34.90 & 36.70 & 40.50 & 37.87 \\
    Llama-3.2-11B-Vision & 43.30 & 40.90 & 42.10 & 44.10 & 39.90 & 41.60 & 47.20 & 42.73 \\
    Pixtral-12B & 47.00 & 43.90 & 40.10 & 47.80 & 32.60 & 36.20 & 48.30 & 42.27 \\
    Qwen-2.5-VL-7B & 49.70 & 46.10 & 47.80 & 49.80 & 41.20 & 41.70 & 51.10 & 46.77 \\
     \rowcolor{highlightcolor}Aya-Vision-8B & 40.20 & 41.40 & 39.50 & 38.50 & 38.10 & 40.10 & 41.80 & 39.94 \\
     \midrule
    Molmo-72B & 52.80 & 49.00 & 52.80 & 55.40 & 48.00 & 51.20 & 51.50 & 51.53 \\
    Llama-3.2-90B-Vision & 56.60 & 52.90 & 55.20 & 54.30 & 46.60 & 45.00 & 56.20 & 52.40 \\
    Qwen-2.5-VL-72B & 62.40 & 60.60 & 64.00 & 62.00 & 60.80 & 59.70 & 62.70 & 61.74 \\
     \rowcolor{highlightcolor} Aya-Vision-32B & 44.90 & 42.90 & 46.60 & 45.30 & 45.00 & 44.10 & 47.00 & 45.11 \\
    \bottomrule
    \end{tabular}
    }
    \caption{xMMMU}
    \label{tab:xmmmu}
\end{table}
\definecolor{highlightcolor}{RGB}{211, 227, 252}

\begin{table}
    \centering
    \resizebox{1.0\textwidth}{!}{
    \begin{tabular}{lcccccccccccc}
    \toprule
     & arb_Arab & deu_Latn & fra_Latn & ita_Latn & jpn_Jpan & kor_Hang & rus_Cyrl & vie_Latn & tha_Thai & \textbf{avg} \\
    \midrule
    Pangea-7B & 8.53 & 29.96 & 32.39 & 23.87 & 9.30 & 13.44 & 7.67 & 21.38 &  15.15 & 17.97 \\ 
    Molmo-7B-D & 5.83 & 26.24 & 35.67 & 29.86 & 7.61 & 9.86 & 5.03 & 15.05 & 15.15 & 16.70 \\
    Llama-3.2-11B-Vision & 7.97 & 24.24 & 27.99 & 22.85 & 10.75 & 13.08 & 7.01 & 17.31 & 16.88 & 16.45 \\
    Pixtral-12B & 7.68 & 32.54 & 37.92 & 32.69 & 8.33 & 13.08 & 7.14 & 19.12 & 14.29 & 19.20 \\
    Qwen-2.5-VL-7B & 19.26 & 35.31 & 42.66 & 36.76 & 21.98 & 32.80 & 10.45 & 37.33 & 22.51 & 28.78 \\
     \rowcolor{highlightcolor} Aya-Vision-8B & 13.69 & 28.72 & 35.89 & 28.39 & 10.51 & 13.08 & 6.35 & 17.99 & 7.79 & 18.05 \\
     \midrule
    Molmo-72B & 6.54 & 30.34 & 35.44 & 30.54 & 9.42 & 10.04 & 8.73 & 18.21 & 17.32 & 18.51 \\
    Llama-3.2-90B-Vision & 19.91 & 36.35 & 40.29 & 35.29 & 17.27 & 30.11 & 10.98 & 29.30 & 25.97 & 27.28 \\
    Qwen-2.5-VL-72B & 23.19 & 35.78 & 43.91 & 39.14 & 21.98 & 35.66 & 12.83 & 42.87 & 27.27 & 31.40 \\
     \rowcolor{highlightcolor} Aya-Vision-32B & 116.33 & 34.83 & 40.52 & 32.20 & 15.03 & 14.57 & 10.28 & 23.91 & 11.45 & 22.12 \\
    \bottomrule
    \end{tabular}
    }
    \caption{MTVQA}
    \label{tab:mtvqa}
\end{table}
\definecolor{highlightcolor}{RGB}{211, 227, 252}

\begin{table}
    \centering
    \resizebox{1.0\textwidth}{!}{
    \begin{tabular}{lccccccccccccc}
    \toprule
     & hin_Deva & ind_Latn & kor_Hang & spa_Latn & eng_Latn & zho_Hans & jpn_Jpan & \textbf{avg} \\
    \midrule
    Pangea-7B & 29.00 & 36.50 & 28.50 & 34.00 & 26.50 & 36.00 & 35.00 & 32.21 \\
    Molmo-7B-D & 4.00 & 24.50 & 8.50 & 42.50 & 65.50 & 2.00 & 16.50 & 23.36 \\
    Llama-3.2-11B-Vision & 13.00 & 35.50 & 13.78 & 43.00 & 55.50 & 23.00 & 16.33 & 28.59 \\
    Pixtral-12B & 50.50 & 66.50 & 60.00 & 72.50 & 74.00 & 64.00 & 64.00 & 64.50 \\
    Qwen-2.5-VL-7B & 20.50 & 58.50 & 53.00 & 66.50 & 78.00 & 71.50 & 59.00 & 58.14 \\
    \rowcolor{highlightcolor} Aya-Vision-8B & 56.50 & 60.50 & 56.00 & 60.00 & 60.50 & 55.50 & 61.50 & 58.64 \\
    \midrule
    Molmo-72B & 19.5 & 53.5 & 27.0 & 64.5 & 65.5 & 42.5 & 45.5 & 45.43 \\
    Llama-3.2-90B-Vision & 38.50 & 54.50 & 42.35 & 60.50 & 63.00 & 53.00 & 46.00 & 51.12 \\
    Qwen-2.5-VL-72B & 44.50 & 77.00 & 71.94 & 80.50 & 82.00 & 71.00 & 71.00 & 71.13 \\
    \rowcolor{highlightcolor} Aya-Vision-32B & 68.50 & 72.00 & 62.50 & 77.00 & 72.50 & 66.50 & 71.50 & 70.07 \\
    \bottomrule
    \end{tabular}
    }

    \caption{xChatBench}
    \label{tab:xchat}
\end{table}
\definecolor{highlightcolor}{RGB}{211, 227, 252}

\begin{table}
    \centering
    \resizebox{1.0\textwidth}{!}{
    \begin{tabular}{lccccccccccccc}
    \toprule
     & tha_Thai & tel_Telu & ben_Beng & eng_Latn & spa_Latn & jpn_Jpan & zho_Hans & swh_Latn & deu_Latn & rus_Cyrl & fra_Latn & \textbf{avg} \\
    \midrule
    Pangea-7B & 49.60 & 5.60 & 0.00 & 82.00 &  74.8 & 22.00 & 68.00 & 54.0 &  68.4 &   68.0 & 63.2 & 50.51 \\
    Molmo-7B-D & 24.50 & 2.41 & 6.02 & 73.90 & 39.36 & 41.77 & 58.06 & 0.00 & 52.61 & 47.79 & 36.14 & 34.78 \\
    Llama-3.2-11B-Vision & 64.26 & 6.88 & 18.88 & 84.74 & 71.89 & 55.24 & 73.90 & 56.63 & 76.31 & 77.11 & 70.68 & 59.68 \\
    Pixtral-12B & 63.86 & 36.55 & 57.83 & 89.16 & 82.73 & 64.66 & 73.90 & 23.69 & 79.92 & 78.71 & 74.30 & 65.94 \\
    Qwen-2.5-VL-7B & 58.44 & 4.42 & 37.75 & 85.14 & 43.37 & 61.85 & 72.29 & 4.09 & 74.30 & 63.27 & 26.10 & 48.27 \\
    \rowcolor{highlightcolor} Aya-Vision-8B & 12.45 & 0.00 & 6.83 & 84.34 & 77.91 & 67.87 & 74.70 & 4.90 & 75.90 & 80.72 & 73.49 & 50.83 \\
    \midrule
    Molmo-72B & 79.52 & 11.65 & 55.82 & 96.39 & 89.56 & 69.08 & 86.35 & 57.03 & 88.76 & 90.76 & 81.12 & 73.27 \\
    Llama-3.2-90B-Vision & 84.34 & 7.63 & 26.51 & 96.39 & 26.91 & 81.53 & 77.91 & 82.73 & 89.96 & 87.95 & 6.02 & 60.72 \\
    Qwen-2.5-VL-72B & 87.95 & 13.25 & 64.26 & 95.18 & 93.17 & 86.35 & 91.16 & 65.06 & 89.52 & 91.57 & 80.32 & 77.98 \\
    \rowcolor{highlightcolor} Aya-Vision-32B & 39.36 & 0.00 & 14.46 & 87.95 & 82.33 & 75.50 & 80.32 & 23.69 & 81.53 & 76.31 & 72.29 & 57.61 \\
    \bottomrule
    \end{tabular}
    }
    \caption{MGSM}
    \label{tab:mgsm}
\end{table}

\definecolor{highlightcolor}{RGB}{211, 227, 252}

\begin{sidewaystable}
    \resizebox{1.0\textwidth}{!}{
    \begin{tabular}{lccccc>{\columncolor{highlightcolor}}cccc>{\columncolor{highlightcolor}}c}
    \toprule
     & Pangea-7B & Molmo-7B-D & Llama-3.2-11B-Vision & Pixtral-12B & Qwen-2.5-VL-7B & Aya-Vision-8B & Molmo-72B & Llama-3.2-90B-Vision & Qwen-2.5-VL-72B & Aya-Vision-32B \\
    \midrule
    swh_Latn & 39.25 & 22.75 & 29.75 & 44.36 & 33.50 & 29.00 & 52.75 & 57.75 & 54.25 & 36.50 \\
    spa_Latn & 54.25 & 44.25 & 66.75 & 69.00 & 68.75 & 65.25 & 73.50 & 80.00 & 81.00 & 62.91 \\
    jpn_Jpan & 39.75 & 32.00 & 51.00 & 59.90 & 61.75 & 59.75 & 67.50 & 78.50 & 82.50 & 67.50 \\
    kor_Hang & 46.75 & 33.00 & 50.75 & 64.75 & 61.25 & 56.50 & 73.25 & 75.50 & 79.75 & 64.25 \\
    deu_Latn & 54.00 & 41.10 & 66.50 & 67.50 & 64.75 & 67.75 & 77.75 & 80.25 & 82.00 & 68.75 \\
    por_Latn & 55.75 & 44.00 & 64.25 & 69.10 & 65.75 & 65.00 & 74.75 & 84.50 & 83.75 & 62.25 \\
    zho_Hans & 53.75 & 45.00 & 63.50 & 68.09 & 66.08 & 63.75 & 63.25 & 73.75 & 80.75 & 61.50 \\
    ben_Beng & 40.25 & 29.25 & 30.25 & 55.75 & 53.25 & 40.25 & 64.50 & 61.00 & 73.91 & 48.50 \\
    eng_Latn & 65.00 & 49.00 & 71.25 & 74.94 & 72.43 & 71.00 & 74.25 & 83.75 & 87.75 & 69.25 \\
    ind_Latn & 47.75 & 36.00 & 64.75 & 59.00 & 61.46 & 58.75 & 74.75 & 81.50 & 81.25 & 65.50 \\
    hin_Deva & 39.00 & 32.50 & 48.50 & 59.55 & 54.00 & 55.00 & 66.75 & 71.50 & 75.75 & 42.25 \\
    arb_Arab & 38.75 & 33.75 & 52.50 & 62.50 & 62.03 & 59.00 & 57.25 & 74.50 & 77.14 & 67.50 \\
    fra_Latn & 45.00 & 44.25 & 64.50 & 68.75 & 68.17 & 63.50 & 72.50 & 62.25 & 82.96 & 68.00 \\
    yor_Latn & 20.25 & 27.00 & 15.25 & 29.55 & 30.00 & 29.75 & 35.50 & 40.50 & 37.50 & 29.00 \\
    ita_Latn & 52.50 & 40.75 & 64.75 & 70.03 & 71.43 & 65.00 & 76.75 & 83.50 & 83.25 & 63.25 \\
    \hdashline
    \textbf{avg}      & 46.13 & 36.97 & 53.62 & 61.52 & 59.64 & 56.62 & 67.00 & 72.58 & 76.23 & 58.46 \\
    \bottomrule
    \end{tabular}
    }
    \caption{global MMLU}
    \label{tab:global_mmlu}
\end{sidewaystable}
\definecolor{highlightcolor}{RGB}{211, 227, 252}

\begin{sidewaystable}
    \centering
    \resizebox{1.0\textwidth}{!}{
    \begin{tabular}{lccccc>{\columncolor{highlightcolor}}cccc>{\columncolor{highlightcolor}}c}
    \toprule
     & Pangea-7B & Molmo-7B-D & Llama-3.2-11B-Vision & Pixtral-12B & Qwen-2.5-VL-7B & Aya-Vision-8B & Molmo-72B & Llama-3.2-90B-Vision & Qwen-2.5-VL-72B & Aya-Vision-32B \\
    \midrule
    eng_Latn->arb_Arab & 27.50 & 11.26 & 26.62 & 21.90 & 24.79 & 38.22 & 32.05 & 36.78 & 36.17 & 38.93 \\
    eng_Latn->heb_Hebr & 27.36 & 11.07 & 28.32 & 23.68 & 19.92 & 38.25 & 30.52 & 40.87 & 32.02 & 41.85 \\
    eng_Latn->por_Latn & 47.01 & 31.52 & 49.24 & 50.88 & 47.69 & 51.41 & 50.69 & 54.33 & 53.93 & 52.30 \\
    eng_Latn->jpn_Jpan & 22.71 & 11.50 & 22.20 & 19.08 & 22.98 & 26.95 & 25.79 & 28.23 & 29.58 & 29.10 \\
    eng_Latn->hin_Deva & 20.26 & 6.20  & 27.05 & 20.88 & 13.37 & 29.13 & 23.10 & 34.45 & 24.96 & 30.39 \\
    eng_Latn->fra_Latn & 46.68 & 35.49 & 48.79 & 49.68 & 45.37 & 51.42 & 49.84 & 53.81 & 52.83 & 52.17 \\
    eng_Latn->ita_Latn & 28.62 & 19.36 & 31.40 & 32.01 & 28.09 & 33.60 & 32.34 & 34.90 & 33.06 & 36.19 \\
    eng_Latn->rus_Cyrl & 31.08 & 22.31 & 33.58 & 36.53 & 32.28 & 37.22 & 37.43 & 39.67 & 40.48 & 38.80 \\
    eng_Latn->zho_Hans & 31.53 & 21.22 & 28.82 & 24.64 & 34.01 & 33.28 & 36.57 & 35.74 & 38.41 & 34.57 \\
    eng_Latn->ind_Latn & 40.46 & 20.05 & 39.66 & 35.33 & 35.76 & 43.29 & 39.94 & 45.89 & 45.37 & 44.46 \\
    eng_Latn->spa_Latn & 27.87 & 21.15 & 28.33 & 29.59 & 27.41 & 31.11 & 30.32 & 30.84 & 31.03 & 32.17 \\
    eng_Latn->pes_Arab & 14.60 & 8.46  & 27.43 & 20.94 & 18.37 & 30.31 & 24.73 & 33.71 & 27.57 & 31.99 \\
    eng_Latn->tur_Latn & 25.82 & 8.23  & 27.36 & 21.50 & 22.16 & 30.99 & 26.95 & 37.01 & 32.10 & 34.17 \\
    eng_Latn->vie_Latn & 35.55 & 20.18 & 36.98 & 32.29 & 35.20 & 40.15 & 37.29 & 41.88 & 41.17 & 40.38 \\
    eng_Latn->pol_Latn & 19.89 & 11.24 & 25.59 & 22.57 & 22.67 & 28.56 & 26.03 & 30.27 & 28.39 & 30.35 \\
    eng_Latn->ell_Grek & 12.27 & 5.14  & 26.23 & 22.63 & 17.28 & 34.06 & 20.68 & 33.77 & 25.70 & 36.46 \\
    eng_Latn->ron_Latn & 37.39 & 15.83 & 40.17 & 33.91 & 31.78 & 43.51 & 35.85 & 47.82 & 41.55 & 47.06 \\
    eng_Latn->deu_Latn & 34.45 & 21.54 & 39.59 & 41.33 & 36.73 & 40.97 & 41.49 & 45.62 & 43.48 & 44.45 \\
    eng_Latn->kor_Hang & 18.76 & 8.69  & 20.56 & 19.00 & 18.01 & 25.96 & 23.37 & 25.92 & 26.23 & 27.44 \\
    eng_Latn->ces_Latn & 23.85 & 12.34 & 33.97 & 29.70 & 28.59 & 36.25 & 33.26 & 40.81 & 36.71 & 38.53 \\
    eng_Latn->nld_Latn & 24.02 & 15.67 & 28.94 & 26.00 & 27.17 & 31.37 & 30.59 & 33.41 & 31.63 & 33.20 \\
    eng_Latn->ukr_Cyrl & 19.24 & 7.90  & 29.56 & 30.41 & 26.02 & 33.77 & 26.55 & 35.85 & 33.29 & 36.40 \\
    \hdashline
    \textbf{avg}                & 28.04 & 15.74 & 31.84 & 29.29 & 27.98 & 35.90 & 32.52 & 38.25 & 35.71 & 37.79 \\
    \bottomrule
    \end{tabular}
    }
    \caption{flores}
    \label{tab:flores}
\end{sidewaystable}
\definecolor{highlightcolor}{RGB}{211, 227, 252}

\begin{sidewaystable}
    \resizebox{1.0\textwidth}{!}{
    \begin{tabular}{lccccc>{\columncolor{highlightcolor}}cccc>{\columncolor{highlightcolor}}c}
    \toprule
     & Pangea-7B & Molmo-7B-D & Llama-3.2-11B-Vision & Pixtral-12B & Qwen-2.5-VL-7B & Aya-Vision-8B & Molmo-72B & Llama-3.2-90B-Vision & Qwen-2.5-VL-72B & Aya-Vision-32B \\
    \midrule
    ('Irish', 'Ireland')         & 56.40 & 42.33 & 53.99 & 57.67 & 76.38 & 47.24 & 57.06 & 76.99 & 57.98 & 56.13 \\
    ('Swahili', 'Kenya')         & 64.10 & 49.45 & 53.11 & 60.07 & 72.53 & 54.95 & 67.77 & 79.85 & 55.31 & 66.18 \\
    ('Igbo', 'Nigeria')          & 46.00 & 40.50 & 44.00 & 41.50 & 48.00 & 34.67 & 41.50 & 52.00 & 36.55 & 38.00 \\
    ('Minangkabau', 'Indonesia') & 47.80 & 44.62 & 51.79 & 51.39 & 68.13 & 52.40 & 58.17 & 76.49 & 51.79 & 61.75 \\
    ('Sundanese', 'Indonesia')   & 53.00 & 41.00 & 44.00 & 49.00 & 73.50 & 46.50 & 52.00 & 72.50 & 56.50 & 52.53 \\
    ('Chinese', 'China')         & 74.00 & 70.10 & 63.34 & 69.45 & 89.71 & 65.16 & 75.56 & 83.60 & 85.53 & 75.24 \\
    ('Spanish', 'Mexico')        & 62.20 & 54.49 & 53.56 & 63.16 & 79.57 & 57.59 & 64.71 & 74.61 & 68.94 & 67.70 \\
    ('Tamil', 'India')           & 51.90 & 35.98 & 58.41 & 51.87 & 75.70 & 44.39 & 58.41 & 86.45 & 58.88 & 61.68 \\
    ('Hindi', 'India')           &       & 51.74 & 68.16 & 30.85 & 84.58 & 62.69 & 78.11 & 90.05 & 75.12 & 78.11 \\
    ('Spanish', 'Argentina')     & 68.30 & 57.74 & 57.36 & 69.43 & 80.75 & 64.02 & 75.47 & 78.87 & 75.85 & 75.85 \\
    ('Korean', 'South Korea')    & 70.70 & 56.55 & 59.66 & 73.45 & 85.86 & 74.39 & 74.14 & 85.17 & 77.59 & 80.00 \\
    ('Urdu', 'India')            &       & 50.45 & 54.55 & 39.09 & 80.00 & 47.27 & 69.55 & 83.64 & 64.09 & 63.93 \\
    ('Filipino', 'Philippines')  & 58.60 & 45.32 & 51.72 & 64.53 & 74.88 & 44.06 & 64.53 & 82.76 & 65.02 & 66.34 \\
    ('Chinese', 'Singapore')     & 65.60 & 70.67 & 62.26 & 68.40 & 87.26 & 66.82 & 83.02 & 85.38 & 76.42 & 79.72 \\
    ('Spanish', 'Colombia')      & 64.70 & 61.00 & 54.36 & 68.46 & 80.91 & 58.51 & 73.86 & 85.48 & 75.10 & 71.67 \\
    ('Indonesian', 'Indonesia')  & 62.10 & 53.64 & 56.31 & 62.86 & 78.83 & 56.69 & 63.83 & 81.07 & 66.50 & 67.88 \\
    ('Spanish', 'Uruguay')       & 49.80 & 44.44 & 48.25 & 58.41 & 70.16 & 43.91 & 61.27 & 69.52 & 57.78 & 61.90 \\
    ('Portuguese', 'Brazil')     & 72.90 & 68.31 & 57.75 & 73.59 & 84.86 & 66.78 & 77.46 & 85.56 & 76.76 & 78.01 \\
    ('Norwegian', 'Norway')      & 64.50 & 47.49 & 54.52 & 64.21 & 80.60 & 53.20 & 69.90 & 78.93 & 68.56 & 66.22 \\
    ('Oromo', 'Ethiopia')        & 35.50 & 43.93 & 34.11 & 35.51 & 43.46 & 32.71 & 42.06 & 46.73 & 35.05 & 36.45 \\
    ('Bengali', 'India')         & 59.10 & 47.00 & 55.59 & 48.25 & 79.72 & 49.82 & 68.88 & 84.97 & 61.27 & 64.31 \\
    ('Bulgarian', 'Bulgaria')    & 53.90 & 45.80 & 49.06 & 22.91 & 69.19 & 44.74 & 57.68 & 67.39 & 61.99 & 56.49 \\
    ('Amharic', 'Ethiopia')      & 36.30 & 33.48 & 39.32 & 32.91 & 58.37 & 29.44 & 45.30 & 62.82 & 36.48 & 29.18 \\
    ('Malay', 'Malaysia')        & 59.70 & 51.75 & 56.19 & 61.90 & 79.68 & 57.01 & 69.84 & 80.32 & 62.50 & 72.38 \\
    ('Egyptian_Arabic', 'Egypt') & 49.30 & 43.07 & 49.26 & 43.35 & 74.38 & 51.49 & 58.62 & 71.92 & 61.08 & 68.47 \\
    ('Telugu', 'India')          & 54.50 & 43.50 & 55.50 & 32.50 & 73.50 & 47.50 & 57.00 & 83.50 & 58.50 & 57.79 \\
    ('Spanish', 'Ecuador')       & 63.50 & 56.27 & 55.52 & 70.72 & 78.73 & 57.82 & 69.89 & 78.18 & 66.02 & 71.43 \\
    ('Spanish', 'Spain')         & 72.60 & 66.04 & 69.81 & 82.39 & 92.14 & 74.53 & 79.56 & 90.88 & 83.33 & 87.07 \\
    ('Kinyarwanda', 'Rwanda')    & 35.70 & 34.63 & 35.32 & 34.47 & 43.83 & 32.76 & 40.43 & 54.89 & 38.30 & 40.43 \\
    ('Javanese', 'Indonesia')    & 49.50 & 46.46 & 47.81 & 51.18 & 67.34 & 48.15 & 54.88 & 76.09 & 55.22 & 55.56 \\
    ('Romanian', 'Romania')      & 64.60 & 51.66 & 58.94 & 67.88 & 85.10 & 62.79 & 70.20 & 87.09 & 75.83 & 74.17 \\
    ('Urdu', 'Pakistan')         & 66.20 & 50.00 & 57.41 & 56.94 & 80.56 & 50.93 & 69.44 & 88.43 & 65.74 & 69.44 \\
    ('Japanese', 'Japan')        & 48.30 & 43.78 & 50.74 & 49.26 & 69.46 & 48.28 & 57.14 & 64.04 & 58.62 & 59.11 \\
    ('Breton', 'France')         & 34.60 & 30.86 & 34.57 & 35.80 & 44.20 & 34.41 & 35.06 & 48.64 & 37.78 & 39.36 \\
    ('Sinhala', 'Sri_Lanka')     & 39.10 & 28.89 & 48.00 & 28.44 & 62.05 & 28.89 & 45.78 & 67.56 & 45.50 & 39.56 \\
    ('Russian', 'Russia')        & 74.00 & 64.50 & 66.50 & 37.00 & 84.00 & 66.33 & 84.00 & 85.50 & 79.00 & 80.00 \\
    ('Marathi', 'India')         &       & 43.56 & 48.02 & 31.19 & 80.20 & 50.75 & 68.81 & 84.65 & 61.39 & 66.17 \\
    ('Spanish', 'Chile')         & 70.50 & 64.96 & 60.26 & 71.37 & 81.62 & 63.52 & 76.07 & 85.04 & 73.08 & 77.16 \\
    ('Mongolian', 'Mongolia')    & 42.30 & 33.33 & 39.42 & 39.74 & 54.81 & 28.53 & 47.76 & 55.77 & 39.10 & 36.01 \\
    \hdashline
    \textbf{avg}                          & 57.20 & 48.96 & 52.78 & 52.59 & 73.71 & 51.32 & 63.20 & 76.24 & 61.69 & 62.80 \\
    \bottomrule
    \end{tabular}
    }
    \caption{CVQA}
    \label{tab:cvqa}
\end{sidewaystable}

\definecolor{highlightcolor}{RGB}{211, 227, 252}

\begin{sidewaystable}
    \resizebox{1.0\textwidth}{!}{
    \begin{tabular}{lccccc>{\columncolor{highlightcolor}}cccc>{\columncolor{highlightcolor}}c}
    \toprule
     & Pangea-7B & Molmo-7B-D & Llama-3.2-11B-Vision & Pixtral-12B & Qwen-2.5-VL-7B & Aya-Vision-8B & Molmo-72B & Llama-3.2-90B-Vision & Qwen-2.5-VL-72B & Aya-Vision-32B \\
    \midrule
    eng_Latn & 24.70 & 26.90 & 41.58 & 43.30 & 36.40 & 40.99 & 53.81 & 51.60 & 53.80 & 46.07 \\
    spa_Latn & 46.20 & 47.80 & 50.54 & 57.83 & 57.80 & 51.36 & 69.16 & 69.68 & 72.50 & 60.46 \\
    hin_Deva & 24.30 & 30.00 & 29.75 & 17.29 & 33.80 & 33.56 & 39.08 & 39.13 & 46.20 & 34.20 \\
    nld_Latn & 37.30 & 41.50 & 39.88 & 43.71 & 46.40 & 38.86 & 51.57 & 52.26 & 57.40 & 43.65 \\
    ukr_Cyrl & 33.00 & 35.80 & 37.51 & 29.18 & 45.40 & 41.14 & 54.08 & 56.10 & 65.20 & 48.49 \\
    por_Latn & 48.80 & 47.60 & 55.15 & 59.55 & 59.20 & 55.20 & 70.70 & 68.05 & 73.30 & 59.87 \\
    arb_Arab & 20.40 & 21.50 & 38.22 & 27.23 & 36.10 & 31.22 & 40.84 & 39.79 & 36.10 & 34.48 \\
    rus_Cyrl & 25.50 & 29.20 & 26.20 & 22.31 & 28.80 & 27.82 & 38.25 & 31.54 & 39.30 & 27.08 \\
    fra_Latn & 25.50 & 33.10 & 25.72 & 29.92 & 35.20 & 31.18 & 43.96 & 37.80 & 46.20 & 28.65 \\
    pes_Arab & 21.20 & 28.30 & 28.25 & 21.70 & 30.30 & 30.23 & 35.25 & 35.72 & 35.40 & 31.12 \\
    deu_Latn & 17.20 & 19.90 & 28.67 & 44.88 & 26.60 & 43.30 & 57.06 & 50.83 & 49.30 & 44.94 \\
    hrv_Latn & 17.90 & 30.90 & 30.25 & 25.31 & 25.90 & 26.71 & 36.42 & 34.57 & 33.30 & 29.50 \\
    hun_Latn & 23.90 & 25.50 & 28.21 & 28.75 & 28.60 & 25.94 & 32.77 & 30.18 & 37.50 & 27.75 \\
    ben_Beng & 27.80 & 26.00 & 31.25 & 23.88 & 35.00 & 28.30 & 46.50 & 47.38 & 49.50 & 34.46 \\
    tel_Telu & 18.60 & 33.90 & 34.00 & 30.30 & 37.60 & 36.68 & 38.10 & 39.80 & 47.00 & 31.62 \\
    npi_Deva & 17.50 & 25.40 & 28.57 & 14.29 & 22.20 & 16.67 & 23.02 & 27.78 & 23.80 & 25.60 \\
    srp_Cyrl & 26.40 & 27.20 & 26.55 & 26.65 & 27.50 & 25.69 & 34.80 & 31.20 & 35.90 & 28.30 \\
    lit_Latn & 32.60 & 35.30 & 46.32 & 48.68 & 50.30 & 40.18 & 65.15 & 69.56 & 69.10 & 53.09 \\
    \hdashline
    \textbf{avg}      & 31.31 & 32.87 & 34.81 & 33.04 & 39.56 & 34.72 & 46.14 & 45.16 & 52.94 & 38.30 \\
    \bottomrule
    \end{tabular}
    }
    \caption{kaleidoscope}
    \label{tab:kaleidoscope}
\end{sidewaystable}

\end{document}